%% file: paper.tex
\documentclass[]{opendatalab}
\usepackage{pifont}
\usepackage[misc]{ifsym}
\usepackage{longtable}
\usepackage{listings}
\usepackage{wrapfig}
\usepackage[numbers]{natbib}
\usepackage{enumitem}
\usepackage{arydshln}
\usepackage[utf8]{inputenc}
\usepackage[T1]{fontenc}
\usepackage{xstring}
\usepackage{fontawesome5} % Add this to your preamble
\usepackage{colortbl}
\usepackage[table]{xcolor} %给表格添加颜色
\definecolor{myred}{HTML}{A83B27}
\definecolor{mygreen}{HTML}{44985C}
\usepackage{booktabs}      % 用于漂亮的表格线
\usepackage{graphicx}
\usepackage{tabularx}
\tcbuselibrary{listings, minted}

\usepackage{graphicx}         % 用于 \resizebox
\usepackage{booktabs}         % 用于 \toprule, \midrule, \bottomrule
\usepackage{tabularray}
\UseTblrLibrary{booktabs} % 这样就可以在 tblr 环境中使用 \toprule 等命令
\usepackage{xstring}          % 用于检查字符串（判断正负号）
\usepackage{makecell} % 允许表格进行手动换行
\usepackage{algpseudocode}
\usepackage{algorithm}
\usepackage{hyperref}
\usepackage{tocloft}
\makeatletter
\newcommand\andauthor{%
  \g@addto@macro\authorlist{\par\vspace{1mm}}% Inserts a paragraph break and some vertical space
}
\makeatother

\definecolor{groupcolor}{gray}{0.92}
\definecolor{mygreen}{RGB}{40, 160, 40}
\definecolor{myred}{RGB}{200, 30, 30}

\definecolor{codegreen}{rgb}{0,0.6,0}
\definecolor{codegray}{rgb}{0.5,0.5,0.5}
\definecolor{codepurple}{rgb}{0.58,0,0.82}
\definecolor{backcolour}{rgb}{0.95,0.95,0.92}
\definecolor{promptcolor}{HTML}{D1D0F2}
\definecolor{promptcolorheader}{HTML}{bdbcec}
\definecolor{LightBlue}{rgb}{0.85,0.92,1}  % 可以根据需要调整RGB值
\definecolor{lightblue}{RGB}{220,235,250}

\newcommand{\promptbox}[2]{
\begin{tcolorbox}[
top=0.3em,bottom=0.3em,left=0.5em,right=0.5em,
toptitle=0.3em,bottomtitle=0.2em,boxsep=0pt,
colframe=promptcolorheader,colback=promptcolor!50,boxrule=0.5pt,
]
\footnotesize
\end{tcolorbox}
}

\newcommand{\deltaval}[1]{%
  \text{\scriptsize
    \IfBeginWith{#1}{-}{\textcolor{myred}{(#1)}}{\textcolor{mygreen}{(#1)}}%
  }%
}

\usepackage{adjustbox}
\usepackage{booktabs} % 用于后面的表格示例
\usepackage[T1]{fontenc}
\tcbset{
  takeawaysbox/.style={
    title=Takeaway, % Keep your original title setting for now, modify if you add icon
    colback=teal!5!white,
    colframe=teal!65!black,
    fonttitle=\bfseries\sffamily\scshape\small,
    coltitle=white,
    colbacktitle=teal!50!black,
    enhanced,
    attach boxed title to top left={xshift=2mm, yshifttext=-1mm, yshift=-2mm},
    boxed title style={
        outer arc=2mm,
        arc=1.5mm, % Inner arc for the title box
        colframe=teal!65!black,
        colback=teal!50!black,
    },
    width=\linewidth,
    arc=3mm,
    boxrule=0.8pt, % Thicker rule for the main box
  }
}

\lstdefinestyle{mystyle}{
    backgroundcolor=\color{backcolour},   
    commentstyle=\color{codegreen},
    keywordstyle=\color{magenta},
    numberstyle=\tiny\color{codegray},
    stringstyle=\color{codepurple},
    basicstyle=\ttfamily\footnotesize,
    breakatwhitespace=false,         
    breaklines=true,                 
    captionpos=b,                    
    keepspaces=true,                 
    numbers=left,                    
    numbersep=5pt,                  
    showspaces=false,                
    showstringspaces=false,
    showtabs=false,                  
    tabsize=2
}

\definecolor{myLavender}{HTML}{e2d5ba}
\definecolor{myDarkBlue}{HTML}{14bc94}
\definecolor{green1}{HTML}{f3eedd}
\definecolor{purple1}{HTML}{303030}
\definecolor{green2}{HTML}{BFF6BA}
\definecolor{blue2}{HTML}{508AB2}
\definecolor{red1}{HTML}{ff6600}
\definecolor{blue1}{HTML}{0085c7}

\newtcolorbox{findbox}[2][]{
    enhanced,
    colback=green2!18,
    colframe=myDarkBlue!50,
    arc=3pt,
    title=#2,
    fonttitle=\bfseries\large,
    coltitle=blue,
    colbacktitle=myDarkBlue!60,
    attach boxed title to top center={yshift=-2mm},
    boxed title style={
        colframe=myDarkBlue,
        arc=3pt,
    },
    coltext=black!85,
    fontupper=\linespread{1.2}\selectfont,
    #1 % 允许额外参数
}

\newtcbtheorem[number within=section]{exmp}{Example}%
{beforeafter skip balanced=.01in,colback=green1!30,colframe=purple1!70,fonttitle=\small \bfseries, left=.03in, right=.03in,bottom=.02in, top=.02in}{exmp}

\title{MARTI-MARS$^2$: Scaling Multi-Agent Self-Search via Reinforcement Learning for Code Generation}
\author[1*]{Shijie Wang}
\author[3*]{Pengfei Li}
\author[1,4*]{Yikun Fu}
\author[3]{Kaifeng Liu}
\author[3]{Fangyuan Li}
\author[5]{Yang Liu}
\author[1,6]{Xiaowei Sun}
\author[1]{Zonglin Li}
\author[5]{Siyao Zhao}
\author[2]{Jian Zhao}
\author[2,9]{Kai Tian}
\author[3]{Dong Li}
\author[3]{Junqi Gao}
\author[7]{Yutong Zhang}
\author[8]{Yiqun Chen}
\author[1]{Yuqiang Li}
\author[10]{Zoe Li}
\author[3]{Weinan Zhang}
\author[1]{Peng Ye}
\author[1]{Shuyue Hu}
\author[1\textrm{\Letter}]{Lei Bai}
\author[1,2\textrm{\Letter}]{Bowen Zhou}
\author[2,9\textrm{\Letter}\dagger]{Kaiyan Zhang}
\author[1\textrm{\Letter}\dagger]{Biqing Qi}

\affiliation[1]{Shanghai AI Laboratory}
\affiliation[2]{Tsinghua University}
\affiliation[3]{Harbin Institute of Technology}
\affiliation[4]{Shanghai Jiao Tong University}
\affiliation[5]{Institute of Automation}
\affiliation[6]{Fudan University}
\affiliation[7]{High School Affiliated to Fudan University}
\affiliation[8]{Renmin University of China}
\affiliation[9]{Frontis.AI}
\affiliation[10]{University of Washington}

\contribution[*]{Equal contributions} 
\contribution[\dagger]{Project leaders}
\contribution[\textrm{\Letter}]{Corresponding authors}
\metadata[Correspondence:]{\url{qibiqing@pjlab.org.cn, zhang-ky22@mails.tsinghua.edu.cn}}
\metadata[Github:]{\url{https://github.com/TsinghuaC3I/MARTI}}

\abstract{
\textbf{Abstract} $|$ While the complex reasoning capability of Large Language Models (LLMs) has attracted significant attention, single-agent systems often encounter inherent performance ceilings in complex tasks such as code generation. To transcend these boundaries, multi-agent collaboration offers a promising avenue. However, existing frameworks typically rely on prompt-based test-time interactions or multi-role configurations trained with homogeneous parameters, which significantly limit the scope of error correction capabilities and strategic diversity.
In this paper, we propose a \textbf{M}ulti-\textbf{A}gent \textbf{R}einforced \textbf{T}raining and \textbf{I}nference Framework with \textbf{S}elf-\textbf{S}earch \textbf{S}caling (\textbf{MARTI-MARS$^{\mathbf{2}}$}), which integrates policy learning with multi-agent tree search by formulating the multi-agent collaborative exploration process as a dynamic and learnable environment. Specifically, by allowing agents to iteratively explore and refine within the environment, the framework facilitates a evolution from parameter-sharing homogeneous multi-role training to heterogeneous multi-agent training, thereby breaking through the capability limits of single-agent. Complementing the training paradigms, we introduce an efficient inference strategy MARTI-MARS$^2$-T+ to fully exploit the scaling potential of multi-agent collaboration at test time.
We conduct extensive experiments across varied model scales (8B, 14B, and 32B) on challenging code generation benchmark, which serves as a proxy for complex reasoning tasks. Notably, utilizing two collaborating 32B models, MARTI-MARS$^2$ achieves a performance of 77.7\%, outperforming strong baselines like GPT-5.1. Furthermore, MARTI-MARS$^2$ reveals a novel scaling law from multi-agent perspective: \textbf{shifting from single-agent to homogeneous multi-role and ultimately to heterogeneous multi-agent paradigms progressively yields higher RL performance ceilings, robust TTS capabilities, and greater policy diversity, suggesting that policy diversity is a critical dimension for scaling intelligence via multi-agent reinforcement learning.}
}

\begin{document}\maketitle

\begin{figure}[h]
    \centering
    \includegraphics[width=0.9\linewidth]{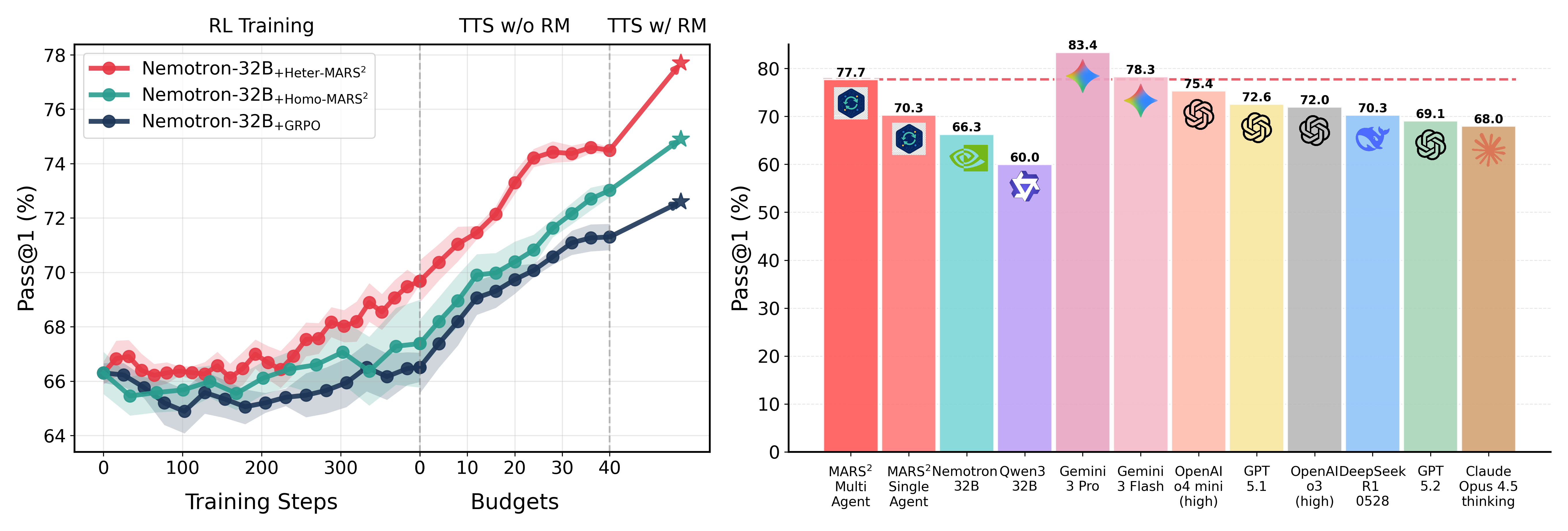}
    \caption{Multi-agent scaling laws and performance of MARTI-MARS$^2$. The left panel shows the scaling advantages in Nemotron-32B of the homogeneous multi-role method (Homo-MARS$^2$) and heterogeneous multi-agent (Nemotron-32B and Qwen3-32B) method (Heter-MARS$^2$) over single-agent (GRPO). The right panel compares the performance of MARTI-MARS$^2$ against leading open-source and closed-source LLMs.}
    \label{fig:scalinglaw}
\end{figure}
%  The left panel shows the scaling advantages of the homogeneous multi-role method (Homo-MARS$^2$) and heterogeneous multi-agent (Nemotron-32B and Qwen3-32B) method (Heter-MARS$^2$) over single-agent (GRPO). The right panel compares the performance of MARTI-MARS$^2$ against leading open-source and closed-source LLMs.

\newpage
\tableofcontents
% \thispagestyle{body}
% \fancyhead[L]{\nouppercase{\textbf{MARS$^2$: Scaling Multi-Agent Self-Search via Reinforcement Learning for Code Generation}\hspace{1em}}}
\newpage

\input{sections/1_intro}
\input{sections/3_method}

\input{sections/4_experiments}
\input{sections/5_casestudy}
\input{sections/2_preliminary}
\input{sections/7_related_work}

\section{Conclusion}
In this work, we propose and address a critical bottleneck in leveraging multi-agent systems for complex reasoning: the disconnect between test-time scaling and pre-deployment coordination training. To bridge this gap, we proposed MARTI-MARS$^{2}$, a novel multi-agent reinforced training and inference framework based on self-search scaling. By modeling the agents' collaborative exploration process as a dynamic and learnable environment, our framework enables a pivotal evolution from single-agent baselines to parameter-sharing homogeneous multi-role paradigms, and ultimately to heterogeneous multi-agent paradigms. Extensive experiments on challenging code generation benchmarks across various model scales demonstrate that MARTI-MARS$^{2}$ significantly enhances reasoning performance: with two collaborating 32B models achieving 77.7\% which surpasses strong proprietary baselines like O4-Mini(High). Besides, we unveil a new dimension of multi-agent scaling laws: the progressive shift towards homogeneous multi-role and heterogeneous multi-agent collaboration yields higher reinforcement learning performance ceilings, more robust test-time scaling capabilities, and greater policy diversity. These findings provide a scalable and effective pathway for advancing LLMs reasoning through multi-agent coordination.

While MARTI-MARS$^{2}$ has demonstrated the power of multi-agent systems in code generation, we believe that our framework serves as a foundational step toward uncovering what multi-agent scaling laws can truly achieve. The current generation of benchmarks, although challenging, often consists of isolated reasoning problems that may not fully exploit the emergent capabilities of large-scale agent coordination. To discover the next phase of multi-agent scaling laws, future research should pivot towards more promising tasks, such as those within enterprise or organizational scenarios.

\section{Acknowledgements}
This work is supported by the Shanghai Municipal Science and Technology Major Project.

\clearpage
\newpage
\bibliographystyle{plainnat}
\setcitestyle{numbers}
\bibliography{main}

\clearpage
% \newpage
\addtocontents{toc}{\protect\setcounter{tocdepth}{-1}}
\beginappendix
\input{sections/8_part4_or_appendix}

\end{document}

%% file: sections/1_intro.tex
% \begin{figure}[h]
%     \centering
%     \includegraphics[width=1\linewidth]{images/main.png}
%     \caption{Performance evaluations and scaling laws of MARTI-MARS$^2$ in LiveCodeBench(v6, 01/25–05/25)~\cite{livecodebench}. The left figure illustrates the multi-agent scaling laws of Nemotron-32B during the RL, TTS, and reward-model-based TTS phases. We compare three different training paradigms: the baseline single-agent method (GRPO), the homogeneous multi-role method (Homo-MARS$^2$), and the heterogeneous multi-agent (Nemotron-32B and Qwen3-32B) method (Heter-MARS$^2$). The right figure compares the performance of MARTI-MARS$^2$ with that of current leading open-source and closed-source LLMs. We present both the independent performance of a single agent and the performance achieved through multi-agent cooperation after heterogeneous multi-agent training.}
%     \label{fig:scalinglaw}
% \end{figure}
%  The paradigms of multi-role and multi-agent systems play a pivotal role in enhancing the upper limit of single-agent capabilities during post-training scaling and test-time scaling.

\section{Introduction}

As the complex reasoning capability of Large Language Models (LLMs) emerges as a primary focus of contemporary research, significant progress has been achieved by optimizing robust chain-of-thought during post-training phase~\cite{luo2023wizardmath,wang2024math,guo2025deepseek,shao2024deepseekmath} and by guiding efficient external search during test-time scaling phase~\cite{jin2025two,liu2025can,muennighoff2025s1,zhang2024accessing,yao2023tree}. Nevertheless, constrained by the inherent reasoning bounds of the base model's pre-training, single agents often encounter performance ceilings in complex tasks such as code generation~\cite{zhang2025survey,el2025competitive}. To break this barrier, initial attempts aim to scale inference computation based on multi-agent systems, such as AutoGen~\cite{wu2023autogenenablingnextgenllm} and MetaGPT~\cite{hong2024metagptmetaprogrammingmultiagent}. However, these frameworks rely heavily on the models' instruction-following capabilities and the pre-defined multi-agent workflow~\cite{pan2025multiagent}. Therefore, recent research (e.g., Kimi K2.5~\cite{moonshotai_kimi_k2.5}, Chain-of-Agents~\cite{li2025chain}, and AT-GRPO~\cite{zhao2025stronger}) have shifted from prompt-based test-time interaction to training LLMs for collaboration~\cite{liu2025llm,chen2025improving,chen2025mao,chen2026beyond}, positing that coordination is an intrinsic capability to be optimized prior to inference. However, most existing multi-agent collaboration frameworks remain constrained to homogeneous multi-role settings, where agents share identical base models or parameters. Such homogeneity inherently limits the diversity of reasoning pathways, thereby capping the potential of heterogeneous agents to offer complementary perspectives and robust error correction mechanisms. 

To further exploit the capabilities of heterogeneous systems, researchers design more general multi-agent training frameworks. MARFT~\cite{liao2025marft} formulates multi-agent collaboration as a sequential decision-making problem to address the challenges of asynchronous interactions and heterogeneous configurations, which fails to achieve efficient parallel generation and constrains scalability in large-scale real-time applications. Multi-Agent Reinforced Training and Inference Framework (MARTI)~\cite{marti2025} proposes the first unified multi-agent framework to support parallel training and inference, demonstrating that learned interaction patterns can guide LLMs toward more robust reasoning pathways. Building on this foundation, CoMAS~\cite{xue2025comas} employs interaction-based intrinsic rewards to train multi-agent, significantly reducing the reliance on external supervision. Nevertheless, effectively unifying multi-agent cooperation and reflective search within RL paradigms remains challenging~\cite{zhoubian2025rest, jin2025your, zuo2025ttrl}. More crucially, the field still lacks a systematic evaluation of the evolutionary trajectory from single agents to homogeneous multi-role paradigms and subsequently to heterogeneous multi-agent paradigms, which limits our empirical understanding of how collective intelligence scales.

\begin{figure}[t]
    \centering
    \includegraphics[width=0.95\linewidth]{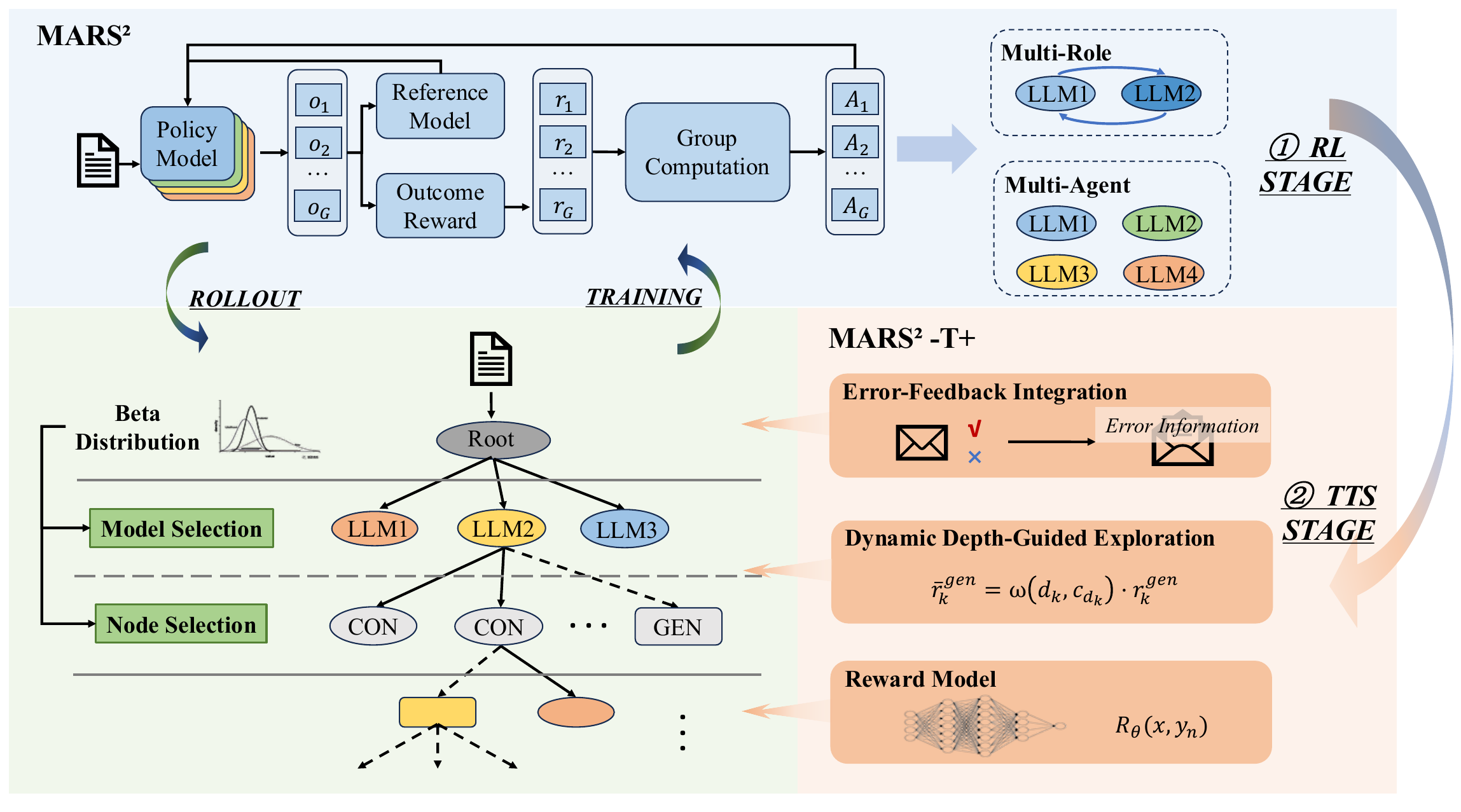}
    \caption{The framework of MARTI-MARS$^{2}$. In RL stage, the multi-role and multi-agent tree search is modeled as a learnable dynamic environment, enabling agents to improve reasoning capabilities via tree-based GRPO algorithm. In TTS stage, an enhanced method MARS$^2$-T+ is introduced, incorporating error message feedback, dynamic depth-guided exploration, and pre-trained reward model to achieve efficient inference.}
    \label{fig:framework}
\end{figure}

Therefore, building upon MARTI, we propose a \textbf{M}ulti-\textbf{A}gent \textbf{R}einforced \textbf{T}raining and \textbf{I}nference Framework with \textbf{S}elf-\textbf{S}earch \textbf{S}caling method (MARTI-MARS$^2$, denoted as MARS$^2$), which conducts large-scale RL training to systematically explore the scaling laws of multi-agent systems. We posit that multi-agent collaboration serves as a critical axis for performance scaling, where optimized interactions can yield consistent gains beyond the limitations of single-model parameters or data. Specifically, MARS$^2$ deeply integrates hierarchical multi-agent tree search based on multi-turn refinement with policy learning. As shown in Figure~\ref{fig:framework}, MARS$^2$ is designed to foster multi-agent co-evolution, which supports both parameter-sharing homogeneous multi-role agents and parameter-independent heterogeneous agents. During the training phase, we model the multi-agent collaborative search process as a learnable dynamic environment, enabling agents to acquire high-quality trajectories for improving fundamental reasoning ability through RL. Bridging to the inference phase, we further introduce an enhanced multi-agent TTS method MARS$^2$-T+. Based on structured error feedback, dynamic depth-guided exploration, and reward model guidance in multi-agent system, MARS$^2$-T+ fully exploit the scaling potential
of multi-agent collaboration at test time. As illustrated in Figure~\ref{fig:scalinglaw}, MARS$^2$ reveals multi-agent scaling laws: \textbf{the transition from single-agent to homogeneous multi-role (Homo-MARS$^2$), and ultimately to heterogeneous multi-agent paradigms (Heter-MARS$^2$), sequentially yields higher RL performance ceilings, stronger TTS capabilities, and greater diversity in the corresponding policies.} The paper collectively suggest a crucial insight: \textbf{integrating heterogeneous multi-agent scaling with RL paradigms represents a vital evolutionary trajectory beyond homogeneous multi-role paradigms, aiming to synthesize diverse cognitive capabilities.}

%MARS$^2$-T+ leverages the data collected during training to optimize the reward model and employs the trained reward model during inference to efficiently guide the search process.
% Building on this foundation, we introduce the reward-model-driven test-time scaling method MARS$^2$-T+, which uses reward signals learned during training to guide deeper inference-time search. 

The main contributions are as follows:
\begin{itemize}
    \item \textbf{We conduct the first systematic large-scale training and scaling analysis of heterogeneous multi-agent systems in complex reasoning tasks, revealing that multi-agent collaboration via RL represents the next frontier of the scaling law beyond TTS.} Extensive experiments on 8B, 14B, and 32B models uncover a robust performance hierarchy driven by interaction diversity: under equal computational budgets, the transition from a single agent to a parameter‑sharing homogeneous multi‑role paradigm yields initial performance gains (Section~\ref{homo}). Advancing to a parameter‑independent heterogeneous multi‑agent paradigm further elevates the performance upper bound (Section~\ref{heter}). The findings confirm that trained multi‑agent collaboration provides a scalable pathway capable of surpassing the performance limits of individual agents.
    \item \textbf{We propose MARTI-MARS$^2$, a unified framework that bridges collaborative multi-agent training with inference scaling.} By allowing multiple agents to actively explore and refine within a dynamic environment, MARS$^2$ facilitates a crucial evolution from parameter-sharing homogeneous multi-role training to heterogeneous multi-agent training, thereby breaking through the capability limits of single-agent optimization (Section~\ref{rl}). Complementing these training paradigms, we introduce MARTI-MARS$^2$-T+ to fully exploit the scaling potential of multi-agent collaboration with reduced computational overhead through efficient guidance and refinement (Section~\ref{tts}). The framework demonstrates that leveraging two collaborating 32B models, MARS$^2$ achieves a performance of 77.7\%, surpassing strong baselines such as O4-Mini (high) and GPT-5.1.
    \item \textbf{We provide a granular analysis of diversity scaling in multi-agent collaboration, demonstrating that the progression from single-agent to homogeneous multi-role and finally to heterogeneous multi-agent paradigms progressively elevates strategic diversity alongside performance.} To comprehensively assess this, we design multi-layered evaluations that measure diversity across semantics, algorithms, and cognitive reasoning (Section~\ref{case_study}). This analysis confirms that the multi-agent performance scaling observed in our experiments is driven by the system's enhanced capacity to discover diverse, high-quality reasoning paths, establishing diversity as a distinct and critical dimension of scaling beyond mere computational power.
    %Crucially, by evaluating these metrics strictly on functionally correct solutions, we confirm that our method generates meaningful algorithmic innovation rather than superficial variation, validating that diversity is a key driver for escaping local optima in complex reasoning.

    % \item \textbf{We propose MARS}$^{\mathbf{2}}$\textbf{, a unified framework that orchestrates population-based co-evolution by integrating multi-role and multi-agent tree search with RL.} MARS$^2$ establishes a collaborative "TTS $\leftrightarrow$ RL" closed-loop where agents actively learn from the diverse and structured trajectories generated by multi-path exploration based on tree search, transforming the search for diversity into sustainable policy improvements.
    % \item \textbf{We propose MARS$^2$-T+, a multi-agent TTS method that enables efficient cooperative reasoning.} By leveraging reward model guidance and structured error feedback across different agents, MARS$^2$-T+ significantly outperforms traditional MCTS and single-agent TTS. It demonstrates that heterogeneous multi-agent cooperation yields more stable and scalable inference trajectories under the same compute budget.
\end{itemize}

%% file: sections/3_method.tex
\section{Method}

\subsection{Overview}

MARS$^2$ establishes a unified training and inference framework centered on multi-agent self-search. The framework consists of a structured multi-agent search and policy-update process during training, and a reward-model-guided inference process, both operating on a shared tree-search structure (Figure~\ref{fig:framework}). During training, MARS$^2$ employs cooperative multi-agent tree search to generate structured reasoning trajectories. Different layers of the search tree correspond to different sub-tasks of the reasoning process, and multiple agents jointly expand nodes, propose candidates, and produce multi-path reasoning sequences. These trajectories and their associated expansion paths are then used to update the policy model, forming a tree-search-driven pipeline for trajectory generation and policy learning. During inference, MARS$^2$ applies a reward model (RM)—trained on trajectories collected during training—to evaluate the final candidate solutions. The RM provides a learned evaluation signal, enabling the system to select outputs that are more consistent with task objectives while preserving the underlying search procedure.

% \subsection{Multi-Agent Reinforcement Learning}
% \subsection{Reinforced Multi-Agent Tree Search Scaling}
\subsection{The training paradigm of MARS$^2$}\label{rl}

\label{training-paradigm}
% MARS$^2$训练阶段的整体工作流程可以概括如下：（i）收集交互数据和奖励信号，（ii）计算优势并进行数据预处理，（iii）将数据分配给代理并动态更新代理参数。
% 具体来说，在rollout阶段，多个代理被纳入树搜索过程以获取每个代理的训练数据。对于每个样本，应用扩展以生成更多样化的轨迹树。通过GRPO~\cite{shao2024deepseekmath}中的重复采样形成的原始组被由完整轨迹树表示的组所取代。基于这些树，我们收集奖励并计算优势。处理后的数据随后根据代理标识分配到相应的代理缓冲区。一旦代理缓冲区的大小达到预定义的阈值，就会动态触发该代理的参数更新。完整的算法过程在算法~\ref{MAMCTS-GRPO_algo}中提供。
The overall workflow of MARS$^2$ training phase can be summarized as follows: (i) collecting interaction data and reward signals, (ii) computing advantages and performing data preprocessing, (iii) distributing data to agents and dynamically updating agent parameters.
Specifically, during the rollout phase, multiple agents are incorporated into the tree search process to obtain training data for each agent. For every sample, the expansion is applied to generate a more diverse trajectory tree. The groups originally formed through repeated sampling in GRPO~\cite{shao2024deepseekmath} are replaced with groups represented by complete trajectory trees. Based on these trees, we collect rewards and compute advantages. The processed data are then assigned to the corresponding agent buffer according to the agent identifier. Once the size of an agent’s buffer reaches a predefined threshold, a parameter update for that agent is dynamically triggered.

\subsubsection{Multi-Agent Instantiation}
Within the unified tree-search framework, MARS$^2$ supports two multi-agent configurations: Homogeneous Multi-Agent (\textbf{Homo-MARS$^2$}) and Heterogeneous Multi-Agent (\textbf{Heter-MARS$^2$}).
In the Homo-MARS$^2$ setup, all agents share the same policy parameters, and behavioral differences arise solely from the distinct system prompts and local contexts of the search tree. In contrast, the Heter-MARS$^2$ setup assigns each agent an independent set of parameters, naturally introducing higher policy diversity and complementary behaviors during collaborative search.
Both configurations integrate seamlessly into the same tree-search procedure, enabling MARS$^2$ to maintain a consistent search structure while flexibly adjusting the agents' heterogeneity based on task demands.

% \subsubsection{MAMCTS-GRPO @Fangyuan}
\subsubsection{Data Collection and Advantage Calculation}
% 在RMATS阶段，算法动态选择不同的代理 $\mathcal{A}_j$（$j \in {1,\dots,m}$，其中 $m$ 表示代理的总数）以执行轨迹树扩展。随着每个节点的扩展，代码验证环境以标量奖励 $r$ 的形式提供即时反馈：如果所有测试用例通过，则分配奖励1，如果有任何测试用例失败，则分配0。在训练期间，所有测试用例都用于构建训练数据。相比之下，在评估期间，我们区分公开和私有测试用例：公开测试用例提供奖励信号以指导轨迹扩展，而私有测试用例仅用于报告最终评估结果。
During the rollout phase, the algorithm dynamically selects different agents $\mathcal{A}_j$ ,$j \in \{1,\dots,m\}$, where $m$ denotes the total number of agents, to perform trajectory tree expansion. As each node is expanded, the code validation environment provides immediate feedback in the form of a scalar reward $r$: a reward of 1 is assigned if all test cases pass, and 0 if any test case fails. During training, all test cases are used to construct the training data. In contrast, during evaluation, we distinguish between public and private test cases: public test cases provide reward signals to guide the trajectory expansion, whereas private test cases are used solely for reporting the final evaluation results.

% 对于给定的问题$q$，经过在MAMCTS扩展完成后，我们得到多个节点的信息集合${\Pi_{\mathcal{N}_1,\mathcal{A}_j_1},\Pi_{\mathcal{N}_2,\mathcal{A}_j_2},\dots,\Pi_{\mathcal{N}_D,\mathcal{A}_j_D}}$，其中$D$表示扩展后的节点$N$的数量,$\mathcal{A}_j_k$代表在节点$\mathcal{N}_k$处所选智能体。每个节点$\mathcal{N}_k$处使用智能体$\mathcal{A}_j_k$生成回答后，生成节点信息$\Pi_{\mathcal{N}_k,\mathcal{A}_j_k}$，其中所包含输入的提示信息$p_k$、输出的回答序列$o_k$、智能体ID$\mathcal{A}_j_k$以及相应的标量奖励$\mathbf{R}_{\mathcal{A}_{j_k}}$的信息，该奖励反映了节点$\mathcal{N}_k$选择的智能体$\mathcal{A}_j_k$生成代码的正确性。

\textbf{Reward Collection.} For a given problem $q$, after the rollout stage is completed, we obtain a set of node information $\{\Pi_{\mathcal{N}_1,\mathcal{A}_{j_1}},\Pi_{\mathcal{N}_2,\mathcal{A}_{j_2}},\dots,\Pi_{\mathcal{N}_{G_{T}},\mathcal{A}_{j_{G_{T}}}}\}$, where $G_{T}$ represents the number of tree nodes, and the subscript ${j_k}\in\{1,\dots,m\}$ denotes the agent ID selected at node $\mathcal{N}_k$. After generating a response using agent $\mathcal{A}_{j_k}$ at Node $\mathcal{N}_k$, node information $\Pi_{\mathcal{N}_k,\mathcal{A}_{j_k}}$ is created, containing the input prompt $p_{k,j_k}$, output sequence $o_{k,j_k}$, agent ID ${j_k}$, and the corresponding scalar reward $r_{k,j_k}$, which reflects the correctness of the code generated by the agent $\mathcal{A}_{j_k}$ at node $\mathcal{N}_k$.

% 为了衡量相对优势，我们将整个树中的所有节点视作一个"组"进行处理。不同于GRPO仅对单一智能体的状态-动作对进行评估，我们的设置关注的是整个多智能体系统的相对表现。尽管在组中包含多个智能体推理结果，我们仍可以利用公式\ref{grpo-norm}计算该组的组间优势，来量化每个智能体在群体中的相对优势，这样做可以帮助智能体识别相比于其他智能体的强项和不足，从而更好地指导其学习与发展。因此，在节点$\mathcal{N}_k$上智能体$\mathcal{A}_{j_k}$取得的优势为：

\textbf{Advantage Calculation.} To measure the relative advantage, we treat all nodes in the entire tree as a single group. Unlike GRPO, which evaluates the state-action pairs of individual agent, our setting focuses on the relative performance of the entire multi-agent system. Although the group contains the inference results of multiple agents, we can still use the formula \eqref{grpo-norm} to compute the advantage at the group-level, quantifying the relative advantage of each agent within the group, which helps agents identify their strengths and weaknesses compared to other agents, providing better guidance for their learning and development. Therefore, the advantage of all tokens gained by the selected agent $\mathcal{A}_{j_k}$ at node $\mathcal{N}_k$ is:

\begin{equation}
\label{advantage_caluate}
\hat{A}_{k,t}^{j_k}=\hat{A}_{k,t}=\frac{r_{k,j_k}-\text{mean}(r_{1,j_1},r_{2,j_2},\dots,r_{G_{T},j_{G_{T}}})}{\text{std}(r_{1,j_1},r_{2,j_2},\dots,r_{G_{T},j_{G_{T}}})}
\end{equation}

% 奖励收集和优势计算过程使我们能够量化每个代理相对于整个群体的表现。通过利用GRPO框架，我们可以在可能由多个代理组成的群体上下文中计算每个代理的优势。这使我们在MAMCTS框架内对代理的贡献进行更有效的评估和比较成为可能，提供了一种改善代理行为和在后续训练迭代中优化性能的方法。

% \subsubsection{MAMCTS-GRPO+: A Stable Version of MAMCTS-GRPO @Pengfei}
\subsubsection{Data dispatching and Dynamic Training}
% \subsubsection{Distributing Data for Agent Training @Fangyuan}
% 讲如何将所有数据分发给不同的agent

% 根据公式\cite{grpo-norm-mamcts}计算出每个节点对应的组间优势$\{\hat{A}_{1,j_1}, \hat{A}_{2,j_2}, \dots, \hat{A}_{D,j_D}\}$之后，将组间优势结果更新到节点信息$\{\Pi_{\mathcal{N}_1,\mathcal{A}_{j_1}},\Pi_{\mathcal{N}_2,\mathcal{A}_{j_2}},\dots,\Pi_{\mathcal{N}_D,\mathcal{A}_{j_D}}\}$中。由于树中的节点对应不同的智能体，我们将节点信息$\Pi_{\mathcal{N}_k}$添加到对应智能体的经验缓冲区$\mathcal{B}_{j_k}$中，以便后续训练。
\paragraph{Distributing Data.}  After calculating the inter-group advantages $\{\hat{A}_{1,t}^{j_1}, \hat{A}_{2,t}^{j_2}, \ldots, \hat{A}_{G_T,t}^{j_{G_T}}\}$ for each node, based on the formula (\ref{advantage_caluate}), the results are incorporated into the node information $\{\Pi_{\mathcal{N}_1,\mathcal{A}_{j_1}},\Pi_{\mathcal{N}_2,\mathcal{A}_{j_2}},\ldots,$ $\Pi_{\mathcal{N}_{G_T},\mathcal{A}_{j_{G_T}}}\}$. Since the nodes in the tree correspond to different agents, we add the node information $\Pi_{\mathcal{N}_k,\mathcal{A}_{j_k}}$ to the experience buffer $\mathcal{B}_{j_k}$ of the corresponding agent for subsequent training.

\paragraph{Asynchronously triggered Dynamic Training.}
% 在多智能体树搜索过程中，每个节点会根据可更新的 Beta 先验分布动态选择不同的 agent 进行扩展。虽然每轮 rollout 的总数据量固定，但各 agent 接收到的数据量并不均衡，导致不同 agent 的数据累积速度不一致。如果直接采用同步训练，将出现等待与阻塞，从而降低系统吞吐与计算效率。为解决这一问题，我们引入了一种动态异步触发的训练机制：为每个 agent 配置独立的数据缓存区，当某一 agent 的缓存数据量达到设定阈值时，才触发其参数更新，而无需等待其他 agent。
During multi-agent tree search, each node is expanded by dynamically selecting an agent according to an updatable Beta prior. Although the total amount of data collected per rollout is fixed, the data received by each agent is highly imbalanced across rollouts, causing different agents to accumulate data at different rates. Using synchronous updates under such conditions would lead to waiting and blocking, reducing system throughput and computational efficiency. To address this issue, we introduce a dynamically and asynchronously triggered training mechanism: each agent maintains its own data buffer, and a parameter update is triggered only when that agent’s buffer reaches a predefined threshold, without requiring synchronization with other agents.
 % $(\mathbf{N}=\mathbf{rollout\_bs} \times \mathbf{n\_sample\_per\_prompt} \times \mathbf{MCTS\ nodes})$

\subsubsection{The optimization objective of MARS$^2$}
% 具体采用的loss形式.MATS-GRPO(普通版本的算法实现)&MATS-GRPO+(添加了用于稳定rl训练的trick 版本)
% Agent 缓存区的每条数据都经过一系列处理，完整的包含了训练所需的一切变量: 完整的输入输出，action_mask，以及对应的log prob信息和经过\ref{}处理得到的归一化后的组内优势。当缓存区数量达标后可以直接使用grpo的优化目标进行训练，我们在此基础上提出了适配Multi-agent Tree search版本的算法--MATS-GRPO（Multi-agent Tree search based GRPO）\ref{main algorithm}，来对multi agent的权重进行更新：
% $$\mathcal{L}_{MATS-GRPO} = xxx$$
% 在实际使用过程中，我们发现单一的grpo优化目标在长思维链推理模型的强化学习训练中表现一般，模型能力很难继续提升甚至还会出现明显的下降。根据xxx工作，在原有的训练目标\ref{gspo loss}上进一步引入：gspo，overlong penalty，tis从多个层面缓解训练不稳定的因素：
% $$\mathcal{L}_{MATS-GRPO+} = xxx$$
% Gspo -- sequence level loss
% Overlong penalty -- 通过错误response case study，发现bad response 基本都是过长的reponse
\paragraph{MARS$^2$.}
Each sample stored in an agent’s buffer already contains all variables required for training, including the complete input–output pair, the action mask, the associated log-probability information, and the normalized intra-group advantage computed using Eq.~\ref{advantage_caluate}. Once the buffer size reaches the predefined threshold, these data can be directly used for training under the GRPO optimization objective in Eq.~\ref{grpo-update}. Building on this foundation, we further introduce an optimization objective tailored to MARS$^2$ setting.
\begin{equation}
\label{MATS-GRPO loss}
\begin{aligned}
\mathcal{J}_{\mathrm{MARS^2}}(\bm{\Theta})
&= \mathbb{E}_{q\sim\mathcal{D},\,\{o_i\}_{i=1}^{G_{\text{Tree}}}\sim\pi_{\bm{\Theta}_{\text{old}}}(\cdot|q)}\Bigg[\sum_{j=1}^{m}\frac{1}{\sum_{\hat{A}_i \in \mathcal{B}_j}|o_i|}
\sum_{\hat A_i\in\mathcal{B}_j} \sum_{t=1}^{|o_i|}\\
&\Bigg(
\min\Bigg(
    w(i,t,\theta_j)\hat{A}_{i,t},\;
    \mathrm{clip}\Big(w(i,t,\theta_j),\,1-\epsilon_{low},\,1+\epsilon_{high}\Big)\hat{A}_{i,t}
\Bigg) \\
&-\beta\,\mathbb{D}_{\mathrm{KL}}\!\big(\pi_{\bm{\theta_j}}\;\|\;\pi_{\mathrm{ref}}\big)
\Bigg)\Bigg],
\end{aligned}
\end{equation}
where we adopt the group-based advantage estimation for $\hat{A}_{i,t}$ by the equation \ref{advantage_caluate}, and denote $\bm\Theta = \{\theta_1,\theta_2,\cdots,\theta_{m}\}$ as the parameters set of all agents.The importance ratio is defined as $$w(i,t,\theta_j)=\frac{\pi_{\bm{\theta_j}}(o_{i,t}|q, o_{i, <t})}{\pi_{\bm{\theta}_{j,\text{old}}}(o_{i,t}|q, o_{i, <t})}.$$

% 

% \paragraph{RMATS+.} In practice, we observed that the aforementioned optimization objective performs suboptimally in reinforcement learning for long-chain reasoning models, often leading to stagnation or deterioration in performance. To stabilize multi-agent training, we analyzed the failure cases encountered in our experiments and leveraged prior \cite{zheng2025group,yu2025dapo,yaoyour} on reinforcement learning stabilization. Guided by these findings, we introduce  RMATS+ optimization objective: 
% 在实践中，我们观察到上述优化目标在强化学习中的长链推理模型表现不佳，通常导致性能停滞或下降。这个问题在代码生成等复杂任务中尤为突出。为了稳定多智能体训练，我们分析了实验中遇到的失败案例，并借鉴了强化学习稳定性的先前工作\cite{zheng2025group,yu2025dapo,yaoyour}。在这些发现指导下，我们引入了RMATS+优化目标：
\paragraph{MARS$^2$+.} In practice, we observed that the aforementioned optimization objective performs suboptimally in reinforcement learning for long-chain reasoning models, often resulting in performance stagnation or deterioration. This issue is particularly pronounced in complex tasks such as code generation. To stabilize multi-agent training, we analyzed the failure cases encountered in our experiments and leveraged prior work on reinforcement learning stabilization \cite{zheng2025group,yu2025dapo,yaoyour}. Guided by these findings, we introduce the MARS$^2$+ optimization objective:
\begin{equation}
\label{MATS-GRPO+ loss}
\begin{aligned}
\mathcal{J}_{\mathrm{MARS^2+}}(\bm{\Theta})
&= \mathbb{E}_{q\sim\mathcal{D},\,\{o_i\}_{i=1}^{G_{\text{Tree}}}\sim\pi_{\bm{\Theta}_{\text{old}}}(\cdot|q)}\Bigg[\sum_{j=1}^{m}\frac{1}{\sum_{\hat{A}_i \in \mathcal{B}_j}|o_i|}
\sum_{\hat A_i\in\mathcal{B}_j} \sum_{t=1}^{|o_i|}\\
&\mathbf{vllm\_kl}(j)\Bigg(
\min\left(
    s(i,j,\theta_j)\hat{A}_{i,t},\; 
    \mathrm{clip}\Big(s(i,j,\theta_j),\,1-\epsilon_{low},\,1+\epsilon_{high}\Big)\hat{A}_{i,t}
\right) \\
&-\beta\,\mathbb{D}_{\mathrm{KL}}\!\big(\pi_{\bm{\theta_j}}\;\|\;\pi_{\mathrm{ref}}\big)
\Bigg)\Bigg],
\end{aligned}
\end{equation}
% where we have
% $$\mathbf{vllm\_kl}(j) = \log\left( \frac{\pi_{\theta_j}^{\text{vllm}}(o_i|q)}{\pi_{\theta_j}^{\text{fsdp}}(o_i|q)} \right)$$
% $$s(i,j,\Theta) = \left( \frac{\pi_{\theta_j}(o_i | x)}{\pi_{\theta_{j, \text{old}}}(o_i | x)} \right)^{\frac{1}{|o_i|}} = \exp\left( \frac{1}{|o_i|} \sum_{t=1}^{|o_i|} \log \frac{\pi_{\theta_j}(o_{i,t} | x, o_{i,<t})}{\pi_{\theta_{j,\text{old}}}(o_{i,t} | x, o_{i,<t})} \right).$$
% 具体来说，RMATS+通过三个组件——GSPO、过长时间处罚和TIS——从不同角度缓解导致训练不稳定因素：
Specifically, MARS$^2$+ is extended with three components—GSPO, Overlong Penalty, and TIS~\cite{zheng2025group, yu2025dapo, yaoyour} that mitigate factors causing instability in training from different perspectives:
\begin{itemize}
\item \textbf{GSPO.} Although GSPO~\cite{zheng2025group} was originally proposed to stabilize Mixture-of-Experts (MoE) model training, our experiments indicate that it is also crucial for stabilizing long-chain reasoning in dense models. By applying a geometric mean over the importance weights across the entire reasoning chain, GSPO smooths the influence of unstable tokens:
$$s(i,j,\theta_j) = \left( \frac{\pi_{\theta_j}(o_i | x)}{\pi_{\theta_{j, \text{old}}}(o_i | x)} \right)^{\frac{1}{|o_i|}} = \exp\left( \frac{1}{|o_i|} \sum_{t=1}^{|o_i|} \log \frac{\pi_{\theta_j}(o_{i,t} | x, o_{i,<t})}{\pi_{\theta_{j,\text{old}}}(o_{i,t} | x, o_{i,<t})} \right).$$
\item \textbf{Overlong Penalty.} Through case study, we observed that most poor responses were truncated outputs. To address this, we adopted the reward shaping strategy used in DAPO \cite{yu2025dapo} to penalize overlong and incorrect responses, thereby discouraging this behavior:
$$
R_{\text{length}}(y) =
\begin{cases} 
0, & |y| \le L_{\text{max}} - L_{\text{cache}} \\[2mm]
\dfrac{(L_{\text{max}} - L_{\text{cache}}) - |y|}{L_{\text{cache}}}, & L_{\text{max}} - L_{\text{cache}} < |y| \le L_{\text{max}} \\[1mm]
-1, & L_{\text{max}} < |y|
\end{cases}
$$
\item \textbf{TIS.} To alleviate the instability caused by the mismatch between inference and training parameters, we applied the same mitigation strategy as described in TIS \cite{yaoyour} to ensure consistent reasoning behavior during training:
$$\mathbf{vllm\_kl}(j) = \log\left( \frac{\pi_{\theta_j}^{\text{vllm}}(o_i|q)}{\pi_{\theta_j}^{\text{fsdp}}(o_i|q)} \right)$$
\end{itemize}

\subsection{Test Time Scaling of MARS$^2$}\label{tts}
% 概述 tts of ramts：包括error message和rm。

% 现有的ReMASS-T在推理阶段仅使用public test case这种不完整的奖励作为引导tree展开的信号，并且节点之间缺少细粒度反馈的传递，在实际使用中我们发现存在refinement不足和节点无效评估的问题，导致TTS阶段只能激发出智能体的次优表现。
% 为了解决这些限制，我们提出了RMATS-T+, 从多模型Refinement增强和reward model分别改进了现有的自适应树搜索算法。对于推理阶段，我们提出了Multi-Agent Refinement-enhanced MCTS,增强系统的refinement.对于节点选择阶段，我们训练了一个reward model来选择节点
% Existing MARS$^2$-T relies solely on public test cases as incomplete reward signals to guide tree expansion during inference, and lacks fine-grained feedback propagation between nodes. In practice, we observe that this limitation results in insufficient refinement and inaccurate node evaluation, ultimately restricting the system to suboptimal performance in the TTS stage.
% To address these limitations, we propose MARS$^2$-T+, which introduces two key improvements over existing adaptive tree search methods: (1) a Multi-Agent Refinement-enhanced tree search method to strengthen the refinement capability during the reasoning phase, and (2) a trained reward model to guide more accurate node selection during the search process.

% 训练阶段的 rollout 使用完整的 test cases 提供精确反馈，而推理阶段只能依赖 public test cases，导致两者之间存在明显的反馈差异。此外，我们观察到 vanilla TTS 方法在搜索过程中缺乏细粒度的中间反馈，导致节点 refinement 不充分、部分分支评估不准确，最终限制了推理阶段的性能发挥。为缓解这些限制，我们提出 MARS$^2$-T+，包括两个关键改进：(1) 一个多智能体协作的 refinement-enhanced 搜索机制，以提升推理阶段的中间推理质量；(2) 一个基于训练数据拟合得到的奖励模型（Reward Model, RM），用于在搜索结束后对最终候选进行更可靠的评分，从而在推理阶段部分弥补训练–推理反馈的差异。
During training, the rollout procedure uses the full set of test cases to provide precise feedback, whereas inference must rely solely on public test cases, resulting in a clear discrepancy between training-time and test-time supervision. In addition, we observe that vanilla TTS such as AB-MCTS method~\cite{inoue2025wider} lacks fine-grained intermediate feedback during the search process, often leading to insufficient refinement and inaccurate evaluation of certain branches, which ultimately hampers inference performance. To mitigate these limitations, we introduce MARS$^2$-T+, which incorporates two key components: (i) a multi-agent, refinement-enhanced search mechanism that improves intermediate reasoning quality during inference, and (ii) a Reward Model (RM) trained on data collected in the training phase to provide more reliable scoring of final candidates, thereby partially bridging the feedback gap between training and inference.

\subsubsection{Multi-Agent Refinement-based tree search}
%% 突出多体

% 现有的研究设计了一些自适应树搜索算法（如：ab-mcts）以在TTS中平衡探索与利用，然而，这些方法未能充分挖掘多智能体协作场景下协同智能的潜力。我们的实验揭露了他们在解决复杂代码生成任务上有两点主要缺陷：
% (1) Insufficient error utilization
% (2) Limited depth for refinement
% 这些限制在多智能体系统中尤为关键，因为协调和跨智能体反馈对于有效解决问题至关重要。附录~X 提供了进一步的实证验证。
% Recent advances in inference-time scaling have explored adaptive tree search methods (e.g., AB-MCTS~\cite{inoue2025wider}) to balance exploration and exploitation in TTS. However, these approaches have yet to fully leverage the potential of collaborative reasoning in multi-agent settings.
While adaptive tree-search strategies have improved the balance between exploration and exploitation in test-time scaling, they still fall short of fully leveraging the potential of collaborative reasoning in multi-agent settings.
Our experiments reveal two key limitations of these methods when applied to complex code generation tasks: (1) \textbf{Insufficient error utilization}; (2) \textbf{Limited depth for refinement}. These limitations are particularly critical in multi-agent systems, as effective coordination and cross-agent feedback are essential for problem solving. Further empirical validation is provided in Appendix \ref{sec:appendix_mamcts}.

% 为了解决这些限制，我们提出了一个更高效的多智能体树搜索算法框架xxx，我们的方法在模型选择和节点扩展上沿用了Multi-LLM AB-MCTS的机制，我们引入了两种轻量但十分有效的机制来鼓励多智能体系统进行更精细和深层的refinement，具体而言，我们的方法通过细粒度的错误反馈提升了智能体间交互的质量，并通过更深的搜索深度显著增加了交互频次。
To address these limitations, we propose multi-agent refinement-based tree search (MARS$^2$-T+). Our approach adopts the model selection and node expansion mechanisms from Multi-Agent AB-MCTS, while integrating two lightweight yet highly effective mechanisms to encourage multi-agent systems to perform more precise and deeper refinement. Specifically, our method enhances the quality of inter-agent interactions through the error-feedback integration mechanism, which provides fine-grained diagnostic signals, and increases the frequency of such interactions through dynamic depth-guided exploration, which drives the search toward greater depth.

% 错误反馈信息集成(feedback)：当一个节点N未通过验证时，我们的方法会通过结构化的诊断反馈来增强其后续的细化步骤。这些反馈包括一组统一的通用信号——执行摘要、错误消息和相应的错误代码，随后是特定于错误类型的上下文。对于"答案错误"的情况，我们提供在公开测试集上的失败的输入输出对以及预期的输出；对于执行错误（例如 IndexError、TimeoutError），我们提供错误类型和详细的错误跟踪。这种结构化的表述使选定的代理能够推理失败的根本原因，而不仅仅是扰动之前的输出。重要的是，增强的反馈仅在细化阶段注入，以保持初始探索的多样性。额外的提示细节在附录~\ref{sec:error_feedback}中提供。
\paragraph{Error-Feedback Integration.} When a node $N$ fails validation, our method augments its subsequent refinement steps with structured diagnostic feedback. This feedback includes a unified set of general signals—an execution summary, the error message, and the corresponding error code—followed by error-type–specific context. For Wrong Answer cases, we provide the failed input–output pairs along with the expected outputs on the public test; for execution errors (e.g., IndexError, TimeoutError), we provide the error type and detailed error trace. This structured formulation enables the selected agent to reason about the underlying cause of failure rather than merely perturbing previous outputs. Importantly, the enhanced feedback is injected only during the refinement phase to preserve the diversity of initial exploration. Additional prompting details are provided in Appendix \ref{sec:error_feedback}.

% 动态深度引导探索：我们的实验表明，现有的算法倾向于生成浅层的树，系统未能利用多智能体环境中的协作潜力。为了鼓励智能体之间进行更深层次的交互，我们在对'GEN'节点的Thompson采样得到的分数引入深度感知的偏差。设d为扩展目标节点的深度，c_d为该深度已有的节点数。我们通过深度相关的权重对'GEN'得到的采样分数进行调制：$\bar{r^{gen}_k=w(d_k) \cdot r^{gen}_k$，其中$w(d)$是随深度衰减的函数，$r^{gen}_k$为在第k个节点上选择'GEN'节点的采样分数。这种方式鼓励算法进行更深度的探索，从而能够在具有挑战性的子问题上进行多步优化

\paragraph{Dynamic Depth-Guided Exploration.} Our experiments reveal that many unresolved cases stagnate in shallow regions of the search tree, where the system fails to exploit the collaborative potential of the multi-agent setting. To encourage deeper interaction among agents, we introduce a depth-aware bias into the Thompson sampling scores for the GEN action. Specifically, let $d$ denote the depth of the node being expanded, and let $c_d$ denote the number of nodes already generated at depth $d$. We modulate the original Thompson sampling score using a weighting function $w(d, c_d)$ that depends on both the depth and the node count at that depth:
\begin{equation}
    \bar{r}^{\text{gen}}_k = w(d_k, c_{d_k}) \cdot r^{\text{gen}}_k,
\end{equation}
where $r^{\text{gen}}_k$ is the Thompson sample for selecting the GEN action at node $k$.We compute the weighting function using the following formulation:
\begin{equation}
    w(d_k, c_{d_k}) = \gamma(d_k)^{\, c_{d_k}}
\end{equation}
where $\gamma \in (0,1)$ is a depth-dependent decay factor. In our experiments, $\gamma(d_k)$ is assigned a value close to 1 at shallow depths(e.g., $\gamma(1)=0.98$) and is designed to gradually decay as $d_k$ increases, thereby encouraging broad exploration during the early stages of search and progressively guiding the algorithm toward deeper and more focused exploration as the tree grows.

\subsubsection{Reward Model of MARS$^2$-T+}
Although the environment provides a definitive binary reward $r \in \{0, 1\}$, this reward signal is inherently \textbf{sparse}. % 尽管环境提供了明确的二元奖励 r 属于 {0, 1}，但这个信号是稀疏的。
%关键在于，奖励信号仅依赖于有限的公开测试用例，这一局限使得这种稀疏性进一步严重加剧。该局限性往往会导致一种我们称之为 **“奖励篡改”** 的现象：解决方案虽能获得 r=1 的奖励（即通过公开测试），却因过拟合公开测试的特性，在全面且隐藏的私有测试中以失败告终。此外，当多个解决方案均通过公开测试时，稀疏的 r=1 信号无法区分它们实际的鲁棒性差异，进而严重制约最终的 pass@1 性能（单次尝试通过率）。
Crucially, this sparsity is severely compounded by the reward's reliance on a limited set of \textbf{public test cases}. % 强调公共测试奖励的误导性。
This limitation often leads to a phenomenon we term \textbf{"reward hacking"}: solutions ($r=1$) which overfit the public tests and fail in comprehensive, hidden private tests. % 奖励欺骗/过拟合问题
Furthermore, when multiple solutions pass the public tests, the sparse $r=1$ signal is insufficient to distinguish their actual robustness, thus severely impeding the final $\texttt{pass@1}$ performance. % 无法区分鲁棒性，阻碍 pass@1 性能

%为了提供更细粒度的节点质量评估（该评估需与私有测试的成功与否真正相关），我们构建了一个专用的奖励模型（RM）R_\theta(x,y)，其训练目标是预测给定解决方案y 通过问题私有测试用例的概率。我们实现了一套完整的奖励模型流程；在后续段落中，我们将详细阐述其训练过程（涵盖数据构建、目标优化与基准测试），并说明该模型如何集成到蒙特卡洛树搜索（MCTS）中。
To furnish a more fine-grained estimation of node quality that truly correlates with private test success, we construct a dedicated Reward Model (RM), $R_\theta(x, y)$, trained to predict the probability of a given solution $y$ passing the problem's private test cases. % 引入RM及其功能
Our implementation features a complete RM pipeline, detailing the training process (including data construction, objective optimization and benchmark) before demonstrating the model's integration into the multi-agent tree search.% 承接

\paragraph{Training Data Construction.}
The foundation of a reliable RM is a large-scale dataset of code solutions with known outcomes. We curate this dataset by leveraging our policy models (agents) as the data generation engine, using the \texttt{DeepCoder} dataset as the problem pool. Further details on data collection and formatting are available in Appendix \ref{sec:appendix_rm_data}. 
% RM的基础是大型代码解法数据集。我们利用策略模型(智能体)作为数据生成引擎，在 DeepCoder 数据集上生成数据。其余见附录

\paragraph{Training Objectives.}
We initialize our RM $R_\theta$ from the \texttt{Skywork-Reward-V2-Qwen3-8B}\cite{liu2025skywork} checkpoint. The model takes $(x, y)$ as input and outputs a scalar score $s = R_\theta(x, y)$. We apply a sigmoid function $\sigma(\cdot)$ to the output score and optimize against the binary ground-truth label $r$ using the Mean Squared Error (MSE) Loss. Other objectives are discussed in Appendix \ref{sec:appendix_rm_loss}.
% 我们从 Skywork-Reward-V2-Qwen3-8B 检查点初始化 RM。模型输入 (x, y) 并输出标量分数 s。 我们对输出分数应用 sigmoid 函数，并使用均方误差(MSE)损失 相对真值标签 r 进行优化。其他目标在附录。
\begin{equation}
\label{eq:mse_loss}
L_{MSE} = (\sigma(R_\theta(x, y)) - r)^2
\end{equation}

\paragraph{Benchmark Construction for RM Evaluation.}
To rigorously evaluate our trained RM, we constructed several evaluation benchmarks from held-out datasets (\texttt{LiveCodeBench\_v6}). The detailed generation process, benchmark statistics, and evaluation metrics (e.g., Pairwise Accuracy, MSE) are deferred to Appendix\ref{sec:appendix_rm_bench}.
%为了严格评估我们训练的RM，我们从保留数据集(如 livecodebench_v6)构建了评估基准。具体细节见附录

%rm的使用
\textbf{Application of RM in MARS$^2$-T+.}
In our MARS$^2$-T+ algorithm, the RM serves as a decisive selector for the final solution rather than a heuristic guiding the search process. Specifically, we focus on the set of candidate nodes $\mathcal{N}_{pass}$ that successfully pass the public tests (i.e., $r_n=1$). Within this functionally correct subset, we invoke the RM to evaluate the quality of each solution. The final solution $y^*$ is determined by selecting the node with the highest RM score:
% 在我们的MARS^2-T+算法中，RM不作为指导搜索的启发式工具，而是作为最终解的决定性选择器。具体来说，我们关注那些成功通过公共测试用例(即 r_n=1)的候选节点集合 N_pass。在这个功能正确的子集中，我们调用 RM 来评估每个解的质量。最终解 y* 选取其中 RM 得分最高的节点：
\begin{equation}
\label{eq:selection_function}
y^* = \underset{n \in \mathcal{N}_{pass}}{\operatorname{arg\,max}} \ R_\theta(x, y_n)
\end{equation}
where $R_\theta(x, y_n)$ denotes the continuous prediction score from the RM for the solution $y_n$ at node $n$. This mechanism ensures that the algorithm prioritizes solutions that are not only verified by public tests but also aligned with the RM's quality estimation, thereby maximizing the probability of passing hidden private tests.
% 其中 R_theta(x, y_n) 表示 RM 对节点 n 的解 y_n 的连续预测分数。这种机制确保算法优先考虑那些不仅通过了公共测试验证，而且符合 RM 质量预估的解，从而最大化通过隐藏 private tests 的概率。

%% file: sections/4_experiments.tex
\section{Experiments}\label{part1}

% \begin{tcolorbox}[takeawaysbox]
% (1) \textbf{AR models are more compute-efficient} than masked diffusion objectives, making them the stronger foundational choice under equal budgets.

% (2) \textbf{SDAR enables efficient AR-to-Block Diffusion conversion}, preserving AR performance while adding parallel decoding.

% (3) \textbf{AR backbones remain superior post-conversion}, with AR-derived SDAR models consistently outperforming MDLM-based ones.
% \end{tcolorbox}

\subsection{Experimental Setup}

% 模型选择：我们在8B和14B两个规模上使用了多种代码生成模型，包括：
% Qwen3-8B Qwen3-14B
% AReal-8B/AReal 14B
% Deepcoder-14B
% 所有模型均使用相同的分词器和标准化的提示格式，以保证公平对比。

% 数据集：在训练中，我们使用了deepcoder发布的训练数据\cite{deepcoder2025}，在测试中，我们使用了LiveCodeBench (v6)\cite{livecodebench}作为我们的评测数据，这一数据在所有我们使用的模型发布后发布，以避免数据泄露的问题。

% 评估指标：我们使用Areal\cite{areal}提供的开源脚本来计算代码任务的准确率，以评估模型性能。同时，我们使用了多种指标来衡量模型性能。我们使用Pass@1来标记模型在单次响应下的准确率。使用Pass@K来标记模型在多次相应下的最高的准确率。在MCTS系统评估中，我们使用System N Node Pass@1来标记系统在N budget下最终输出的准确性。我们使用System N Node avg Pass@1来标记系统中所有满足通过public test cases条件的节点的平均准确性。另外，我们还使用 System N Node Pass@K来标记中表现最优的样本的准确性。

% 训练细节：我们在Openrlhf上修改，使其支持多智能体训练，并在其上进行单智能体和多智能体的训练。对于单智能体，我们使用一个节点，配备8张H200 GPU进行训练。对于多智能体，我们使用智能体数量个节点组成的集群，每个节点配备8张H200 GPU，每个智能体独占一个完整的节点。我们的详细训练参数如图所示\ref{tab:training-config}

% \textbf{Model Selection.} 我们实验的重点聚焦在增强推理模型在code generation任务上的能力，我们在从8B和14B level的开源推理模型中组建了模型池：Qwen3系列的8B和14B模型，AReaL-boba-2系列的8B和14B模型，以及DeepCoder-14B-Preview模型。我们在不同参数水平上组合模型池作为我们多智能体实验的Base model 选择，并且我们在实验中使用统一的提示词格式确保，以确保公平对比。
\textbf{Base LLMs Policies.} \
To comprehensively evaluate the performance of MARS$^2$ in code generation, we construct open-source model pool with parameter scales of 8B, 14B and 32B, encompassing seven recently released representative LLMs. For code-oriented LLMs, we select \textbf{AReaL-boba-2 8B/14B}~\cite{areal} and \textbf{DeepCoder-14B-Preview}~\cite{deepcoder2025} to verify that MARS$^2$ remains an effective enhancement method for improving code generation capabilities, even in models that have already been extensively optimized for code-related tasks. Additionally, we incorporate the general LLMs \textbf{Qwen3 8B/14B/32B}~\cite{yang2025qwen3} and \textbf{Nemotron 32B}~\cite{ahmad2025opencodereasoning} to further demonstrate the generalizability and universality of our method.
% Our experiments primarily focus on enhancing the reasoning capabilities of models in code generation tasks. To this end, we construct a model pool comprising 8B and 14B scales open-source reasoning models, including the \textbf{Qwen3 8B/14B}~\cite{yang2025qwen3} series, the \textbf{AReaL-boba-2 8B/14B}~\cite{areal} series, and the \textbf{DeepCoder-14B-Preview}~\cite{deepcoder2025} model. We then form different combinations of these models at various parameter scales as the base models for our multi-agent experiments. In addition, we adopt a unified prompt format across all models to ensure fair comparisons.

% We evaluate multiple code generation models at both the 8B and 14B scales, including: \textbf{Qwen3-8B} and \textbf{Qwen3-14B}, \textbf{AReal-8B} and \textbf{AReal-14B}, and \textbf{Deepcoder-14B}.

% All models use the same tokenizer and a standardized prompting format to ensure a fair comparison.

\textbf{Training Dataset.} \
We select the open-source coding dataset released by DeepCoder~\cite{deepcoder2025} as the foundation for training data. To optimize training efficiency, we systematically filter two types of extreme samples: those for which the model’s solutions achieve a perfect score and those for which the solutions receive a zero score. Following this filtering strategy, a training dataset comprising 7,992 coding prompts is constructed. RL training for our proposed method and all baselines are based on this dataset.

%We utilize the code dataset released by DeepCoder~\cite{deepcoder2025} for training. For evaluation, we adopt \textbf{LiveCodeBench(v6)}~\cite{livecodebench} as our benchmark. As LiveCodeBench(v6) was released after the development of the base LLMs, it guaranties the absence of data leakage.

\textbf{Baseline.} \
% 对于单智能体的baseline，我们选择vanilla grpo作为主要的baseline，同时为了排除RL算法增加的额外trick对结果的影响，我们在vanilla grpo中也引入了同样的RL trick，如3.2节\ref{oursmethods}中所提到的一样。对于多智能体，我们选择vanilla GRPO和Homo-MARS$^2$作为Baseline。
To verify the effectiveness of MARS$^2$ and reveal the multi-agent scaling law, we adopt vanilla GRPO as the single-agent baseline. To ensure that no additional RL-specific tricks bias the results, we incorporate the same RL techniques described in Section \ref{training-paradigm} into the vanilla GRPO setup. 
% \textbf{Evaluation Metrics.}
% We use the open-source evaluation scripts provided by AReal~\cite{areal} to compute code-task accuracy.
% Multiple metrics are used to comprehensively assess the models.
% We report Pass@1 to measure single-sample accuracy, and Pass@K to measure the best accuracy among $K$ sampled responses.
% In the MCTS-based system evaluation, we report \emph{System-$N$-Node Pass@1}, which measures the accuracy of the system's final output under an $N$-node inference budget.
% We also report \emph{System-$N$-Node Avg Pass@1}, defined as the average accuracy across all nodes whose candidate programs pass the public test cases.
% Additionally, we report \emph{System-$N$-Node Pass@K} to capture the best-case performance across all candidate samples generated within the system.

\textbf{Evaluation Benchmark and Metrics.} \ 
For evaluation, we adopt \textbf{LiveCodeBench(v6, 01/25–05/25)}~\cite{livecodebench} as our benchmark (LCB), which is released after the development of the base LLMs to guarantee the absence of data leakage. Besides, we choose three principal evaluation metrics to comprehensively evaluate the performance of MARS$^2$ from different dimensions and demonstrate the multi-agent scaling law.
\begin{enumerate}
    \item \textbf{Pass@1}. We employ \texttt{Pass@1} as the primary metric to assess the LLMs fundamental reasoning and code generation capabilities before and after RL, which measures the probability that the LLM generates a correct solution in a single sampling attempt. This metric directly reflects the improvement in individual capabilities of LLMs based on MARS$^2$.
    \item \textbf{Pass@1(MCTS)}. During the TTS phase, we conduct N complete rollouts in the structured environment established by MARS$^2$ using tree search. Among these N candidate solutions, the optimal one (such as the solution with the highest score from the reward model or the first to pass the test) is selected for final evaluation as the result, which are denoted as \texttt{Pass@1(MCTS)}. The metric reflects the effectiveness of our method in enhancing system performance through structured search during inference.
    \item \textbf{Pass@N}. The \texttt{PASS@N} metric evaluates the probability that at least one of N generated solutions passes all unit tests. This metric not only indicates the LLMs average performance but also reflects its capacity to generate both effective and diverse solutions.
\end{enumerate}

% We evaluate model performance using the open-source accuracy scripts from AReaL~\cite{areal}.
% To provide comprehensive assessment, we report several metrics at both the model and system levels.

% For single-model evaluation, we use \textbf{Pass@1} to measure one-shot accuracy and \textbf{Pass@K} to measure the best accuracy among $K$ sampled responses.

% For MCTS-based system evaluation, we additionally introduce three system-level metrics:
% (1)~\emph{System-$N$-Node Pass@1}, which measures the accuracy of the system's final output under an inference budget of $N$ nodes;
% (2)~\emph{System-$N$-Node Avg Pass@1}, the average accuracy across all nodes whose candidate programs pass the public test cases;
% and (3)~\emph{System-$N$-Node Pass@K}, which captures the best-case accuracy across all candidate programs generated within the system.

\textbf{Training Details.} \
% 我们在MARTI框架上进行智能体强化学习训练的实验，对于单智能体的实验，我们使用单节点8个H200 gpus，对于多智能体，我们为每个智能体的部署分配单节点的8张H200 gpus机器。
We extend MARTI~\cite{marti2025} framework to conduct both single-agent and multi-agent RL. For single-agent training, we use one node equipped with eight H200 GPUs.
For multi-agent training, we use a cluster consisting of as many nodes as the number of agents, with each node containing eight H200 GPUs, and each agent occupying one full node.
The complete set of training hyperparameters is summarized in Appendix~\ref{training_details}.

\subsection{Results of Homo-MARS$^2$}\label{homo}
%% SA-ReMASS提升模型本身能力的同时，也进一步提升了TTS的上限
%% 单智能体的交互仍然无法突破自身局限，导致训练的不稳定
\begin{tcolorbox}[takeawaysbox]
% (1) \textbf{Higher performance ceiling.} Homo-MARS$^2$ enhances the model’s intrinsic capabilities and further increases the upper bound of test-time scaling (TTS).
%显著的性能提升与更快的收敛速度 (Significant Performance Boost & Faster Convergence): 同质化多智能体协同探索（Homo-MARS$^2$）通过汇聚不同角色的经验，有效拓宽了搜索范围，从而显著提升了模型的性能上限和收敛速度，并能有效避免单智能体优化时容易陷入的局部最优困境。对于传统优化方法难以提升、甚至可能导致性能下降的专用代码模型（如 AReaL-8B），Homo-MARS$^2$ 依然能实现显著的性能增益。
% Moreover, it effectively avoids the local optimum pitfalls commonly encountered in single-agent optimization.
(1) \textbf{Significant performance boost \& Faster convergence.} Homo-MARS$^2$ effectively broadens the search space by aggregating the experiences of homogeneous multi-role agents, thereby enhancing the performance ceiling and convergence speed of RL. Even for specialized code LLMs where traditional optimization methods struggle to improve or may even degrade performance, Homo-MARS$^2$ is still able to achieve significant performance gains.

% (2) \textbf{Unstable training process.} Single agent interaction remains inherently constrained by its limited exploration space, which can lead to unstable training and prevents the model from overcoming its intrinsic limitations.
% 同质化设定带来的后期瓶颈 (Late-stage Bottleneck from Homogeneity): 尽管初期效果显著，但由于所有智能体共享参数，其策略会快速趋同。这种多样性的缺失限制了模型在训练后期的持续探索和性能突破，导致过早出现性能停滞。
(2) \textbf{Late-stage bottleneck from homogeneity.} Despite the remarkable initial results, parameter sharing among homogeneous agents leads their policies to rapidly converge. This loss of diversity restricts continued exploration and further performance improvements in the later stages of training, prematurely resulting in performance stagnation.
\end{tcolorbox}

% 我们首先对8B和14B级别的模型展开了全面的单智能体实验分析，实验结果表明，single agent的ReMASS（SA-ReMASS）为RL训练带来了更高的性能表现和更快的收敛速度，但同时由于单智能体探索空间的局限性，仍然无法突破模型内在的潜力。为了保证比较的公平性，我们保证了vanilla grpo和SA-ReMASS一致的训练数据量。
% We first conduct a comprehensive set of single-agent experiments on the 8B and 14B models. To ensure a fair comparison, we match the amount of training data used for vanilla GRPO and Homo-MARS$^2$. The results show that Homo-MARS$^2$ delivers higher performance and faster convergence during RL training. However, due to the inherent limitations of single-agent exploration, it remains unable to overcome the model’s intrinsic capacity ceiling.

\begin{figure}[t]
    \centering
    \includegraphics[width=1\linewidth]{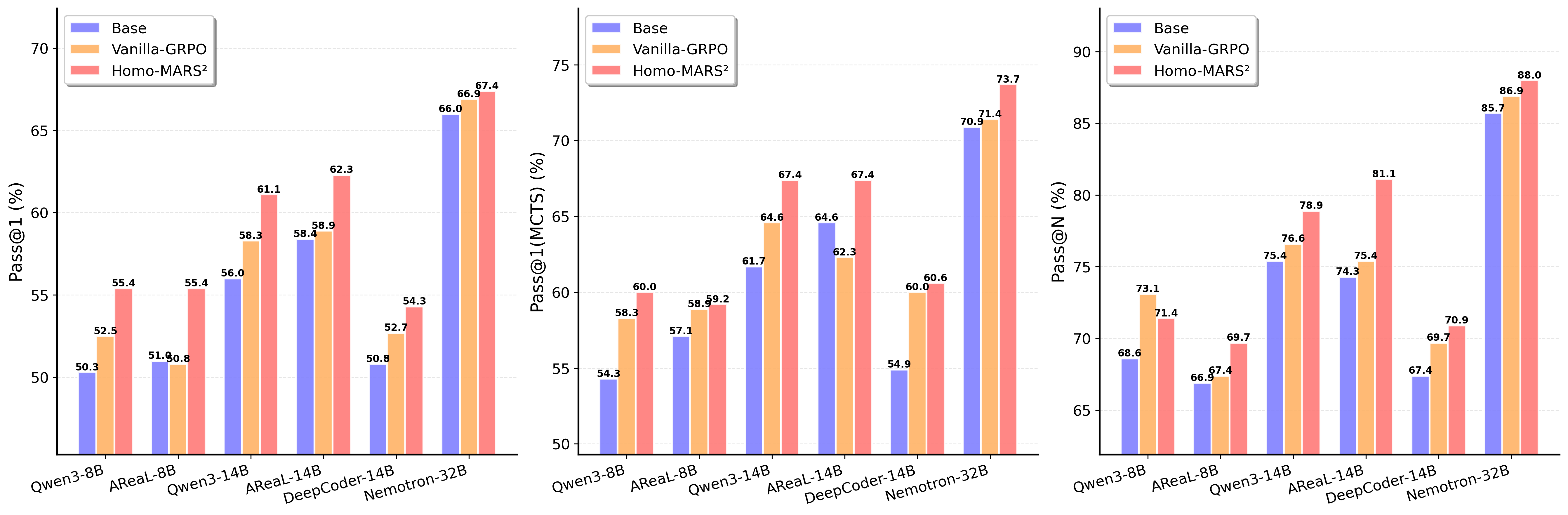}
    \caption{Experimental results of Homo-MARS$^2$ and baseline methods on LCB benchmarks. The inference budgets are $N=60$.}
    \label{fig:in-domain-parameters}
\end{figure}

%v2
%为探究MARS$^2$训练范式的有效性，我们首先在一系列8B和14B参数规模的模型上，开展了基于同构多智能体设定的综合实验（Homo-MARS$^2$）。在该设定下，所有同构的智能体角色共享同一套模型参数，协同进行探索与策略学习。为确保对比的公平性，所有报告的性能指标均在相同的训练步数下测得。详细的实验结果如表1所示。
To investigate the effectiveness of the MARS$^2$ training paradigm, we first conduct comprehensive experiments based on a homogeneous multi-role setting (Homo-MARS$^2$) for base LLMs policies with 8B, 14B,and 32B parameters. In this setting, all homogeneous agent roles share the same set of parameters and collaborate in exploration and policy learning. To ensure fairness in comparison, all reported performance metrics were measured at the same number of training steps. Detailed experimental results are presented in Figure~\ref{fig:in-domain-parameters}.

\begin{figure}[h]
    \centering

    % ---- Row 1 ----
    \begin{subfigure}[b]{0.48\textwidth}
        \centering
        \includegraphics[width=\linewidth]{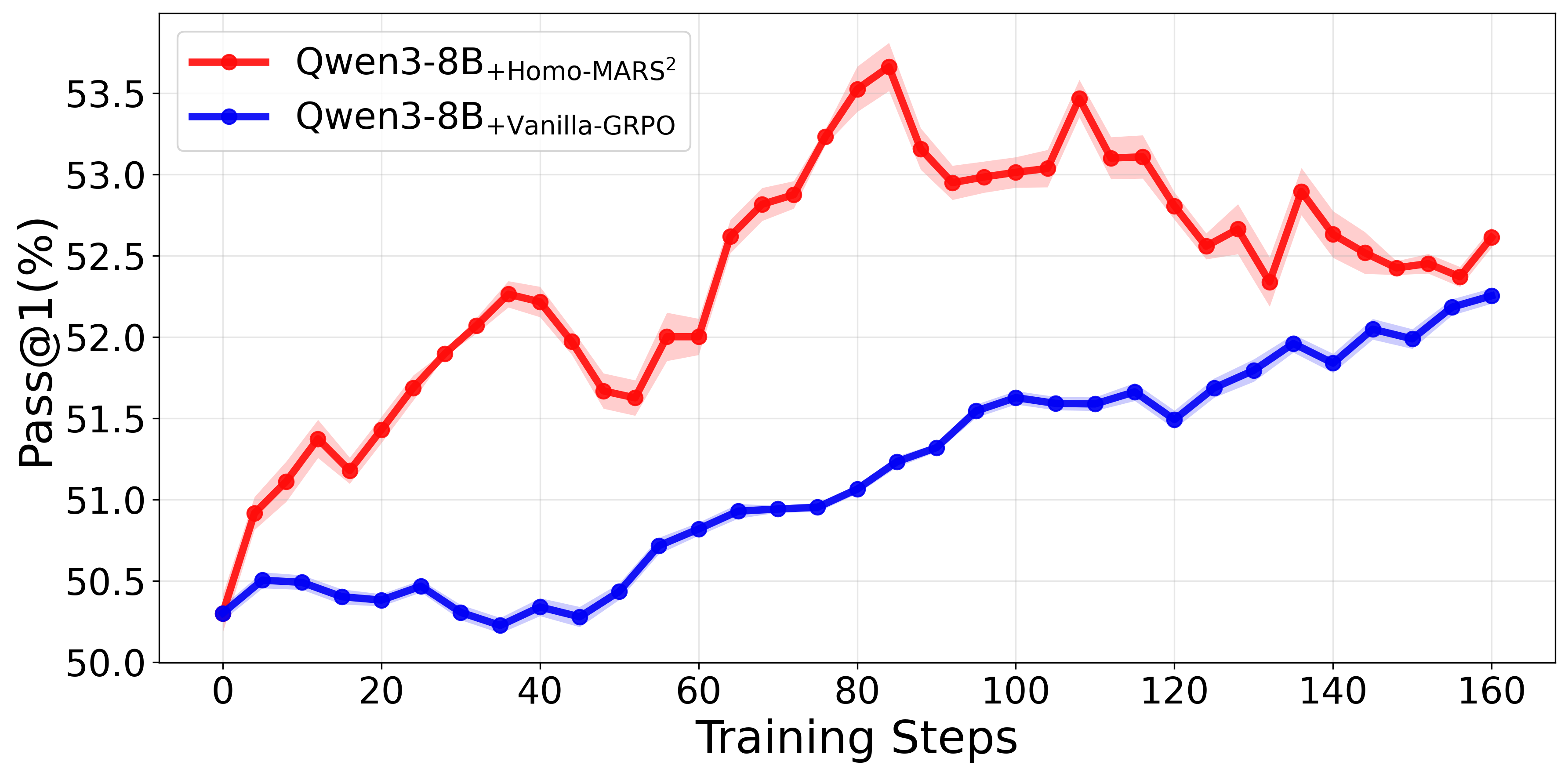}
        \subcaption{Pass@1 accuracy.}
        \label{fig:qwen3_8b_single_analysis}
    \end{subfigure}
    
    \vspace{1em}  % 行间距
    
    % ---- Row 2 ----
    \begin{subfigure}[b]{0.32\textwidth}
        \centering
        \includegraphics[width=\linewidth]{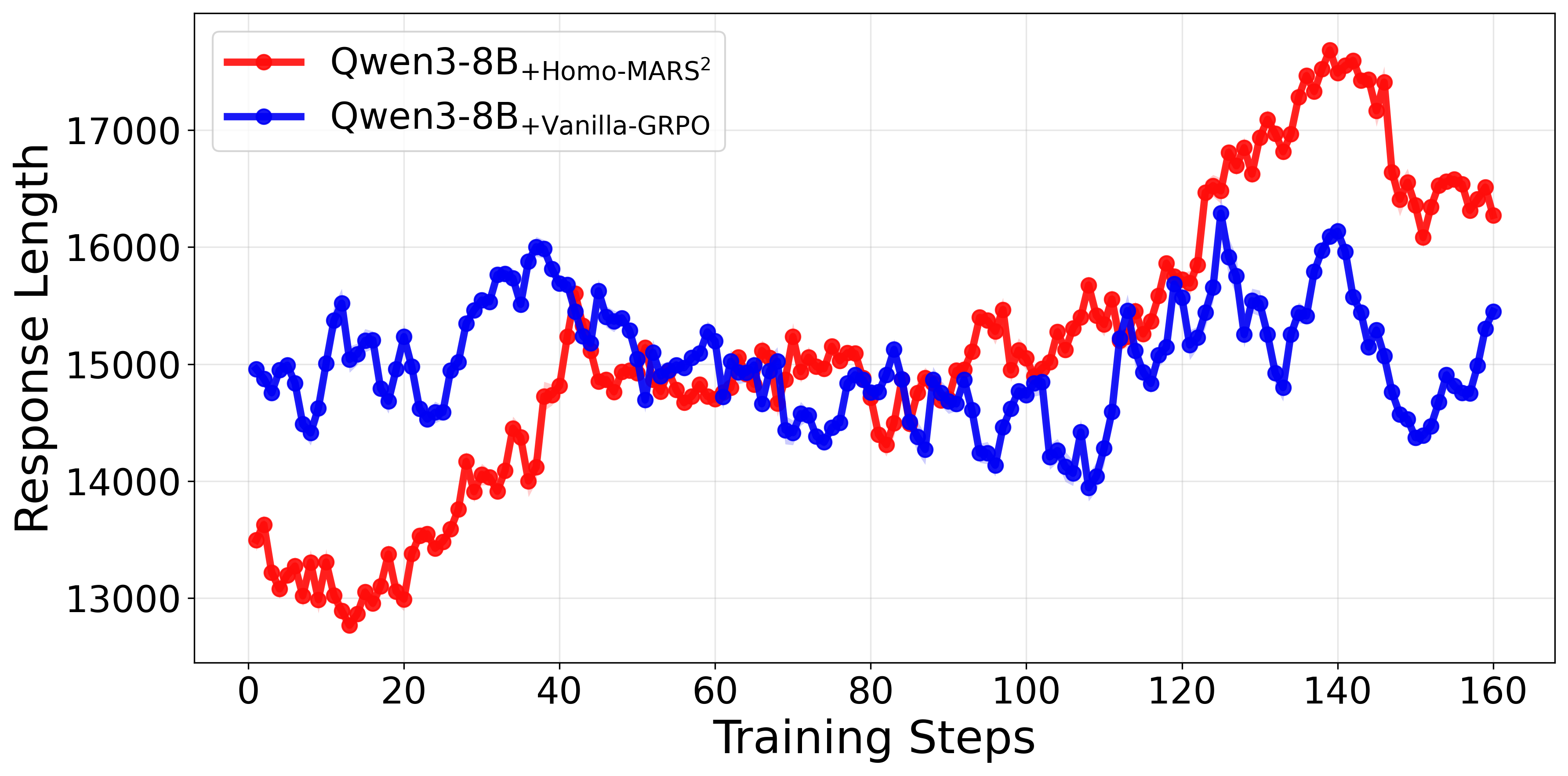}
        \subcaption{Response length.}
        \label{fig:response}
    \end{subfigure}
    \hfill
    \begin{subfigure}[b]{0.32\textwidth}
        \centering
        \includegraphics[width=\linewidth]{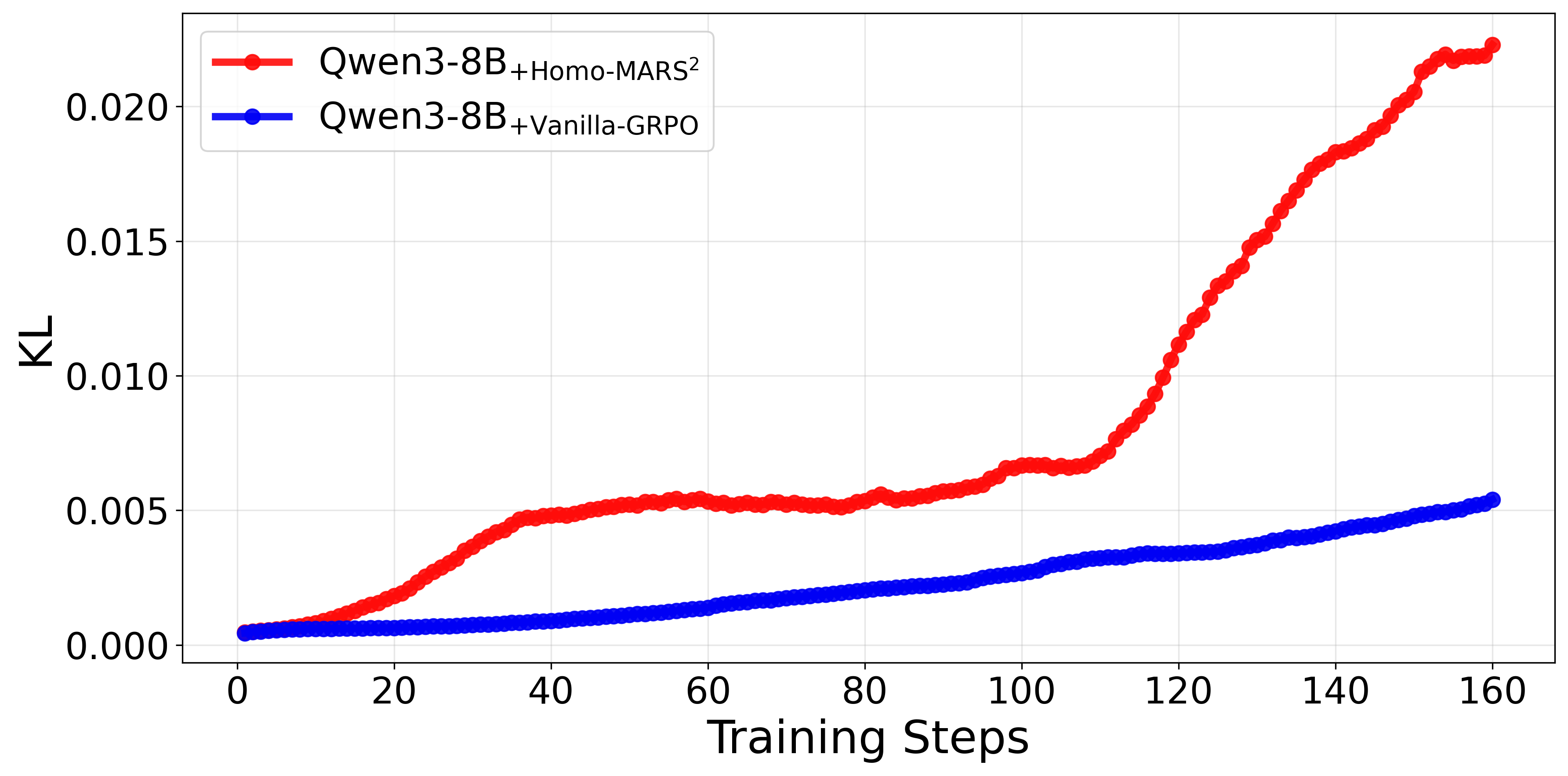}
        \subcaption{KL divergence.}
        \label{fig:kl}
    \end{subfigure}
    \hfill
    \begin{subfigure}[b]{0.32\textwidth}
        \centering
        \includegraphics[width=\linewidth]{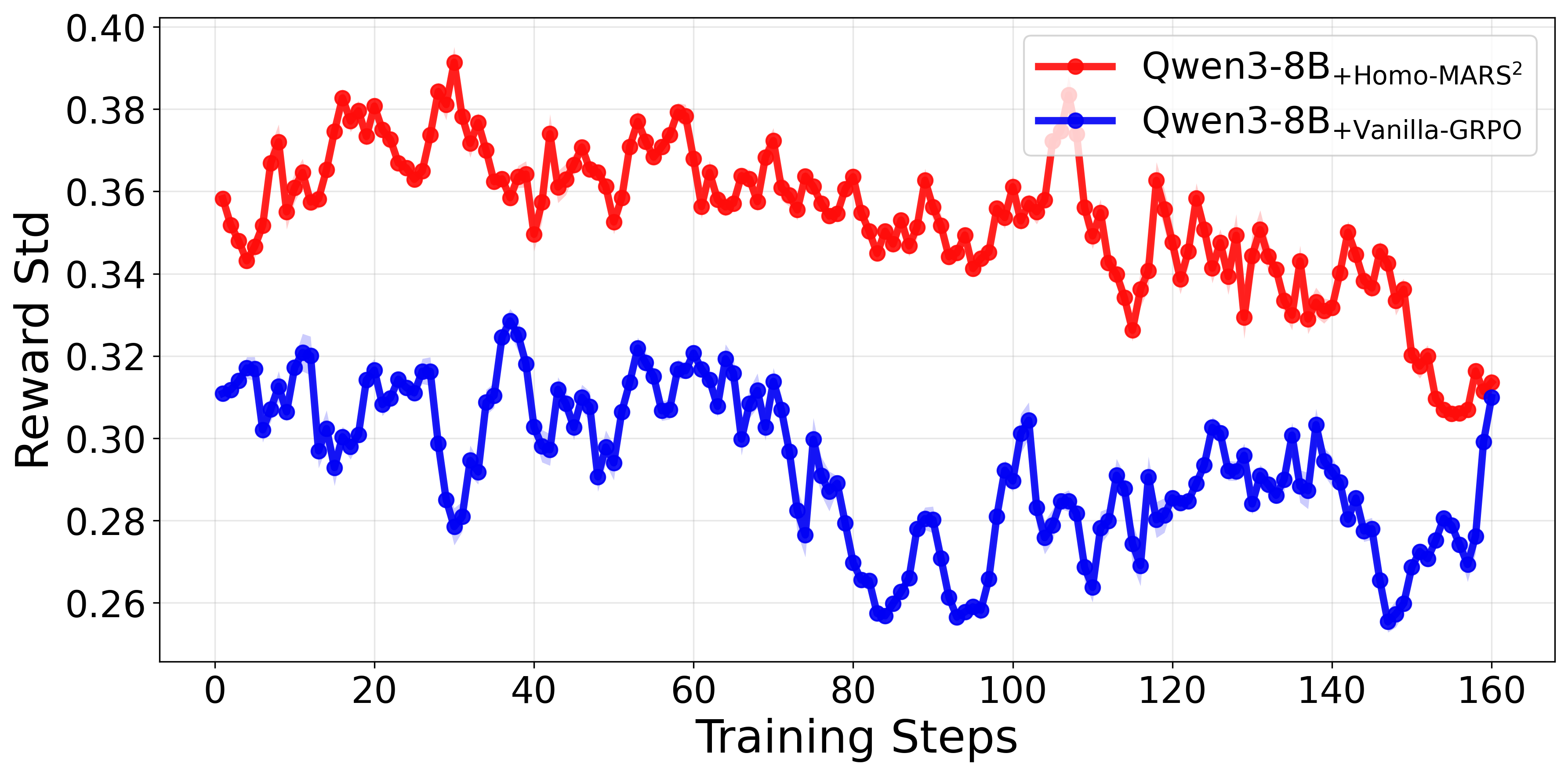}
        \subcaption{Reward standard deviation.}
        \label{fig:reward}
    \end{subfigure}

    \caption{Performance of Homo-MARS$^2$ and GRPO on Qwen3-8B across training steps.}
    \label{satraincurve}
\end{figure}

\textbf{Higher performance and faster convergence speed.} \
% 这里讲结论的时候要给一些数据支撑，可以先预填上，实验结论这里要调整一下了
% 核心结论 结合训练的曲线图说明SA-RMATS（single agent RAMTS）能获得比vanilla grpo更快的训练速度，在前期就很快收敛到model-level的最优表现。这说明通过rmats方法能够引导模型快速的探索到模型本身的最优解空间，但本质上还是无法突破模型自身的探索空间，导致在进行system-level的评估时，和vanilla训练到模型相差不大
%%%%%%%%%%%%%%%%%%%%%%%%%%%%%
% % 结论1 更高的能力上限。经过single agent ReMASS训练后的base model，几乎在所有指标上均超过了经过vanill-grpo训练的模型，在Model Level上Model-Pass@1最大的差距达4.6%，system level上System-Pass@1最大差距达到7.3%，表明我们所提出的ReMASS训练范式在强化学习训练期间提供了更有效的优化信号。
% After training with single-agent MARS$^2$(Homo-MARS$^2$), the base models consistently outperform those trained with Vanilla-GRPO across nearly all metrics. At the Model Level, the largest improvement in \texttt{Model-Pass@1} reaches 4.6\%, while at the System Level, the maximum increase in \texttt{System-Pass@1} is 7.3\%. These results indicate that the proposed MARS$^2$ training paradigm provides more effective optimization signals during reinforcement learning.
% 实验结果显示，Homo-MARS$^2$在五个不同的基础LLM策略中都展现出超过基线方法Vanilla GRPO的性能。具体而言，在模型个体能力层面，Homo-MARS$^2$ 在所有模型上的Pass@1指标均为最高。与 Vanilla GRPO 相比，Homo-MARS$^2$ 获得了最高4.6%的显著的性能增益。在系统级搜索性能层面，借助高质量的奖励模型，BoN指标的最大提升达到了5.1%，这有力地证明了 Homo-MARS$^2$ 训练出的策略能够更有效地指导蒙特卡洛树搜索（MCTS）找到优质解。如图1（a）所示，Homo-MARS$^2$的训练曲线在早期阶段拥有更快的收敛速度和更好的性能表现，这得益于其多角色探索的机制。通过让同构的多角色智能体从不同“视角”探索解空间，并通过共享参数池来聚合各自的成功经验，该方法极大地拓宽了探索空间，从而实现了对高质量训练轨迹的高效发现与利用。
The experimental results demonstrate that Homo-MARS$^2$ surpasses the baseline method across five distinct base LLMs policies. Specifically, regarding individual model capabilities, Homo-MARS$^2$ achieved the highest Pass@1 scores on all models, yielding a significant performance gain of up to 4.6\% over Vanilla GRPO. In terms of system-level search performance, the Pass@1(MCTS) metric improves by as much as 5.1\%, which strongly indicates that policies trained by Homo-MARS$^2$ more effectively guide MCTS toward optimal solutions. As illustrated in Figure~\ref{fig:qwen3_8b_single_analysis}, the training curve for Homo-MARS$^2$ exhibits faster convergence and superior performance in the early stages, attributable to its multi-role exploration mechanism. The method enables homogeneous multi-role agents to explore the solution space from diverse perspectives and aggregates their successful experiences through a shared parameter pool. Homo-MARS$^2$ substantially broadens the exploration scope, thereby facilitating the efficient discovery and utilization of high-quality training trajectories.

\textbf{Stronger deep optimization capabilities.} \
It is noteworthy that the performance of LLMs that specifically designed for code tasks (such as AReaL and DeepCoder), shows relatively limited potential for improvement through traditional optimization methods like Vanilla-GRPO. As indicated in Figure~\ref{fig:in-domain-parameters}, Vanilla-GRPO even led to a slight performance degradation in the Pass@1 metric for the AReaL-8B, decreasing from 51.0\% to 50.8\%. However, by overcoming the limitations of a single optimization trajectory and thus effectively avoiding local optima, Homo-MARS$^2$ still achieves significant performance gains on these models.

% \begin{figure}[t]
%     \centering

%     % ---- Row 1 ----
%     \begin{subfigure}[b]{0.48\textwidth}
%         \centering
%         \includegraphics[width=\linewidth]{images/comparison/qwen3_8b_single.png}
%         \subcaption{Homo-MARS$^2$ and GRPO on Qwen3-8B.}
%         \label{fig:qwen3_8b_single}
%     \end{subfigure}
%     \hfill
%     \begin{subfigure}[b]{0.48\textwidth}
%         \centering
%         \includegraphics[width=\linewidth]{images/comparison/areal_8b_single.png}
%         \subcaption{Homo-MARS$^2$ and GRPO on AReaL-boba-2-8B.}
%         \label{fig:areal_8b_single}
%     \end{subfigure}

%     \vspace{1em}  % 增加行间距，可以调整数值如 0.5em, 1em, 2em 等
%     % ---- Row 2 ----
%     \begin{subfigure}[b]{0.48\textwidth}
%         \centering
%         \includegraphics[width=\linewidth]{images/comparison/qwen_14b_single.png}
%         \subcaption{Homo-MARS$^2$ and GRPO on Qwen3-14B.}
%         \label{fig:qwen3_14b_single}
%     \end{subfigure}
%     \hfill
%     \begin{subfigure}[b]{0.48\textwidth}
%         \centering
%         \includegraphics[width=\linewidth]{images/comparison/areal_14b_single.png}
%         \subcaption{Homo-MARS$^2$ and GRPO on AReaL-boba-2-14B.}
%         \label{fig:areal_14b_single}
%     \end{subfigure}

%     \caption{Scaling trend of generation budgets and agents across different models and parameter sizes.}
%     \label{fig:tts_all}
% \end{figure}

\textbf{The inherent limitations of Homo-MARS$^2$.} \
% 当训练步数持续增加时，我们观测到一个关键现象：Homo-MARS$^2$ 在达到早期的性能峰值后，其性能提升趋于停滞，甚至在后期出现了轻微的回落，如 图 1(a) 所示。这一观察深刻地揭示了参数共享的多角色智能体设定的内在局限性。由于同构的多角色智能体共享参数，它们在训练过程中倾向于学习到相似的策略和行为模式，导致了“策略同质化”（policy homogeneity）。这种策略上的趋同虽然在初期有助于通过快速聚合有效经验来提升性能，但在训练后期，它反而成为了探索多样性的限制。如图 1(d)所示，Homo-MARS$^2$ 的奖励标准差（reward std）在后期显著下降，意味着智能体探索到的策略回报趋于一致，多样性降低，这使得策略网络过早地收敛到局部最优，从而导致了 Pass@1 性能的瓶颈。同时，我们观察到 Homo-MARS$^2$ 在训练后期表现出不稳定的行为。如 图 1(b) 和 图 1(c) 所示，其响应长度（response length）和 KL 散度均出现显著且不必要的增长。这表明，在探索能力受限的情况下，模型可能试图通过生成更长、更偏离初始策略的回答来“强行”寻找性能突破口，但这反而导致了模型行为的退化和不稳定。
%因此，虽然Homo-MARS$^2$在训练早期验证了多智能体协同框架的潜力，但其“同构”特性也为模型的持续优化带来了新的挑战。
As the number of training steps increases, we observe a key phenomenon: after Homo-MARS$^2$ reaches an early performance peak, its improvement stagnates and even exhibits a slight decline in later stages, as shown in Figure~\ref{fig:qwen3_8b_single_analysis}. Due to parameter sharing among homogeneous multi-role agents, they tend to learn similar strategies and behavioral patterns during training, leading to policy homogeneity. While such convergence in strategies facilitates rapid aggregation of effective experiences and performance gains in the early training phase, it subsequently constrains the diversity of exploration. As illustrated in Figure~\ref{fig:reward}, the reward standard deviation of Homo-MARS$^2$ drops significantly in the later stages, indicating that the agents' explored strategy returns become increasingly homogeneous, thereby reducing diversity. This causes the policy network to prematurely converge to a local optimum, resulting in a performance bottleneck for Pass@1. Meanwhile, Homo-MARS$^2$ tends to become unstable in the later training stages. As shown in Figures~\ref{fig:response} and \ref{fig:kl}, there is a marked and unnecessary increase in both response length and KL divergence. This suggests that, when exploration capacity is limited, the LLMs may attempt to achieve performance breakthroughs by forcibly generating longer responses or deviating further from the initial policy, which instead leads to behavioral degradation and unstable performance.

Therefore, although Homo-MARS$^2$ demonstrates the potential of the multi-agent framework in the initial phase of training, its homogeneous characteristic also presents new challenges for the model’s continuous optimization.

% Across all evaluated models, we observe that training with \textbf{MCTS-enhanced GRPO} consistently yields larger performance improvements than standard GRPO under the same number of training steps. This phenomenon holds for both model-level capabilities and system-level performance, indicating that incorporating MCTS into the reinforcement learning loop provides a more effective optimization signal.

% \textbf{Higher Efficiency.} \
% Models trained with \textbf{MCTS-GRPO} not only outperform both the base models and the GRPO-only models, but also exhibit better compatibility with the MCTS-based inference system. Consequently, they achieve higher accuracy under significantly smaller inference budgets, simultaneously reducing computational cost while delivering superior performance.

% \paragraph{Multi-agent RL @Pengfei}
% \subsubsection{Results of RMATS}
\subsection{Results of Heter-MARS$^2$}\label{heter}
%% MA-ReMASS引入多智能体后带来了比SA-ReMASS更高的训练上限，
%% MA-ReMASS能够打破原始RL训练方式的局限性，在已经收敛的模型上能够获得继续的提升。
\begin{tcolorbox}[takeawaysbox]
% (1) \textbf{More Significant Improvements Compared to Homo-MARS$^2$.} By introducing multi-agent interactions, Heter-MARS$^2$ surpasses the performance limits observed under Homo-MARS$^2$, providing a higher upper bound for reinforcement learning training.
(1) \textbf{Breaking Through Individual Performance Ceilings.} Heter-MARS$^2$ compels each agent to explore a broader solution space to adapt to dynamically changing peer policies, which leads to superior and more stable individual performance.

(2) \textbf{Enhancing Group Collaboration Efficiency.} At the system level, through complementarity and error correction among diverse strategies, heterogeneous multi-agent systems demonstrate high-quality collaborative problem-solving capability. Their collective performance significantly surpasses that of any single agent or homogeneous team.
%At the system level, heterogeneous agent teams exhibit a "1+1>2" collaborative problem-solving capacity through mechanisms analogous to peer review, involving complementarity and error correction. 

(3) \textbf{Revealing Multi-Agent Scaling Law.} Heter-MARS$^2$ demonstrates that scaling up agent diversity and interaction complexity serves as a powerful, orthogonal pathway to increasing model parameters. This finding establishes multi-agent collaboration as a effective paradigm for extending the reasoning frontiers of LLMs.
%reveals a novel scaling law from multi-agent perspective. The framework reveals multi-agent systems as an effective and complementary pathway to increasing model size for enhancing the capabilities of LLMs reasoning capabilities.
%not only enables a secondary performance leap by introducing collaborative strategies like MCTS at test time,
% (2) \textbf{Breaking Through the Model’s Training Ceiling.} Even for models that have already converged under standard RL algorithm, Heter-MARS$^2$ enables continued performance improvements, effectively pushing beyond previously reached ceilings.

\end{tcolorbox}
To address the performance bottleneck caused by strategy convergence in Homo-MARS$^2$ during the later stages of training, we propose a superior training framework Heter-MARS$^2$. In this framework, each agent is assigned an independent parameter set, thereby ensuring a high degree of strategy diversity throughout the exploration and learning processes. This design elegantly models the heterogeneous multi-agent tree search process as a learnable dynamic environment, endowing the system with enhanced adaptability and exploration capabilities.

\begin{figure}[t]
    \centering
    \includegraphics[width=1\linewidth]{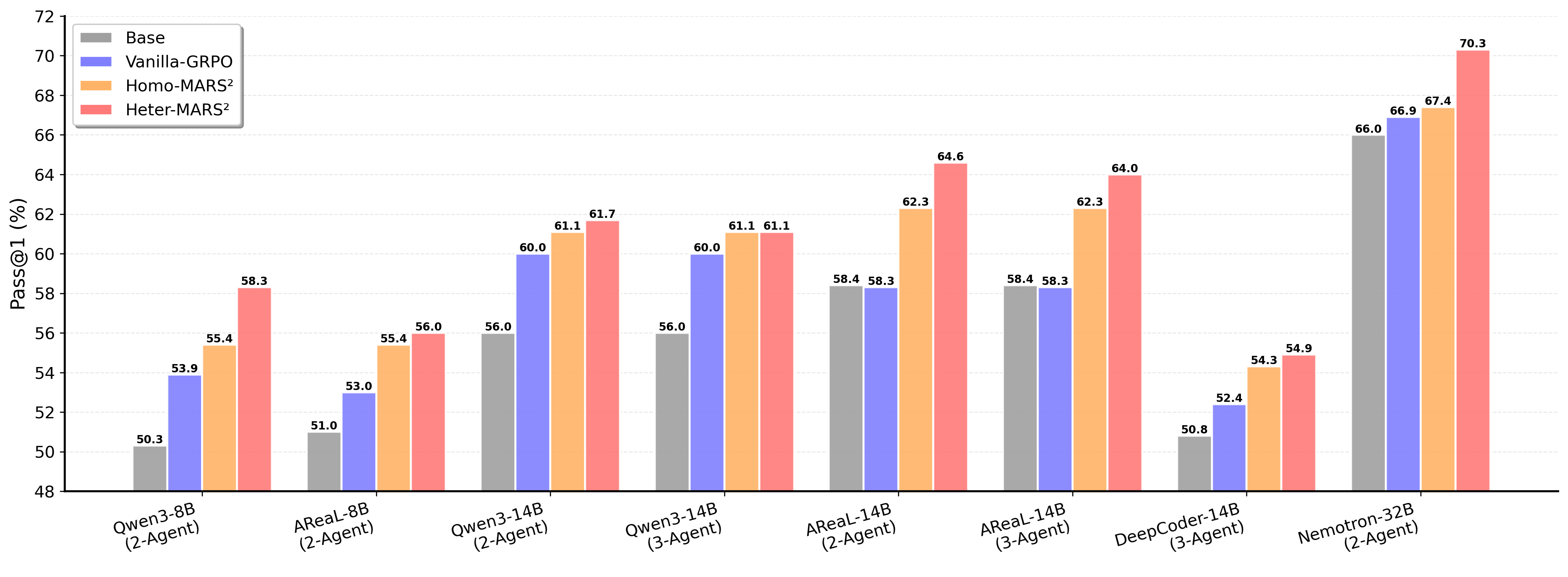}
    \caption{Pass@1 results of Heter-MARS$^2$ and baseline methods on LCB benchmarks.}
    \label{fig:heter_pass1}
\end{figure}

For a comprehensive and equitable evaluation of the effectiveness of Heter-MARS$^2$, we construct multi-agent ensembles using LLMs of 8B and 14B scales for experimentation. In the comparative analysis, we strictly ensure that the total amount of training data utilized by different methods is identical. Notably, given the performance degradation and instability exhibited by Homo-MARS$^2$ in the later stages of training, we select its best checkpoint during the entire training stages as final performance metric, thus enabling the most favorable comparison. Experimental results clearly demonstrate that the introduction of Heter-MARS$^2$ brings significant advantages: it not only effectively improves the system’s performance ceiling and the efficiency of multi-agent collaboration, but also more profoundly unlocks the intrinsic potential of the base LLMs, achieving performance that surpasses Homo-MARS$^2$ and baseline methods.

\textbf{Superior individual performance.} \
To validate the effectiveness of Heter-MARS$^2$ in enhancing the performance of individual agents, we conduct comprehensive Pass@1 performance evaluations on five base LLMs, with the experimental results illustrated in Figure~\ref{fig:heter_pass1}. The evaluation consistently demonstrates that agents trained with Heter-MARS$^2$ significantly and stably outperform the baselines and the Homo-MARS$^2$, which fully attests to the superiority and robustness of our proposed method. Taking Qwen3-8B as an example, there is a clear incremental improvement: the Pass@1 score of Heter-MARS$^2$ increased by 8.0\% compared to the base model, by 4.4\% over the Vanilla-GRPO baseline, and by 2.9\% relative to the peak performance of Homo-MARS$^2$.

\begin{figure}[h]
% \label{mavsgrpo}
    \centering

    % ---- Row 1 ----
    \begin{subfigure}[b]{0.48\textwidth}
        \centering
        \includegraphics[width=\linewidth]{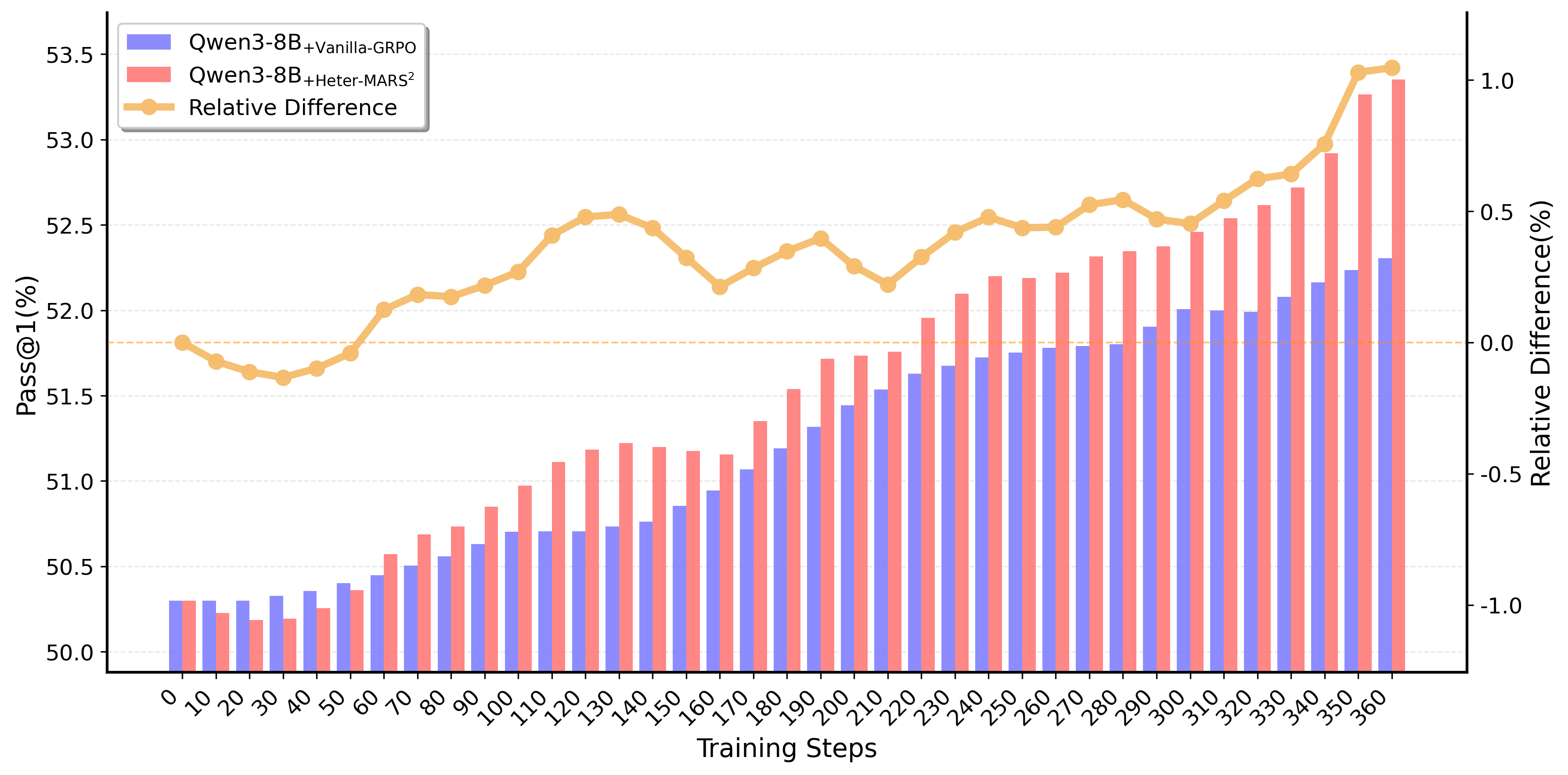}
        \subcaption{Heter-MARS$^2$ and GRPO on Qwen3-8B.}
        \label{fig:qwen3_8b}
    \end{subfigure}
    \hfill
    \begin{subfigure}[b]{0.48\textwidth}
        \centering
        \includegraphics[width=\linewidth]{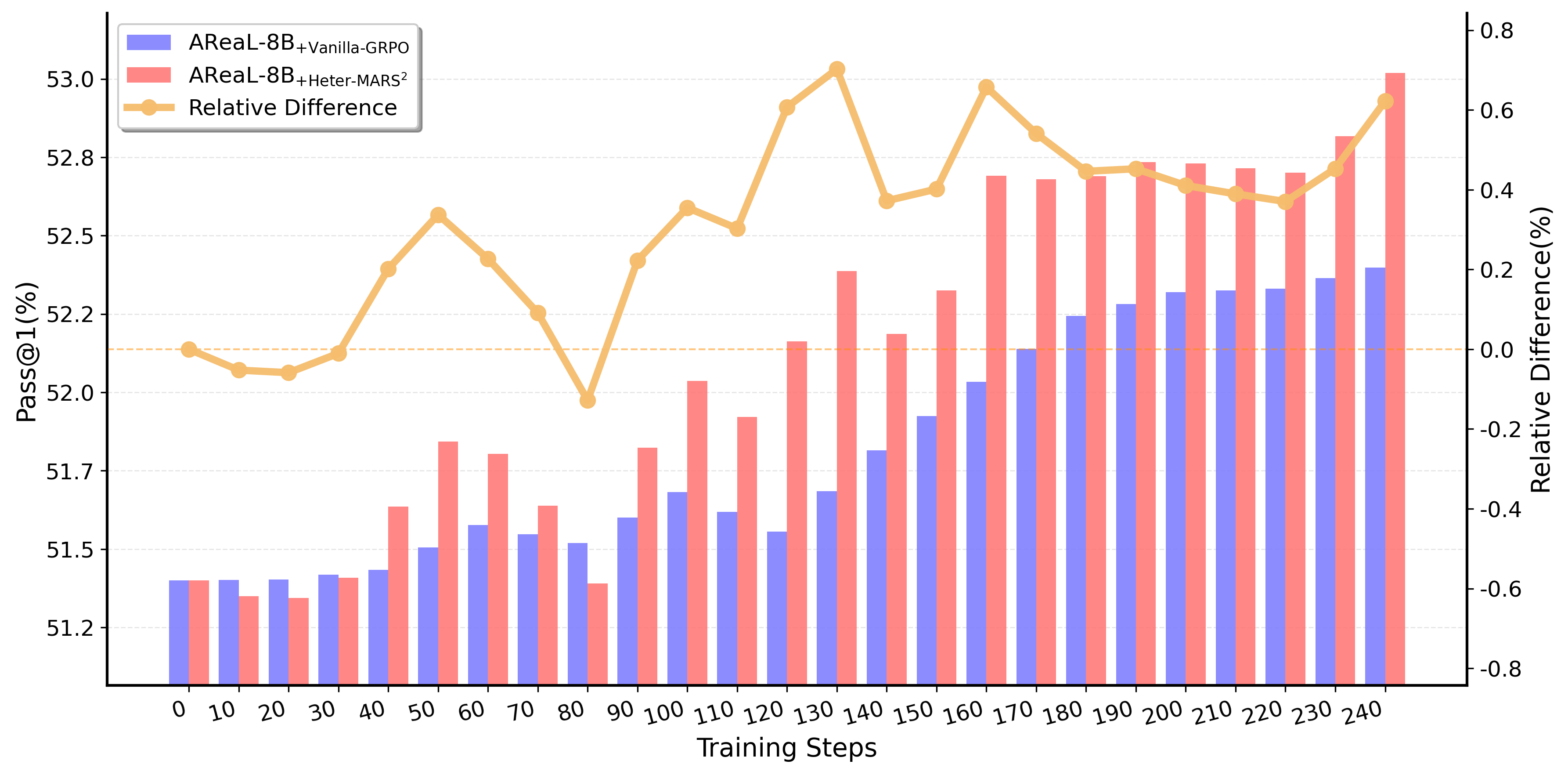}
        \subcaption{Heter-MARS$^2$ and GRPO on AReaL-boba-2-8B.}
        \label{fig:areal_8b}
    \end{subfigure}

    \vspace{1em}  % 增加行间距，可以调整数值如 0.5em, 1em, 2em 等
    % ---- Row 2 ----
    \begin{subfigure}[b]{0.48\textwidth}
        \centering
        \includegraphics[width=\linewidth]{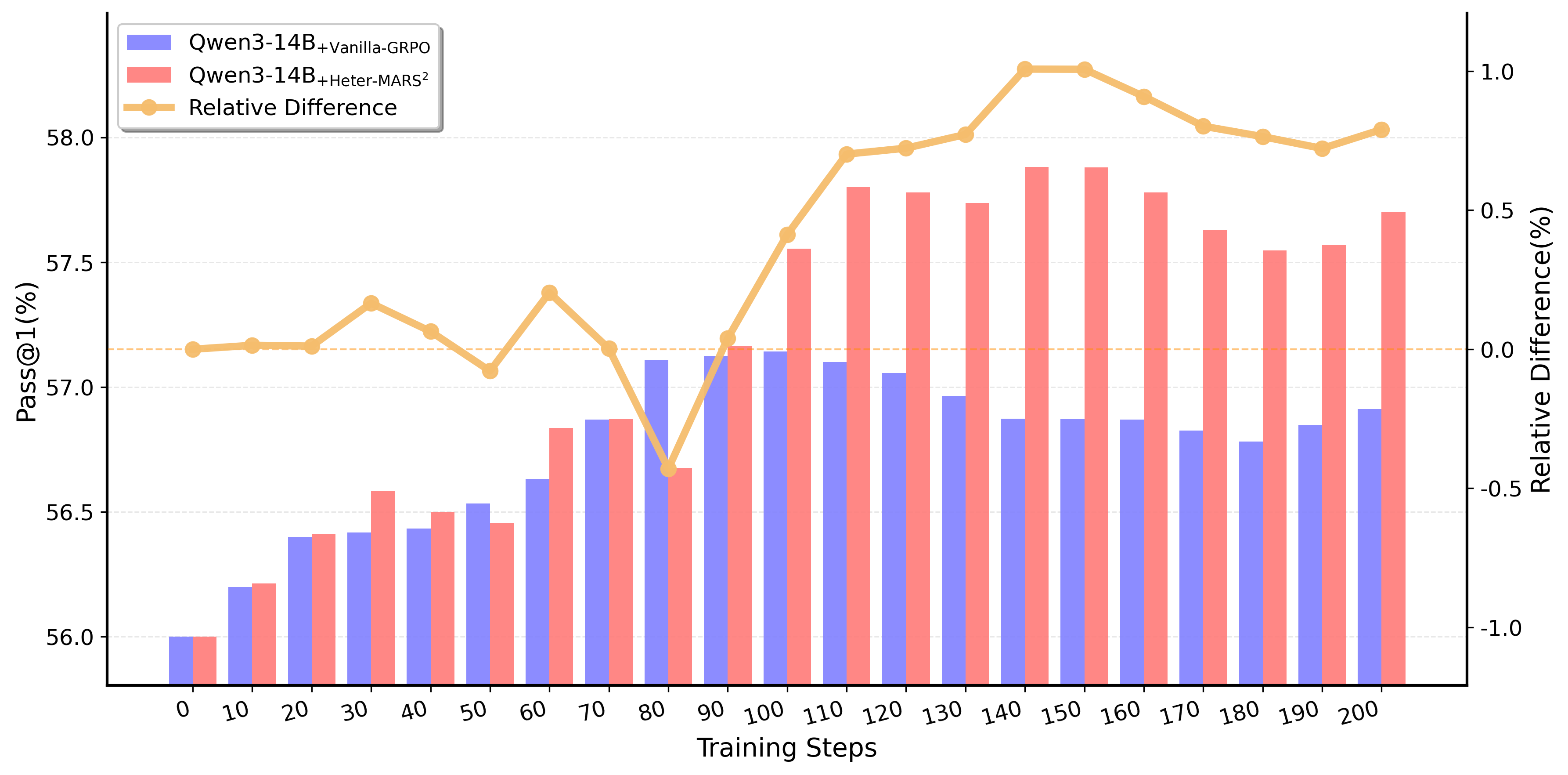}
        \subcaption{Heter-MARS$^2$ and GRPO on Qwen3-14B.}
        \label{fig:qwen3_14b}
    \end{subfigure}
    \hfill
    \begin{subfigure}[b]{0.48\textwidth}
        \centering
        \includegraphics[width=\linewidth]{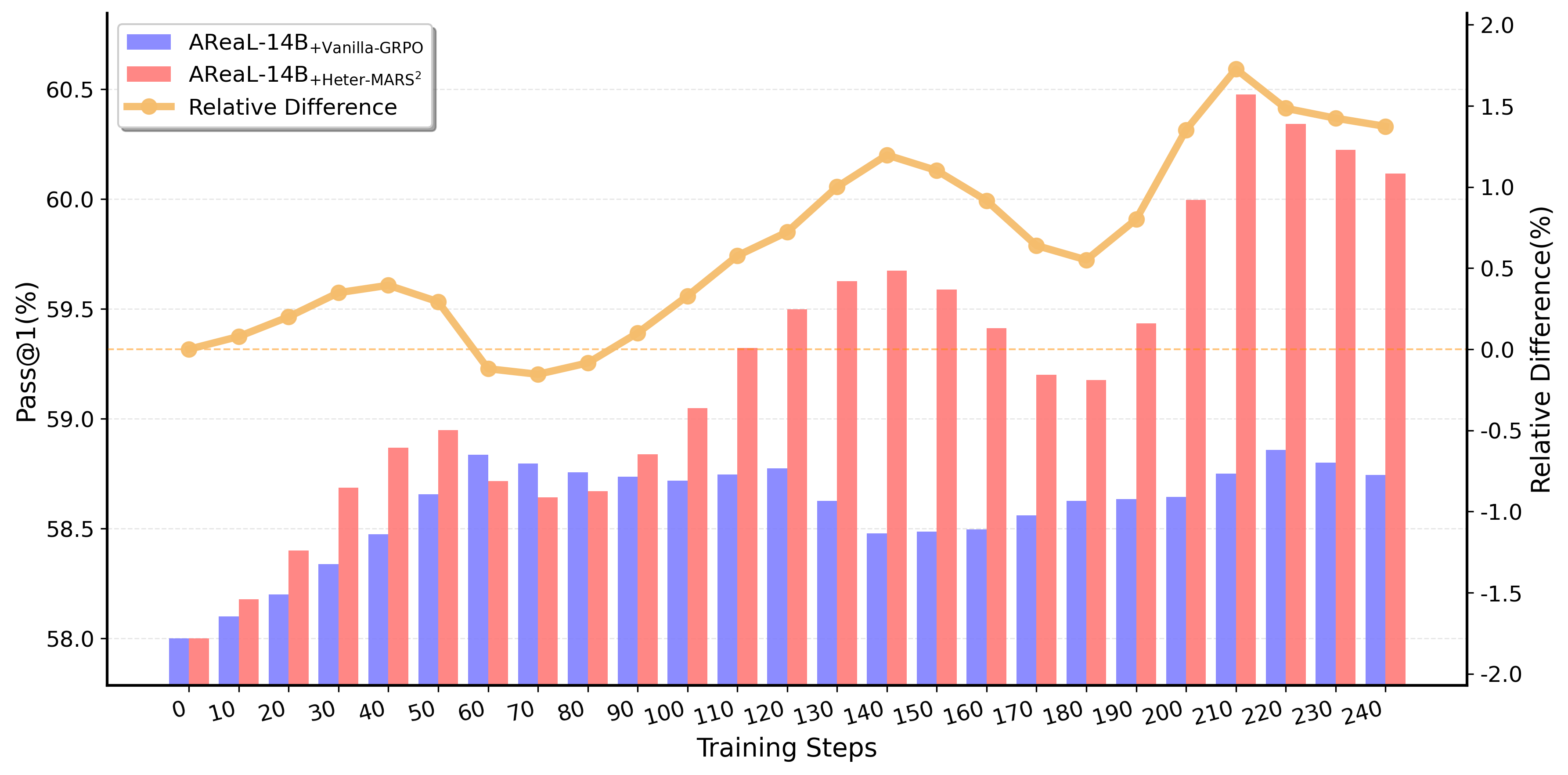}
        \subcaption{Heter-MARS$^2$ and GRPO on AReaL-boba-2-14B.}
        \label{fig:areal_14b}
    \end{subfigure}

    \caption{Scaling trend of generation budgets and agents across models and parameter sizes.}
    \label{fig:mavsgrpo}
\end{figure}

% 为了验证 Heter-MARS$^2$ 在提升单个智能体性能方面的有效性，我们在五个基础模型上进行了全面的 Pass@1 性能评估，实验结果如图 1 所示。评估结果一致表明，经过 Heter-MARS$^2$ 训练的智能体，其性能显著且稳定地超越了未经训练的原始模型（Base）、标准的自学习方法（Vanilla-GRPO）以及同构多智能体范式（Homo-MARS$^2$）。这一普遍的性能优势，充分证明了我们所提方法的卓越性与鲁棒性。以 Qwen3-8B 模型为例，我们观察到了一条清晰的性能提升阶梯：Heter-MARS$^2$ 的 Pass@1 分数较原始模型提升了 8.0%，较 Vanilla-GRPO 基线提升了 4.4%，并且相比 Homo-MARS$^2$ 的峰值性能，再次提升了 2.9%。
% 从 Vanilla-GRPO 到 Homo-MARS$^2$，再到 Heter-MARS$^2$ 的性能递进关系，揭示了不同训练范式对个体能力上限的影响。Vanilla-GRPO 通过与静态环境的自我博弈进行探索，容易陷入局部最优策略。Homo-MARS$^2$ 引入了多角色交互，通过参数共享加速了学习进程，因此性能优于 Vanilla-GRPO。然而，这种同构环境（homogeneous environment）中的策略最终会趋于同质化，导致环境的探索价值降低，从而限制了个体性能的进一步提升。Heter-MARS$^2$ 的核心优势在于，它通过维持策略多样性构建了一个持续变化的异构交互环境（heterogeneous interaction environment）。在该环境中，每个智能体必须适应由其他不同策略的智能体所构成的动态条件。这种源于异构性的挑战，迫使个体探索更广阔的解空间以寻求更鲁棒的策略，从而有效避免了在同构环境中出现的性能饱和问题，并将个体性能的上限提升至新的水平。

The progressive relationship in performance from Vanilla-GRPO to Homo-MARS$^2$ and then to Heter-MARS$^2$ reveals the influence of different training paradigms on the upper limit of individual capability. Vanilla-GRPO learns from single-agent exploration, which tends to lead to locally optimal strategies. Homo-MARS$^2$ introduces multi-role interaction and accelerates the learning process by leveraging parameter sharing, thereby outperforming Vanilla-GRPO. However, in such a homogeneous environment, strategies ultimately converge toward homogeneity, diminishing the exploratory value of the environment and thus limiting further improvement in individual performance. The core advantage of Heter-MARS$^2$ lies in its maintenance of strategy diversity, which constructs a continuously evolving heterogeneous interaction environment. In this setting, each agent must adapt to dynamic environments consisting of other agents employing distinct strategies. This heterogeneity-derived challenge forces individuals to explore a broader solution space to seek more robust strategies, effectively circumventing the performance saturation observed in homogeneous environments and elevating the upper bound of individual performance to a new level.

% 为了深入分析 Heter-MARS$^2$ 在训练过程中的动态特性，我们将其性能曲线与 Vanilla-GRPO 基线进行了比较。如图 3 所示，两种方法在学习效率和最终性能上存在显著差异。在训练初期，由于异构智能体需要探索广阔的策略空间，Heter-MARS$^2$ 的性能曲线表现出一定的波动和较为缓慢的提升速度。然而，随着训练的进行，Heter-MARS$^2$ 通过在智能体间共享多样化的成功策略，有效避免了局部最优问题，其性能曲线展现出持续且稳定的上升趋势。最终，Heter-MARS$^2$ 的 Pass@1 性能显著超越了 Vanilla-GRPO 所能达到的上限。这表明，Hetero-MARS$^2$ 通过维持策略多样性提供了一条更可持续的优化路径，从而稳定地提升单智能体RL的能力上限。
In order to conduct an in-depth analysis of the dynamic characteristics of Heter-MARS$^2$ during the training process, we compare its performance curve with the Vanilla-GRPO baseline. As shown in Figure~\ref{fig:mavsgrpo}, there are significant differences between the two methods in terms of learning efficiency and final performance. In the early stages of training, due to the need for heterogeneous agents to explore a vast policy space, the performance curve of Heter-MARS$^2$ exhibits certain fluctuations and a relatively slow rate of improvement. However, as training progresses, Heter-MARS$^2$ effectively avoids the problem of local optima through interaction and reflection among heterogeneous multi-agents, resulting in a continuously and steadily increasing performance curve that surpasses Vanilla-GRPO. This indicates that Heter-MARS$^2$ provides a more sustainable optimization pathway through maintaining policy diversity, thereby consistently elevating the performance ceiling of single-agent RL.

\begin{figure}[h]
    \centering
    \includegraphics[width=1\linewidth]{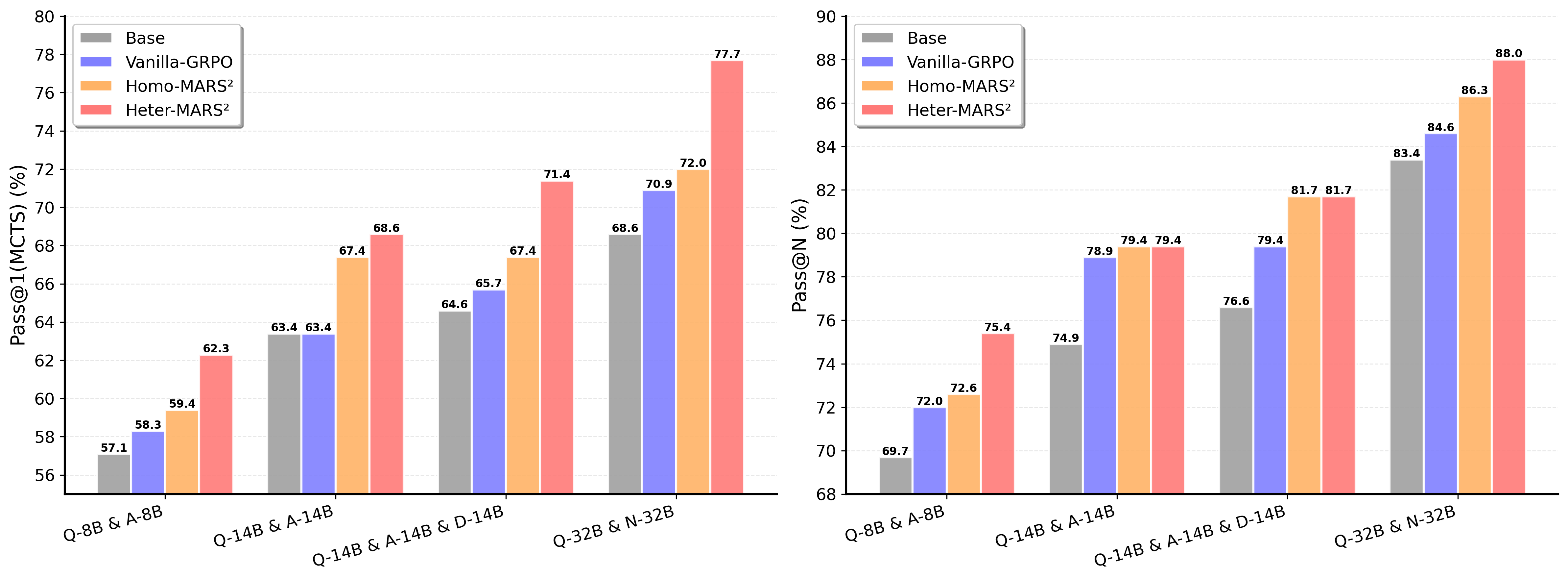}
    \caption{Pass@1(MCTS) and Pass@N results of Heter-MARS$^2$ and baseline methods on LCB benchmarks. The inference budgets are $N = 60$. For simplicity, Q denotes the Qwen3, A denotes the AReaL, D denotes the DeepCoder, N denotes the Nemotron.}
    \label{fig:heter-system}
\end{figure}

\textbf{Enhanced multi-agent collaboration.} \
In addition to enhancements in individual performance, Heter-MARS$^2$ significantly promotes superior system-level collaboration and feedback mechanisms, as demonstrated by its performance in the TTS stage shown in Figure~\ref{fig:heter-system}. A heterogeneous agent team composed of Qwen3-8B and AReaL-8B, trained with Heter-MARS$^2$, achieved a Pass@1(MCTS) score of 62.3\%, marking a notable increase of 5.2 percentage points compared to the base model (57.1\%), and outperforming Vanilla-GRPO (58.3\%) and Homo-MARS$^2$ (59.4\%). Similarly, this upward trend can be observed in the 14B-level model combination, with Heter-MARS$^2$ achieving a maximum performance of 71.2\%. Furthermore, the Pass@N metric exhibits consistent and stable improvements, confirming the comprehensive enhancement of system collaborative problem-solving capabilities provided by our method.

% 除了个体性能的增益，heter-mars还显著促进了卓越的系统级协作与反馈机制，这在对TTS阶段的表现中展现。经过Heter-MARS训练的Qwen3-8B和AReaL-8B组成的异构智能体团队，其Best-of-N（BoN）得分达到了62.3%，相较于基线（57.1%）有了5.2个百分点的显著增长，并且超越了传统Vanilla-GRPO（58.3%）和同构Homo-MARS（59.4%）训练的结果。类似地，在14B级别的模型组合同样观察到了这种增强趋势，并且通过Heter-MARS最高能达到71.4%的bon。此外，Pass@N指标也展现出了一致且稳定的提升，进一步证实了该方法对系统协同解决问题能力的全面增强。
% Heter-MARS通过异构智能体间的交互与反馈，使得多智能体系统能够超越单智能体训练后能力的简单叠加。具体而言，不同模型（如 Qwen3 和 AReaL）拥有各自的知识边界和推理偏好，这些差异被Heter-MARS转化为一种互补优势。多智能体通过高质量的“同行评审”，实现相互启发和纠正偏差，从而共同探索出更优的解题路径。
Specifically, Heter-MARS$^2$ enables multi-agent systems to surpass the simple aggregation of capabilities obtained through single-agent RL by leveraging interactions and feedback among heterogeneous agents. Different agents (such as Qwen3 and AReaL) possess unique knowledge boundaries and reasoning inclinations, which are transformed into complementary advantages by Heter-MARS$^2$. Through high-quality “peer review”, multiple agents inspire and correct each other's biases, collaboratively exploring more optimal problem-solving pathways.

\begin{figure}[h]
    \centering

    % ---- Row 1 ----
    \begin{subfigure}[b]{0.48\textwidth}
        \centering
        \includegraphics[width=\linewidth]{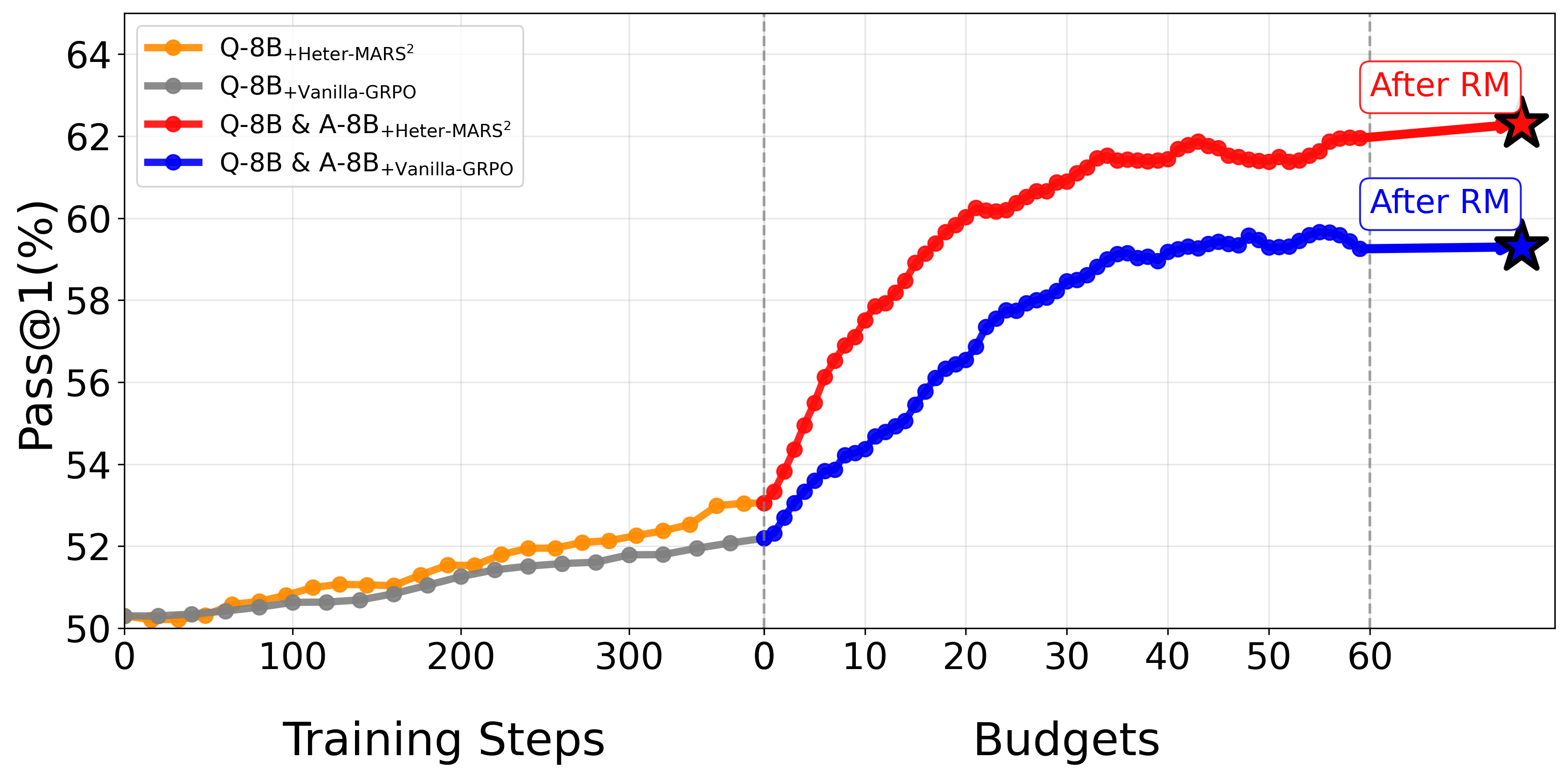}
        \subcaption{Three-stage performance on Qwen3-8B.}
        \label{fig:qwen3_8b_2stage}
    \end{subfigure}
    \hfill
    \begin{subfigure}[b]{0.48\textwidth}
        \centering
        \includegraphics[width=\linewidth]{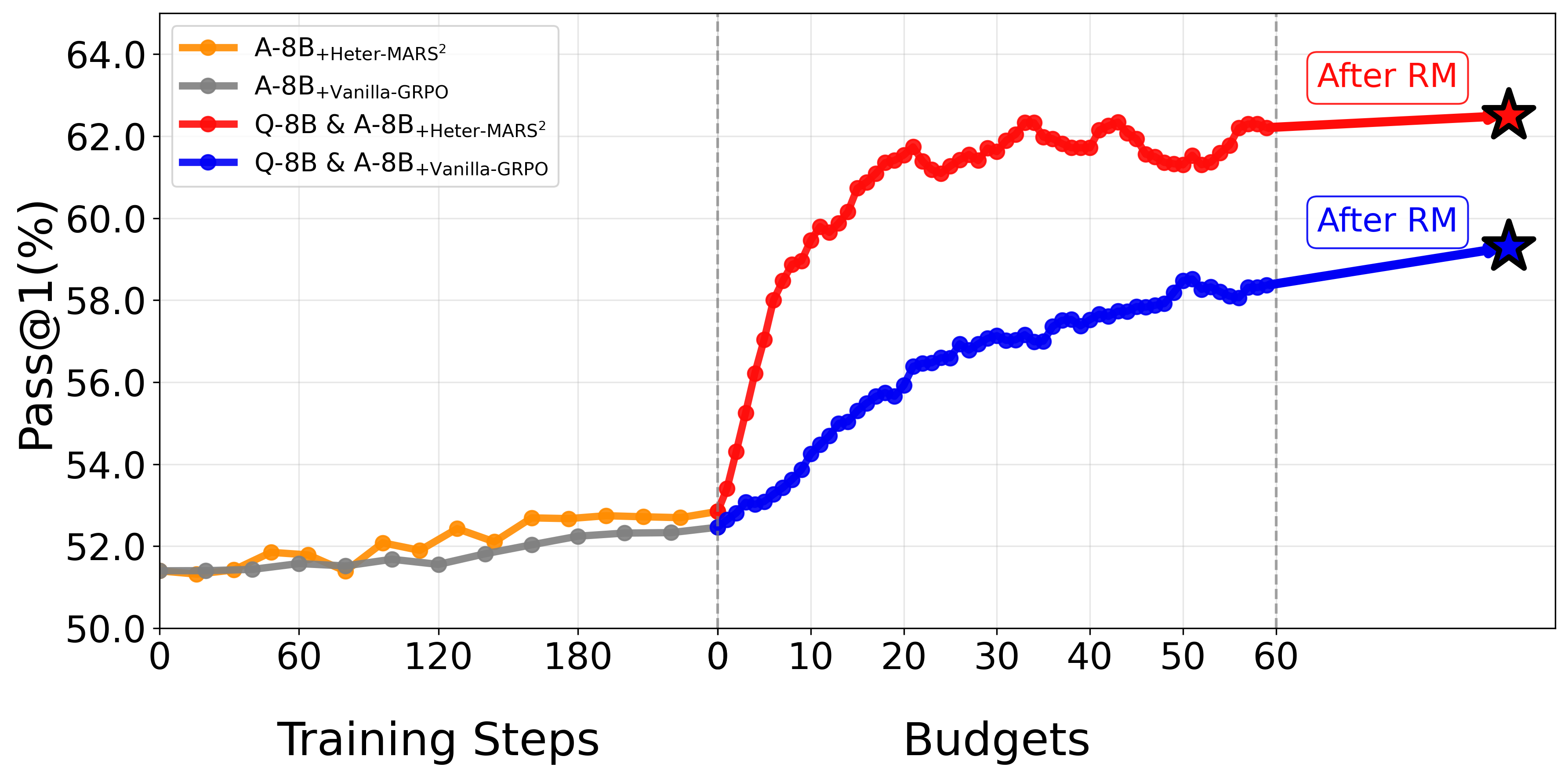}
        \subcaption{Three-stage performance on AReaL-boba-2-8B.}
        \label{fig:areal_8b_2stage}
    \end{subfigure}
    
    \vspace{1em}  % 增加行间距，可以调整数值如 0.5em, 1em, 2em 等
    % ---- Row 2 ----
    \begin{subfigure}[b]{0.48\textwidth}
        \centering
        \includegraphics[width=\linewidth]{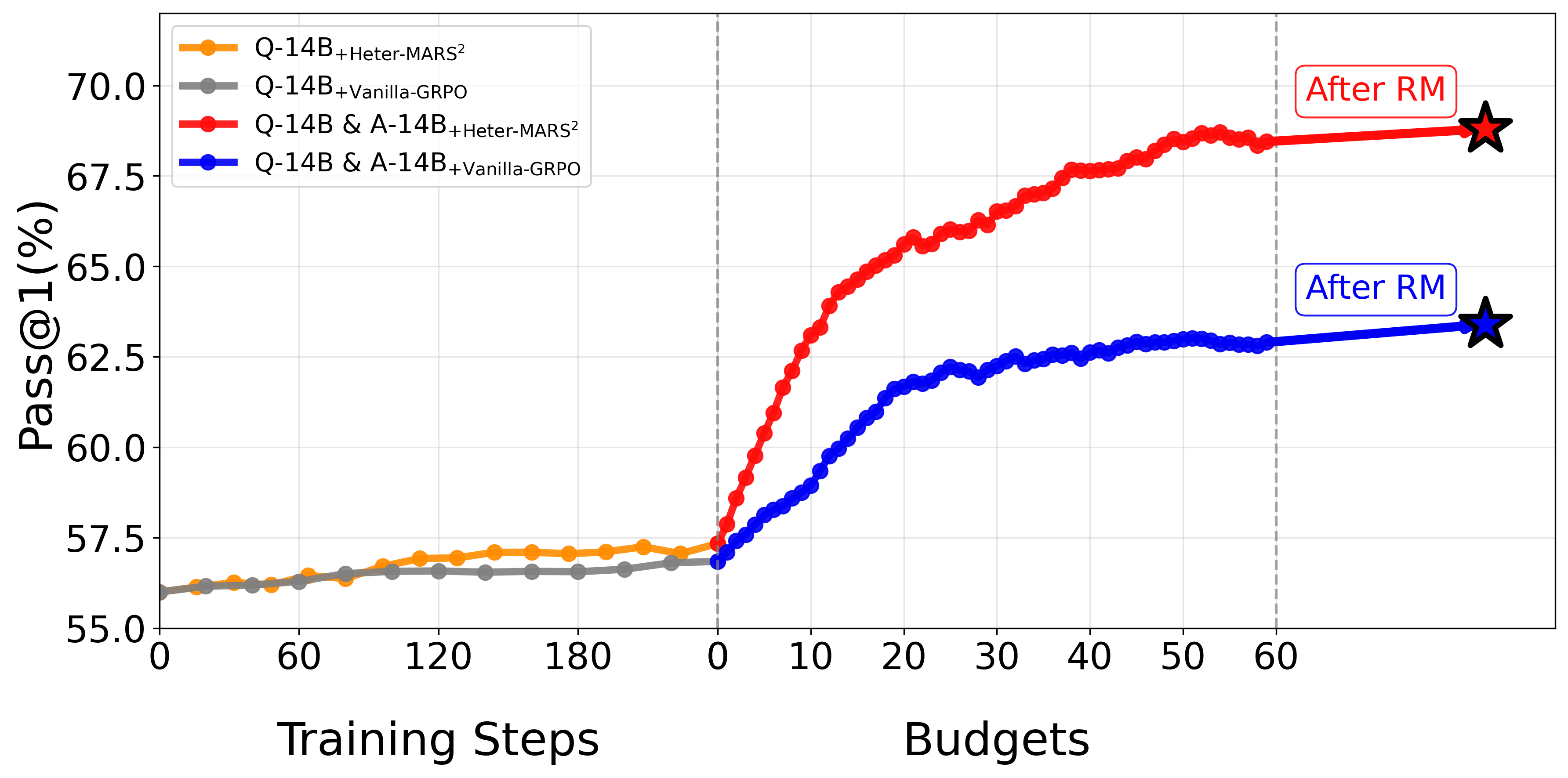}
        \subcaption{Three-stage performance on Qwen3-14B.}
        \label{fig:qwen_14b_2stage}
    \end{subfigure}
    \hfill
    \begin{subfigure}[b]{0.48\textwidth}
        \centering
        \includegraphics[width=\linewidth]{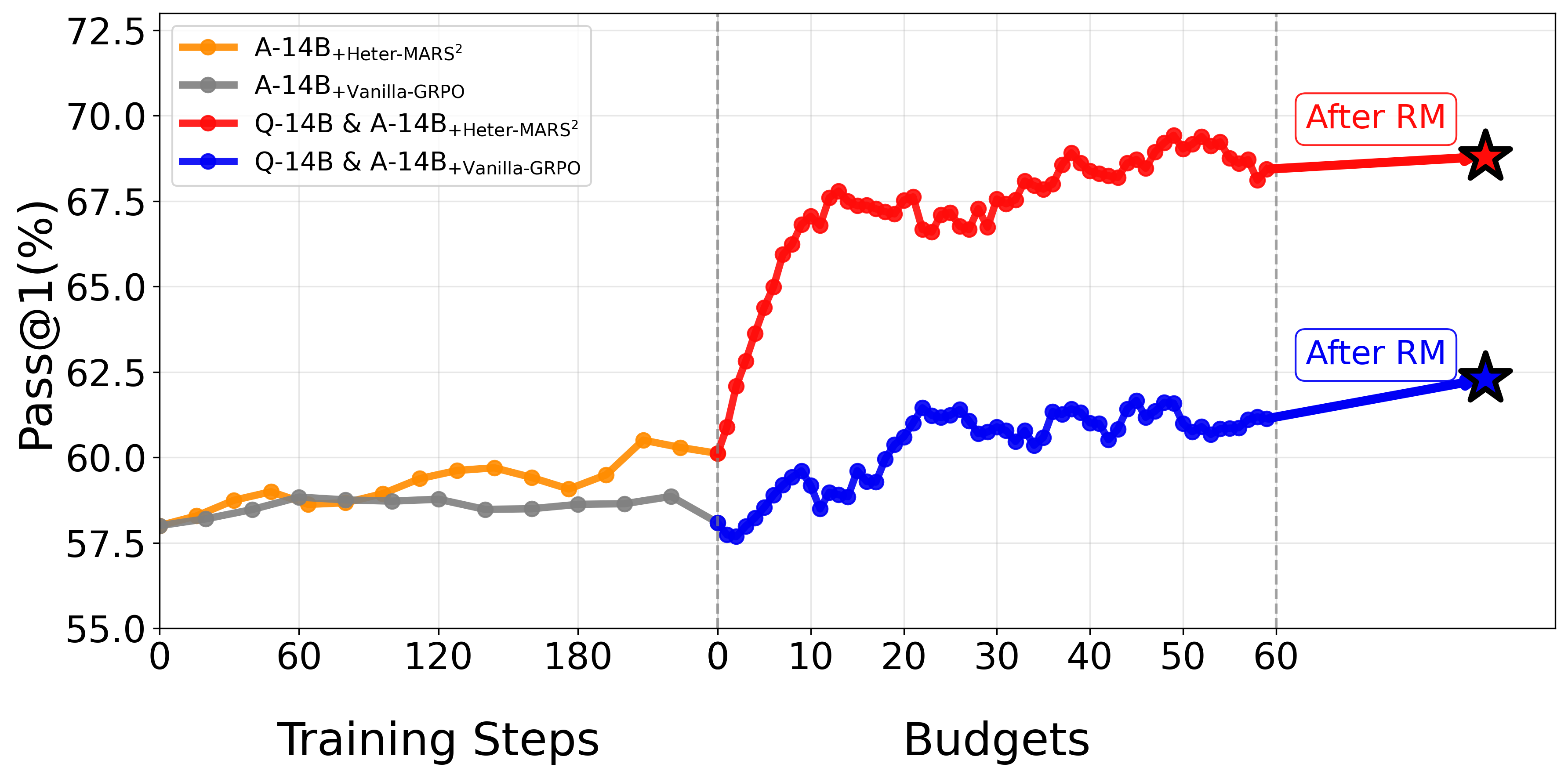}
        \subcaption{Three-stage performance on AReaL-boba-2-14B.}
        \label{fig:areal_14b_2stage}
    \end{subfigure}

    \vspace{1em}  % 增加行间距，可以调整数值如 0.5em, 1em, 2em 等
    % ---- Row 2 ----
    \begin{subfigure}[b]{0.48\textwidth}
        \centering
        \includegraphics[width=\linewidth]{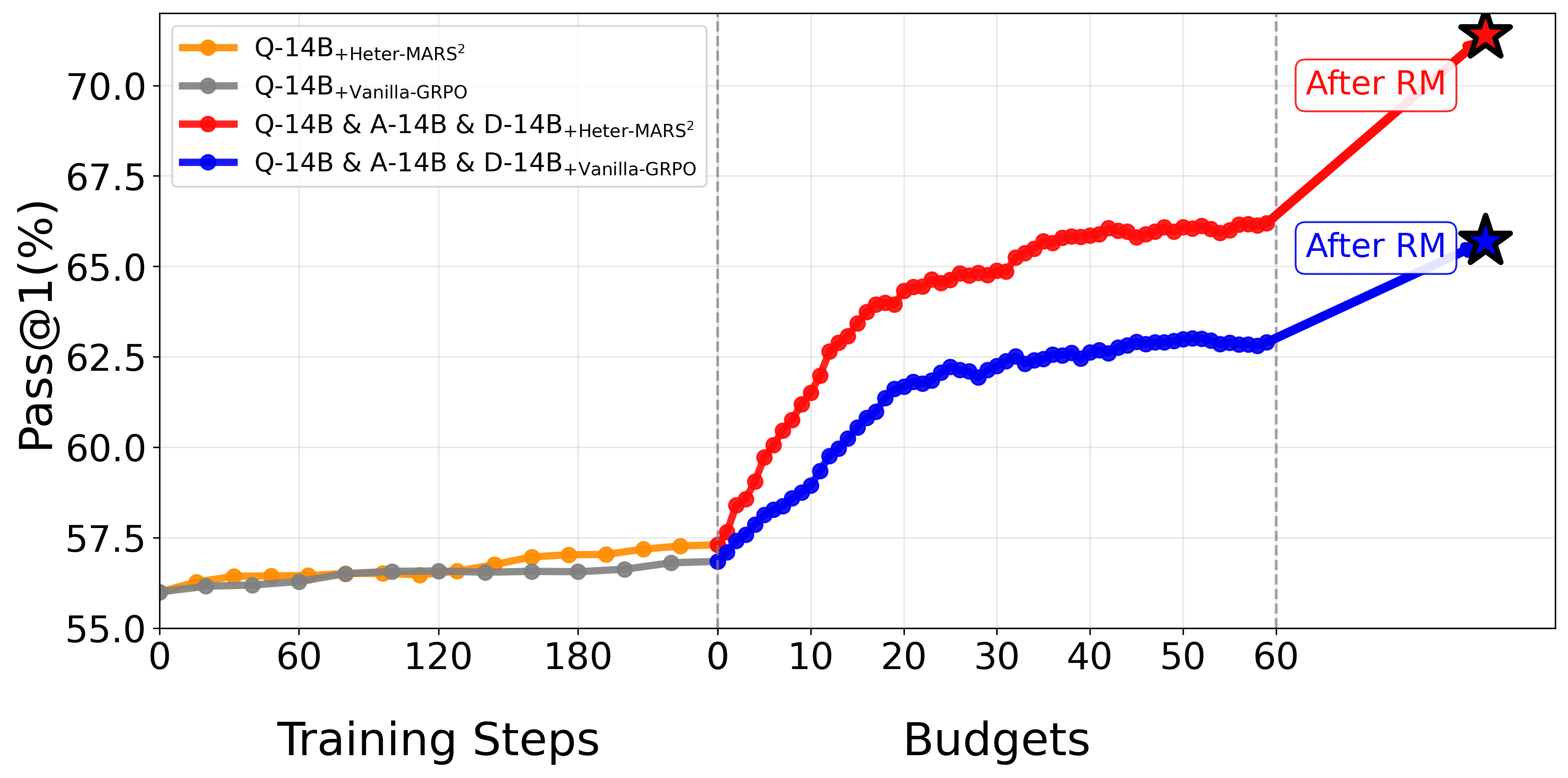}
        \subcaption{Three-stage performance on Qwen3-14B.}
        \label{fig:qwen_14b_2stage_3}
    \end{subfigure}
    \hfill
    \begin{subfigure}[b]{0.48\textwidth}
        \centering
        \includegraphics[width=\linewidth]{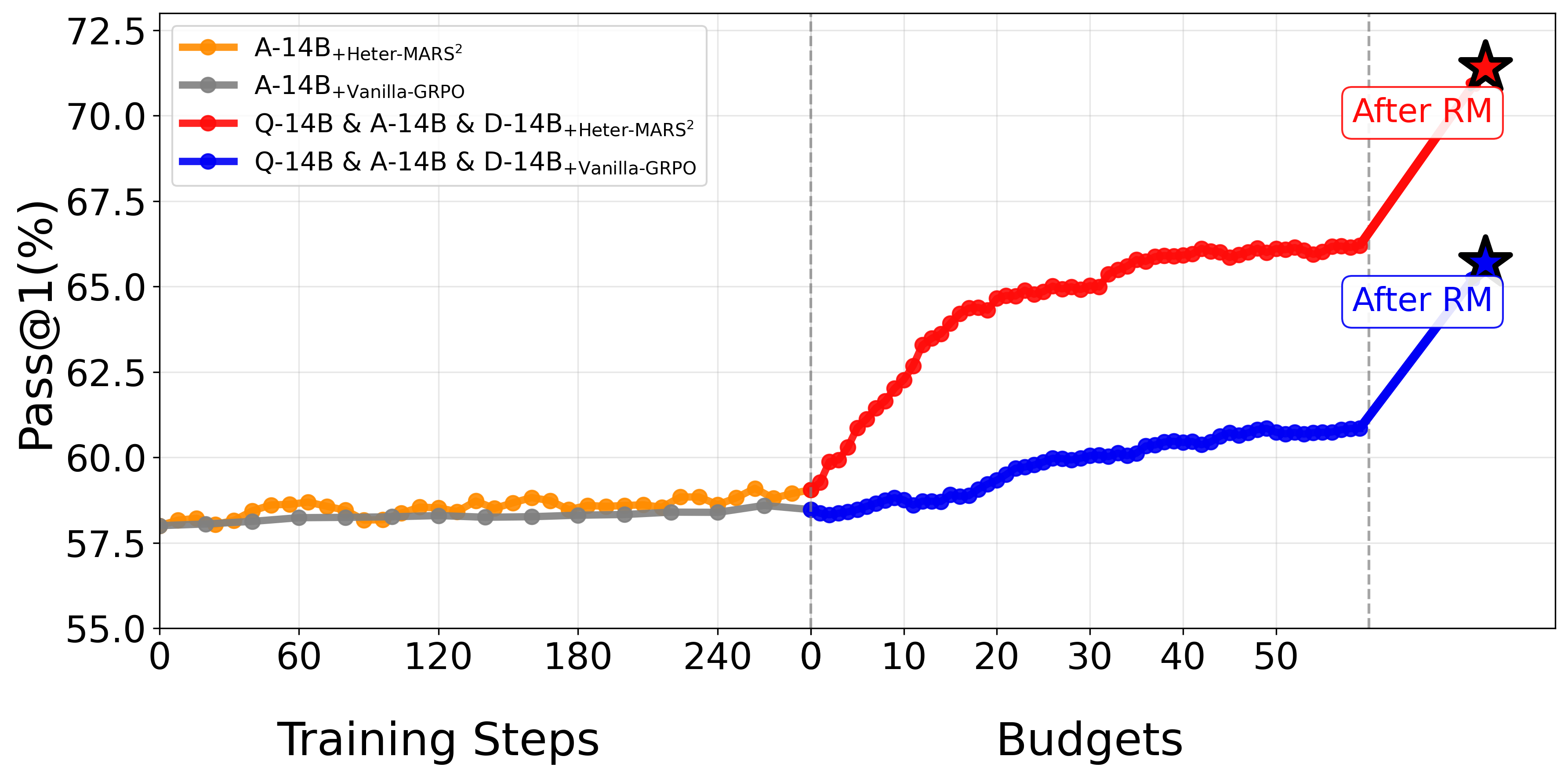}
        \subcaption{Three-stage performance on AReaL-boba-2-14B.}
        \label{fig:areal_14b_2stage_3}
    \end{subfigure}

    \caption{Three-stage scaling trend across different models and parameter sizes.}
    \label{fig:tts_2stage_all}
\end{figure}

\textbf{Multi-dimensional collaborative scaling law.} \ 
% Experiments with varying agent counts show that even when introducing a relatively weaker model—DeepCoder-14B-Preview Heter-MARS$^2$ still yields substantial improvements. The combination TA-14B improves from a pre-training \texttt{System-pass@1 (Node=60)} of 64.6\% to 71.4\% after Heter-MARS$^2$ training. As shown in Figure \ref{numagent} for Qwen3-14B, a clear scaling trend with the number of agents is observed. However, when AReaL-14B is combined with the weaker DeepCoder-14B, performance slightly decreases; this anomaly is likely due to the large disparity in agent capabilities, which may have hindered the strongest model.
% 为了进一步探究 Heter-MARS带来的能力扩展性，我们考察了其在不同阶段的性能表现，这揭示了一种超越传统Scaling Law的新维度。如图8所示，我们分别绘制了系统在后训练（Post-training）阶段和测试时（Test Time）阶段的性能曲线。具体而言，我们对比了 Heter-MARS和 GRPO 在不同模型组合（8B双模型、14B双模型、14B三模型）下的表现。
In order to further investigate the scalability enabled by Heter-MARS$^2$, we examined its performance at different stages, revealing a new dimension that transcends traditional scaling laws. As shown in Figure~\ref{fig:tts_2stage_all}, we plot the performance curves during both the post-training phase and test-time phase, which compare Heter-MARS$^2$ and GRPO across various model combinations. The results clearly demonstrate the scaling phenomenon along two orthogonal dimensions:
\begin{itemize}
% 从后训练到测试时的扩展（Post-training to Test-time Scaling）: 在后训练阶段，即不进行树搜索、仅依赖策略模型进行推理时，Heter-MARS训练的系统性能已经超越了 GRPO。这证明了通过多智能体交互与反馈，智能体自身的策略网络已经被有效优化，实现了第一层能力扩展。进入测试时阶段，通过 MCTS 进行协作式树搜索（TTS），系统的性能在后训练的基础上实现了第二次、也是更大幅度的飞跃。Heter-MARS训练的智能体不仅自身能力更强，而且更擅长在测试时通过协作探索和反馈机制，突破更高的性能上限。特别地，当我们训练的高质量奖励模型介入指导搜索时，这种性能增益被进一步放大，达到了系统的峰值表现。
    \item \textbf{Scaling from post-training to test-time:} During the post-training phase, where no tree search is conducted and inference relies solely on the policy model, the system trained by Heter-MARS$^2$ already outperforms GRPO. This indicates that, through multi-agent interaction and feedback, the policy networks of the agents have been effectively optimized. When entering the test phase, the system undergoes collaborative tree search via MCTS, whose performance exhibits a second and more significant leap beyond post-training results. Heter-MARS$^2$ not only possess stronger individual capabilities, but also excel at surpassing higher performance limits during testing through collaborative exploration and feedback mechanisms. Notably, when our high-quality reward model is involved to guide the search process, the performance boost is further amplified, reaching the peak performance of the system.
    % 从单智能体到多智能体协作的扩展（Single-agent to Multi-agent Scaling）: 图中显示，无论是随机选择节点还是利用我们训练的高质量奖励模型来指导节点选择，Heter-MARS系统的性能增益（从 Post-training 到 Test-time 的曲线斜率）均显著大于 GRPO。特别地，当使用奖励模型指导搜索时，性能达到了峰值，这凸显了高质量反馈在引导协作、挖掘系统集体智能潜力中的关键作用。
    % 无论是后训练阶段的策略能力，还是测试时阶段的协作能力，实验结果明确显示，由多个智能体组成的系统其性能上限远高于任何单个基线模型。这意味着，通过 Heter-MARS的异构协作框架，我们成功实现了 1+1 > 2 的效果。智能体间的互补性知识和交叉验证机制，打破了单个模型的能力瓶颈，使得整个系统能够解决更复杂的问题。这种从单智能体到多智能体的扩展，为提升AI系统能力开辟了一条与单纯增加模型参数量（Scaling Up Model Size）并行且互补的新路径。
    \item \textbf{Scaling from single-agent to multi-agent:}  Whether in terms of policy capability at the post-training stage or collaborative ability at the test-time stage, the experimental results clearly demonstrate that the performance ceiling of multi-agents is significantly higher than that of any single-agent. The complementary knowledge and cross-validation mechanisms among heterogeneous agents break the capability bottleneck of individual models, enabling the entire system to solve more complex problems. This expansion from single-agent to multi-agent systems paves a parallel and complementary path for enhancing LLMs capabilities, alongside simply scaling up model size.
\end{itemize}

\subsection{Results of MARS$^2$-T+}
% ReMASS-T+进一步提升了智能体系统的上限
% 经过ReMASS训练的智能体在协作交互能力方面获得了明显的增强
% 探索训练前所使用TTS方法的潜力和局限性，潜力意味着有助于扩大探索空间，局限性往没有训练过导致的效果不佳/受限的角度上引导
% TTS阶段的实验，我们首先详细分析了多智能体AB-MCTS的性能上限和局限性，在此基础上我们验证了所提出的ReMASS-T+的有效性。在经过ReMASS训练后模型上，我们进一步使用ReMASS-T+方式进行TTS扩展，获得了更高的上限
% 
% In the TTS-stage experiments, we first systematically analyze the performance upper bound and limitations of the Vanilla multi-agent TTS. Building on this analysis, we validate the effectiveness of the proposed MARS$^2$-T+ method. Subsequently, we further apply MARS$^2$-T+ to models trained with MARS$^2$ for TTS scaling, achieving a higher performance upper bound.

\paragraph{The Limitations of Vanilla-TTS.} 

\begin{figure}
    \centering
    \includegraphics[width=0.8\linewidth]{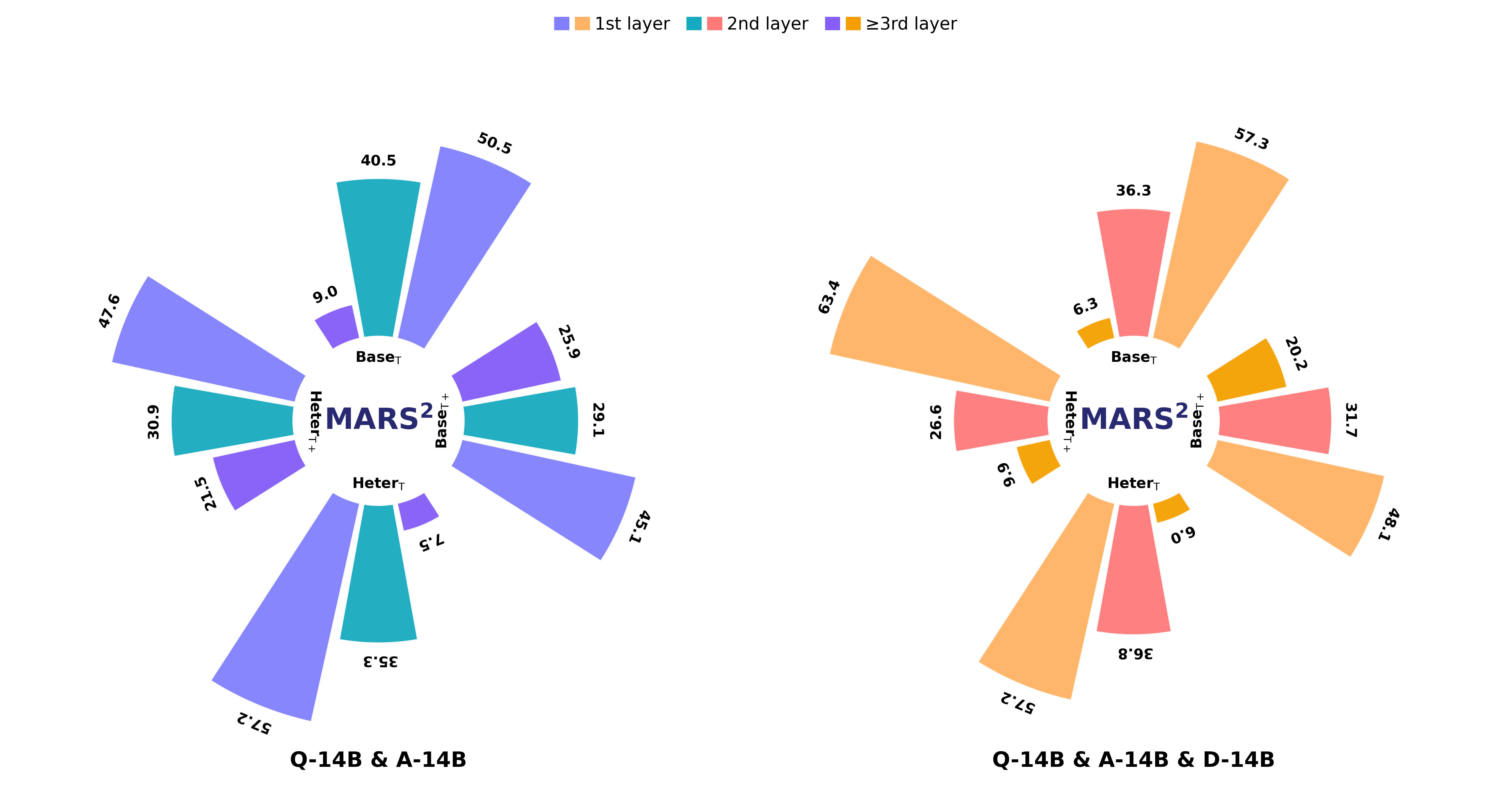}
    \caption{
    % qwen3-14b & areal 14B & deepcoder-14b训练前后节点深度分布
    Node depth distributions of Qwen3-14B \& AReaL-14B \& DeepCoder-14B before and after training}
    \label{fig:placeholder}
\end{figure}

\begin{figure*}[t]
    \centering  

    %------------------ First Row ------------------%
    \begin{subfigure}[b]{0.46\textwidth}  
        \centering
        \includegraphics[width=\linewidth]{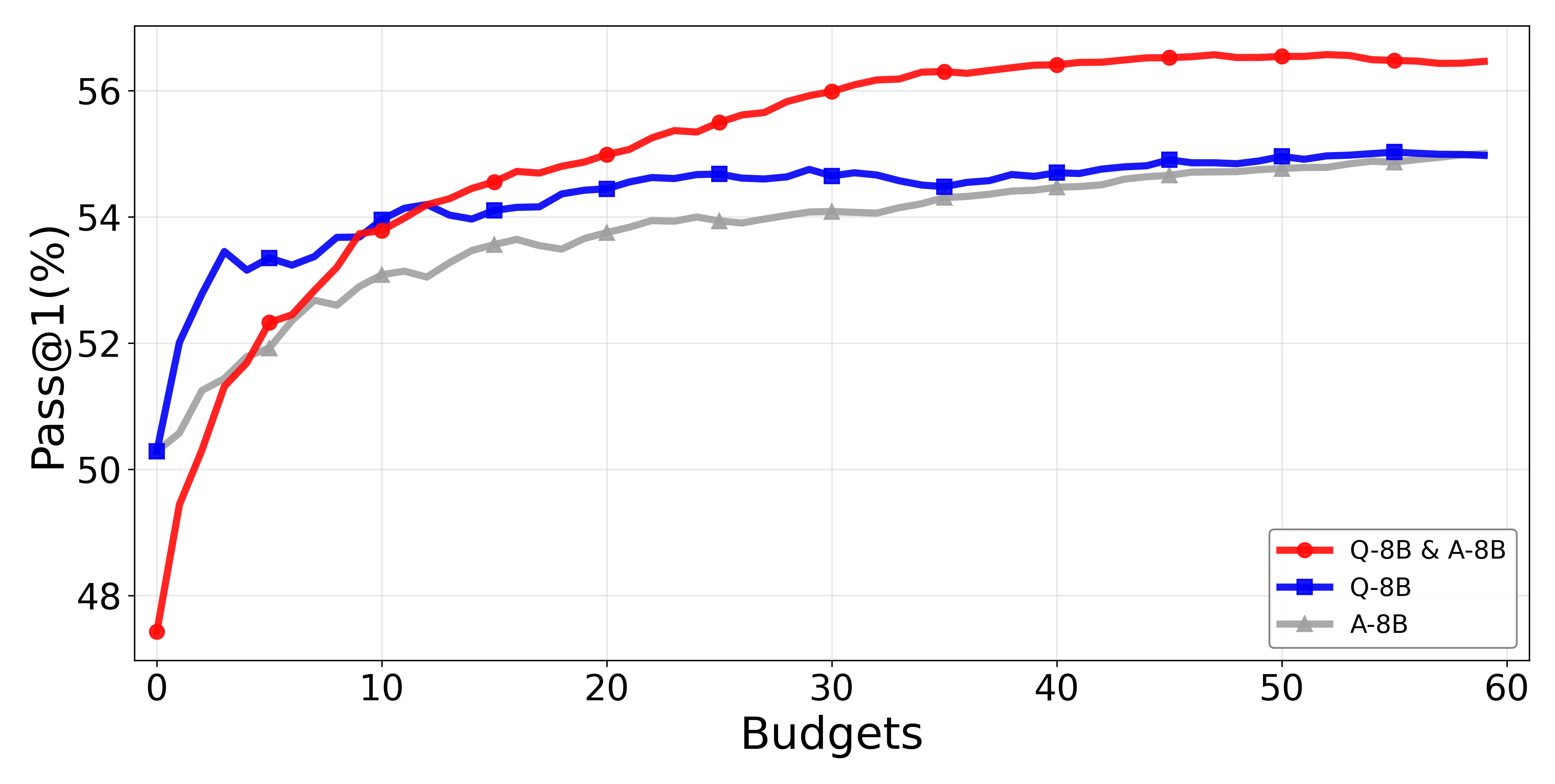}  
        \subcaption{8B Pass@1(MCTS)(Base)}  
        \label{fig:base-8b-pass1}
    \end{subfigure} 
    \hfill
    \begin{subfigure}[b]{0.46\textwidth}
        \centering
        \includegraphics[width=\linewidth]{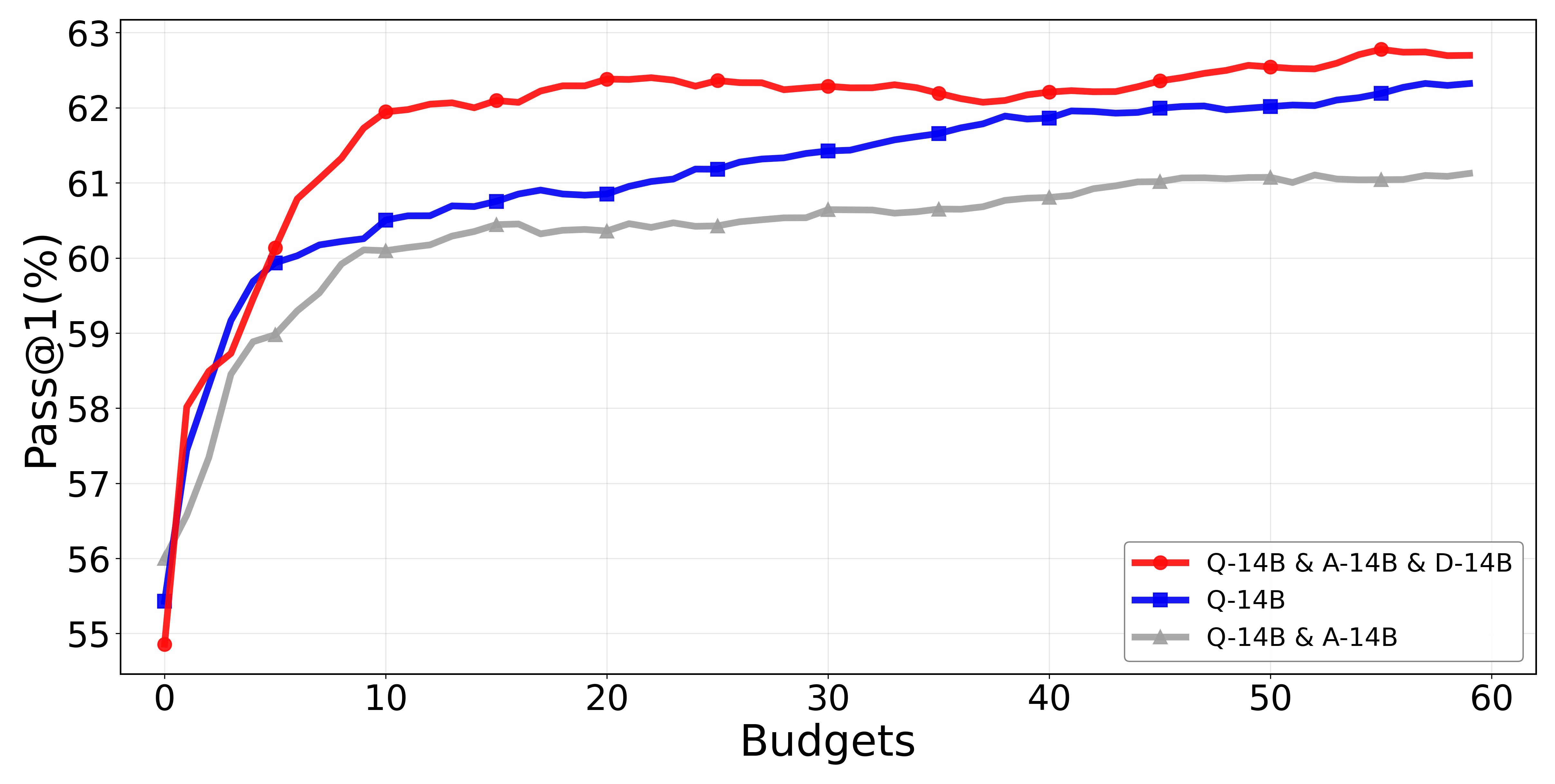}  
        \subcaption{14B Pass@1(MCTS)(Base)}
        \label{fig:base-14b-pass1}
    \end{subfigure}

    %------------------ Second Row ------------------%
    \vspace{0.5em}

    \begin{subfigure}[b]{0.46\textwidth}  
        \centering
        \includegraphics[width=\linewidth]{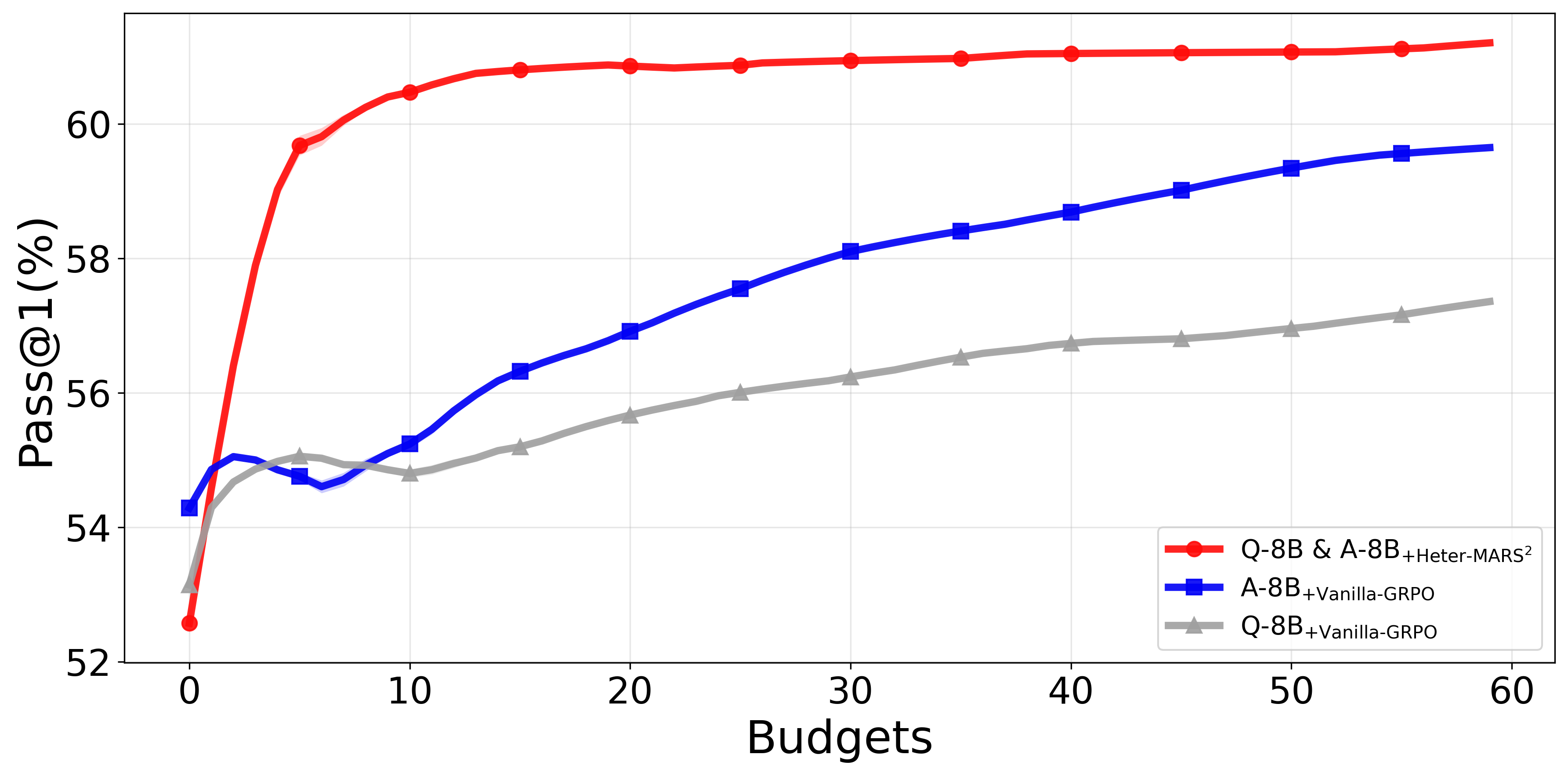}  
        \subcaption{8B Pass@1(MCTS)(MARS$^2$)}  
        \label{fig:trained-8b-pass1}
    \end{subfigure}
    \hfill  
    \begin{subfigure}[b]{0.46\textwidth}
        \centering
        \includegraphics[width=\linewidth]{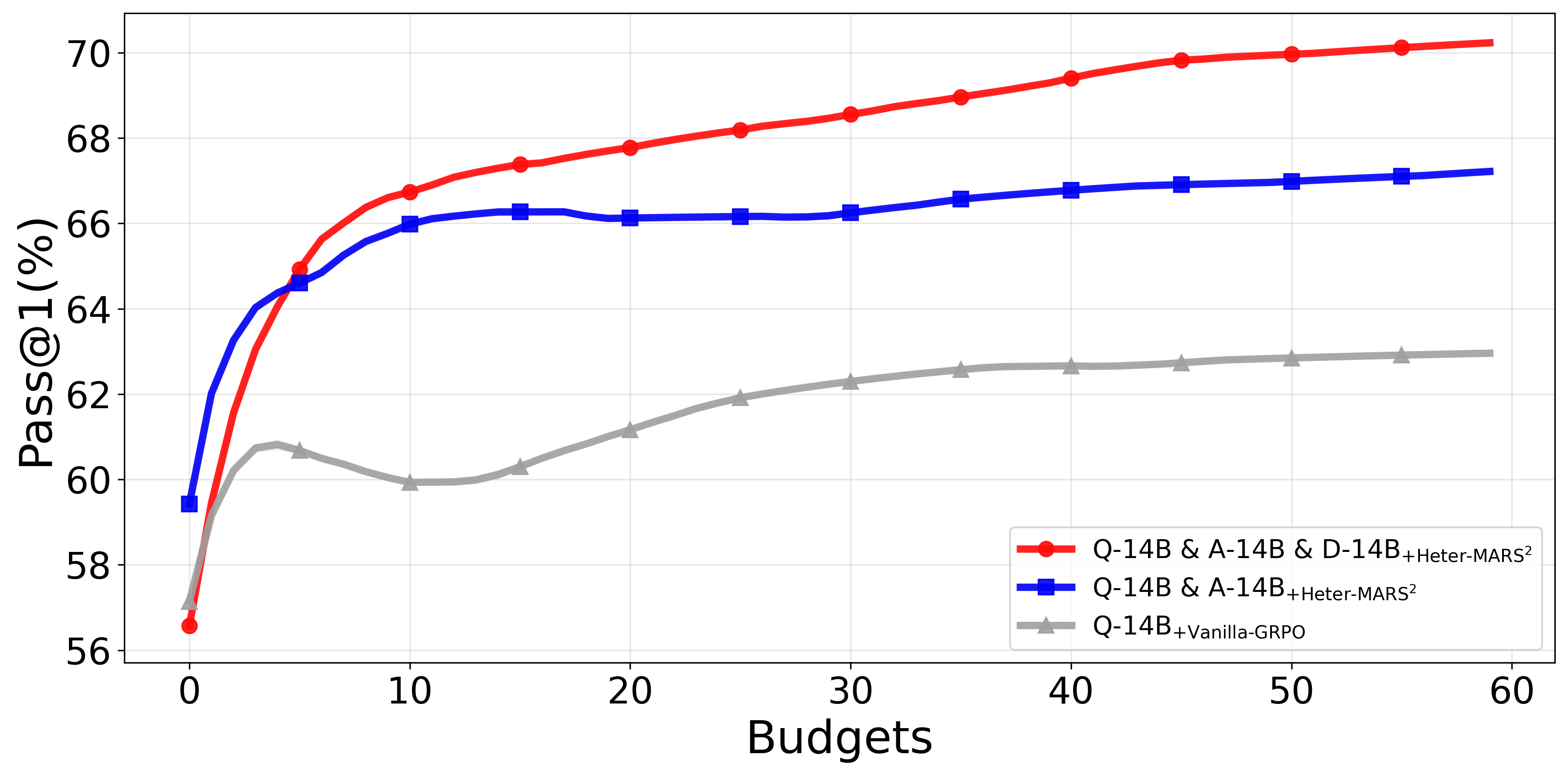}  
        \subcaption{14B Pass@1(MCTS)(MARS$^2$)}
        \label{fig:trained-14b-pass1}
    \end{subfigure}

    \caption{System-level performance before (top) and after (bottom) training with MARS$^2$.}
    \label{fig:tts-base train} 
\end{figure*}

\begin{figure}[t]
    \centering  

    \begin{subfigure}[b]{0.48\textwidth}  
        \centering
        \includegraphics[width=\linewidth]{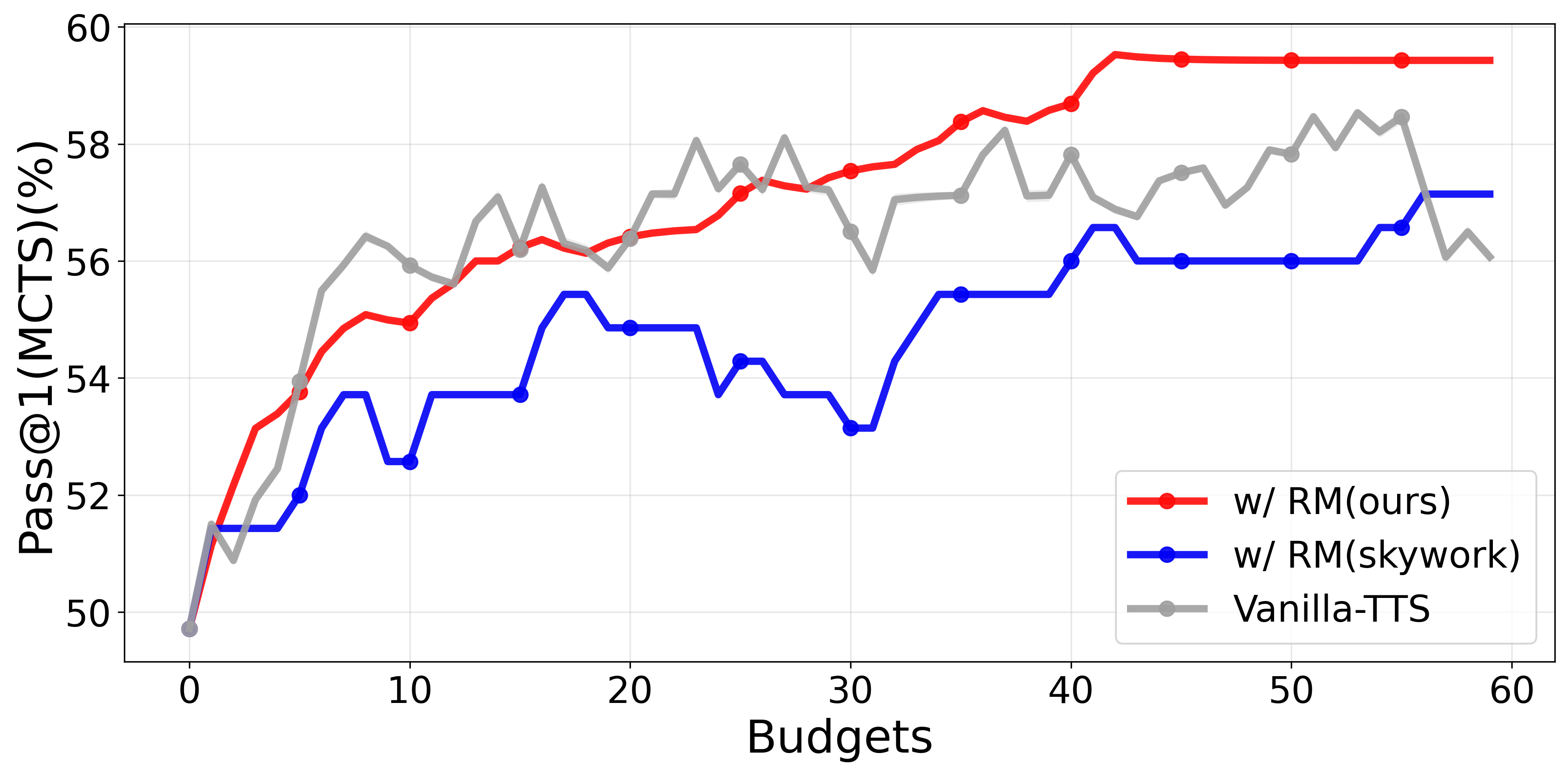}  
        \subcaption{8B Pass@1(MCTS)}  
        \label{fig:qwen3_14b_123.png}  
    \end{subfigure}
    \hfill  
    \begin{subfigure}[b]{0.48\textwidth}
        \centering
        \includegraphics[width=\linewidth]{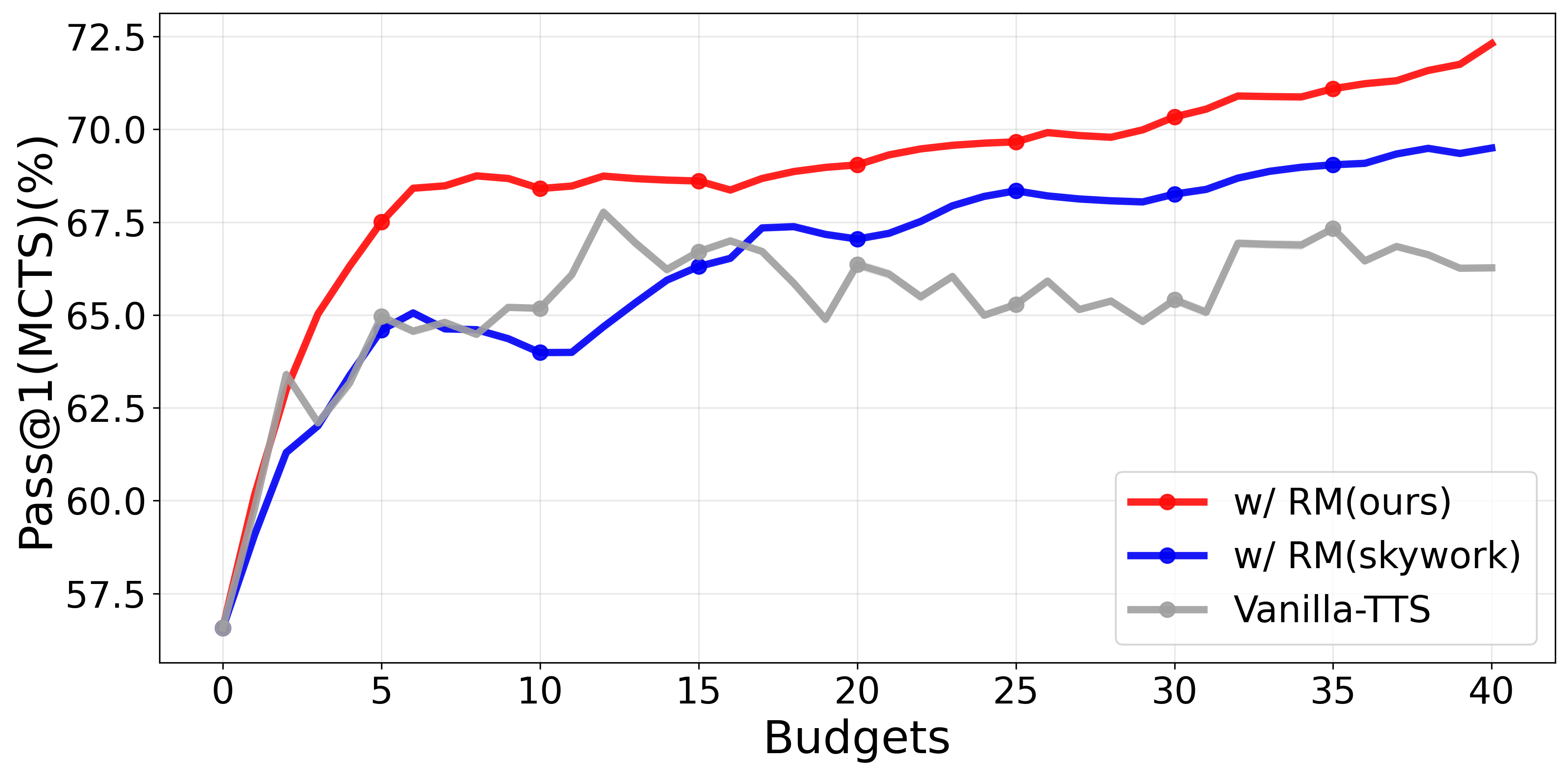} 
        \subcaption{14B Pass@1(MCTS)}
        \label{fig:areal_14b_123.png}
    \end{subfigure}

    \caption{Ablation experiments of reward model.}
    \label{fig:rmablation}
\end{figure}

We begin by systematically evaluating the performance of Vanilla multi-agent tree search (Vanilla TTS) across a variety of models and model combinations. The results show that while Vanilla TTS provides modest system-level gains under small inference budgets, its improvements quickly saturate as the budget increases, revealing a clear upper bound (Figure~\ref{fig:tts-base train} (Base)). A deeper analysis identifies three key limitations:
\begin{itemize}
    \item \textbf{Insufficient exploitation of error signals.} As Appendix~\ref{box:ab-mcts-prompt} shows, Vanilla TTS supplies only binary "pass/fail" feedback from parent to child nodes, discarding richer diagnostic information such as input–output mismatches or runtime error logs. This coarse feedback is insufficient for guiding effective refinement in complex code-generation tasks.
    \item \textbf{Limited depth for refinement.} As shown in Figure~\ref{fig:placeholder}, over 90\% of nodes in the 14B model group remain within the first three levels of the search tree, with fewer than 7\% reaching level four or deeper. This strong shallow bias becomes even more pronounced in multi-agent settings; even after the model is strengthened through MARS$^2$ training, the proportion of deeper nodes under Vanilla TTS remains very low, leaving substantial multi-agent collaboration potential underutilized.
    \item \textbf{Unstable candidate selection.} Vanilla TTS selects the last node that passes the public test cases as the final output. Because public test cases are sparse and cover only a narrow portion of the task space, this selection rule tends to favor solutions overfitted to visible examples. As illustrated in Figure~\ref{fig:rmablation}, Pass@1(MCTS) exhibits substantial variance under increasing budgets and fails to deliver consistent improvements even when the compute budget is doubled.
\end{itemize}

\begin{tcolorbox}[takeawaysbox]
% (1) \textbf{MARS$^2$-T+ Raises the System’s Performance Ceiling.} MARS$^2$-T+ delivers additional gains, further extending the capability upper bound of multi-agent systems.

% (2) \textbf{MARS$^2$ Strengthens Agents’ Cooperative Reasoning.} MARS$^2$ training markedly improves agents’ collaborative interaction, enabling deeper search and larger performance gains.

(1) \textbf{MARS$^2$-T+ fundamentally breaks the ceiling imposed by Vanilla TTS.}
% MARS$^2$-T+ elevates test-time search from shallow probing to a feedback-enhanced, depth-guided, and evaluation-driven reasoning process, structurally overcoming the limitations of Vanilla TTS. By integrating corrective feedback and reward-based evaluation, T+ enables reasoning capabilities that surpass the previous performance ceiling.
MARS$^2$-T+ upgrades test-time search into a feedback-enhanced, depth-guided, and evaluation-driven reasoning process, overcoming the core structural limits of Vanilla TTS and enabling performance beyond the previous ceiling.

(1) \textbf{MARS$^2$ training and MARS$^2$-T+ form a mutually reinforcing reasoning ecosystem.}
% MARS$^2$ training establishes structured collaboration among agents, while MARS$^2$-T+ amplifies and deepens this coordination during inference, yielding pronounced non-linear gains. The scaling trends in both multi-agent and heterogeneous-agent settings indicate that collaborative reasoning is genuinely activated, no longer constrained by model size or sampling depth.
MARS$^2$ training establishes structured agent collaboration, and MARS$^2$-T+ amplifies and deepens this coordination during inference; by jointly enhancing collaborative reasoning, they produce pronounced non-linear system-level gains.

(3) \textbf{The components of MARS$^2$-T+ constitute a co-evolving triad rather than isolated techniques.}
% Error feedback improves correction quality, depth guidance drives deeper exploration, and the reward model stabilizes final selection. These components form a coordinated chain—from information augmentation $\rightarrow$ reasoning guidance $\rightarrow$ decision optimization—where each is indispensable. Their integration delivers stable, sustained gains, and MARS$^2$-T+ achieves its maximal benefit only when all three mechanisms operate jointly.
Error feedback improves correction, depth guidance drives exploration, and the reward model stabilizes decisions. Their coordinated interaction—from information augmentation $\rightarrow$ reasoning guidance $\rightarrow$ decision optimization—is indispensable, and MARS$^2$-T+ reaches maximal benefit only when all three operate together.

\end{tcolorbox}

\textbf{Effectiveness of MARS$^2$-T+.} \
% 针对上述问题，我们提出了MARS$^2$对其进行改进，为了验证我们改进的有效性，我们进行了全面的评估。
To address these limitations, we introduce MARS$^2$ and conduct a comprehensive set of experiments to evaluate the effectiveness of our improvements. Across all 8B-scale model configurations, MARS$^2$-T+ consistently outperforms Vanilla TTS. For instance, with Qwen3-8B (Table~\ref{tab:Tplusresult}), replacing Vanilla TTS with MARS$^2$-T+—while keeping model parameters unchanged—raises Pass@1(MCTS) from 54.3\% to 57.7\% and PASS@N from 68.6\% to 71.4\%. This indicates that, under the same compute budget, MARS$^2$-T+ not only increases the likelihood of producing correct solutions but also yields more stable and higher-quality candidate sets. Unlike Vanilla TTS, which relies solely on sparse public test cases and therefore provides limited guidance, MARS$^2$-T+ more effectively leverages intermediate refinement signals and final-stage evaluations, directing computational resources toward more promising search branches and ultimately improving budget utilization efficiency.

% 训练和T+的协同作用
\textbf{Synergistic Effects Between MARS$^2$ Training and MARS$^2$-T+ Inference Enhancement.} \
We further examine how MARS$^2$ training interacts with T+ inference by comparing multi-agent configurations before and after training. The results show that MARS$^2$-T+ brings only modest gains on Base multi-model systems—for example, Qwen3-8B + AReaL-8B improves only marginally under MARS$^2$-T+. In contrast, once the same model pair is trained with MARS$^2$, MARS$^2$-T+ unlocks substantially higher performance, pushing the Pass@1(MCTS) score to 62.9\%, a level unattainable by any Base multi-agent configuration even with increased inference budgets(Table \ref{tab:Tplusresult}). This contrast highlights that MARS$^2$ training reshapes the collaborative policy landscape across agents, enabling MARS$^2$-T+ to exploit deeper refinement opportunities and achieve performance ceilings that Vanilla TTS and Base model cannot reach.
From the perspective of scalability, the multi-agent TTS behavior before and after training diverges even more clearly: in the Base setting, heterogeneous model combinations do not consistently outperform homogeneous shared-parameter ones, and in some 14B configurations they even underperform (Figure \ref{fig:base-14b-pass1}), suggesting that Vanilla TTS fails to convert policy heterogeneity into effective cooperative gains. After MARS$^2$ training, however, all model combinations exhibit a pronounced multi-agent scaling trend—performance steadily increases and becomes more stable as the number or heterogeneity of participating agents grows. This shift demonstrates that MARS$^2$ training not only improves single-sample quality but also reshapes collaborative strategy structures, enabling MARS$^2$-T+ to operate within a more compatible policy space and realize deeper reasoning expansion. Together, the two stages form a mutually reinforcing capability loop, raising both the performance ceiling and the scalability of LLMs on complex code-generation tasks.

% \begin{table}[H]
%     \caption{Comparison of MARS$^2$-T+ before and after training.}
%     \label{tab:Tplusresult}
%     \centering
%     \small
%     \setlength{\tabcolsep}{4pt}
%     \begin{tabular}{cccc}%四个c代表该表一共四列，内容全部居中
%     \toprule%第一道横线
%     \textbf{Methods} & \textbf{PASS@1} & \textbf{BoN} & \textbf{PASS@N} \\
%     \midrule% 
%     Q-8B$_{\text{+Base}}$ & 50.3 & 54.3 & 68.6 \\
%     Q-8B$_{\text{+Base-T+}}$ & 50.3 & 57.7 & 71.4 \\
%     Q-8B \& A-8B$_{+\text{Base}}$ & 50.3/51.0 & 57.2 & 69.7 \\
%     Q-8B \& A-8B$_{+\text{Base-T+}}$ & 50.3/51.0 & 58.9 & 71.4 \\
%     Q-8B \& A-8B$_{+\text{Homo-MARS$^2$}}$ & 55.4/55.4 & 57.2 & 72.6 \\
%     Q-8B \& A-8B$_{+\text{Homo-MARS$^2$-T+}}$ & 55.4/55.4 & 60.0 & 69.1 \\
%     Q-8B \& A-8B$_{+\text{Heter-MARS$^2$}}$ & 58.3/54.9 & 61.7 & 75.4 \\
%     Q-8B \& A-8B$_{+\text{Heter-MARS$^2$-T+}}$ & 58.3/54.9 & 62.9 & 72.0 \\
%     \bottomrule
%     \end{tabular}
% \end{table}

\begin{table}[ht]
    \caption{Performance comparison between Vanilla TTS and MARS$^2$-T+(w/ RM
    and w/ all).}
    \label{tab:Tplusresult}
    \centering
    \small
    \setlength{\tabcolsep}{5pt}
    \begin{tabular}{lcccccc}
    \toprule
    \multirow{2}{*}{\textbf{Methods}}  
        & \multicolumn{3}{c}{\textbf{Pass@1(MCTS)}} 
        & \multicolumn{3}{c}{\textbf{PASS@N}} \\
    \cmidrule(lr){2-4} \cmidrule(lr){5-7}
        & Vanilla TTS & w/ RM & w/ all & Vanilla TTS & w/ RM & w/ all \\
    \midrule
    % Q-8B (Base)                 & 54.3 & & 57.7 & 68.6 & & 71.4 \\
    Q-8B \& A-8B (Base)         & 57.2 & 57.1 & 58.9 & 69.7 & 69.7 & 71.4 \\
    Q-8B \& A-8B (Homo-MARS$^2$)   & 57.2 & 59.4 & 60.0 & 72.6 & 72.6 & 69.1 \\
    Q-8B \& A-8B (Heter-MARS$^2$) & 61.7 & 62.3 & 62.9 & 75.4 & 75.4 & 72.0 \\
    \hline
    Q-14B \& A-14B (Base)         & 62.9 & 63.4 & 65.1 & 74.9 & 74.9 & 78.9 \\
    Q-14B \& A-14B (Heter-MARS$^2$) & 68.9 & 68.6 & 70.3 & 79.4 & 79.4 & 80.6 \\
    \bottomrule
    \end{tabular}
\end{table}

% rm的必要性/消融
\textbf{Ablation on Reward Models and Their Impact on TTS Stability.} \
As discussed earlier, rollouts during training can access the full set of test cases, whereas inference is restricted to public test cases, creating a clear mismatch in reward signals between the two stages. To quantify the impact of this mismatch on TTS, we conduct a series of Reward Model (RM) ablation studies (Table~\ref{tab:rmbenchmark}). In the 14B multi-agent setting, the performance of Vanilla TTS (gray) fluctuates substantially as the inference budget increases, indicating that supervision derived solely from public test cases is too sparse to provide stable guidance. Introducing a generic Skywork-Reward model (blue) reduces volatility but yields only limited performance gains. In contrast, the RM fine-tuned on training trajectories (orange) produces markedly smoother curves and consistently outperforms the other variants across all budget levels.
A similar trend is observed in the 8B multi-agent configuration: although the fine-tuned RM does not always surpass Vanilla TTS under low budgets, it significantly improves output stability and gradually achieves superior performance as the budget grows. Discriminative metrics further confirm this effect: AUC-ROC improves from 73.0\% to 79.3\%, and Spearman correlation increases from 39.7 to 50.6. Together with the TTS results, these findings show that a task-aligned, high-quality RM is essential for stabilizing multi-agent TTS and mitigating the reward discrepancy between training and inference, serving as a necessary component rather than an optional addition.

\begin{table}[H]
    \centering
    \caption{Performance evaluation of Reward Models. Comparison of base vs. fine-tuned versions across Pre-train and Post-train benchmarks using Adaptive Accuracy, AUC-ROC, and Spearman correlation.}
    \label{tab:rmbenchmark}
    \vspace{0.2cm} % Slightly more space between caption and table
    \small
    \setlength{\tabcolsep}{4pt}
    \begin{tabular}{lccccccc}
    \toprule
    \multirow{2}{*}{\textbf{Model}} & \multirow{2}{*}{\textbf{Loss}} & \multicolumn{3}{c}{\textbf{Pre-train (150 nodes)}} & \multicolumn{3}{c}{\textbf{Post-train (60 nodes)}} \\
    \cmidrule(lr){3-5} \cmidrule(lr){6-8}
     & & \textbf{Adapt. Acc} & \textbf{AUC-ROC} & \textbf{Spearman} & \textbf{Adapt. Acc} & \textbf{AUC-ROC} & \textbf{Spearman} \\
    \midrule
    Skywork-Reward (Base) & - & 65.5 & 71.2 & 36.7 & 66.3 & 73.0 & 39.7 \\
    \midrule
    Ours (BT Variant) & BT & 62.1 & 67.4 & 30.1 & 61.8 & 66.7 & 28.9  \\
    \textbf{Ours (MSE Variant)} & \textbf{MSE} & \textbf{69.9} & \textbf{78.7} & \textbf{49.7} & \textbf{71.6} & \textbf{79.3} & \textbf{50.6} \\
    \bottomrule
    \end{tabular}
\end{table}

% 反馈增强和深度引导的必要性
\textbf{Necessity of Refinement and Depth-Aware Guidance in Multi-Agent TTS.} \
To evaluate the necessity of refinement-enhanced feedback and depth guidance in MARS$^2$-T+, we analyze both node-depth distribution and final reasoning performance. As shown in Figure~\ref{fig:placeholder}, Vanilla TTS exhibits a severe shallow-search bias in the 14B model series: 93.7\% of nodes appear within the first two layers, while nodes at depth($\geq 4$) account for fewer than 7\%—a pattern that becomes even more pronounced in heterogeneous settings. After introducing refinement and depth guidance, the proportion of deeper nodes increases markedly (e.g., rising from 6.3\% to 20.2\%), indicating that the mechanism effectively reallocates computation toward more promising deep trajectories. The performance results further corroborate the necessity of refinement-enhanced feedback and depth guidance within the search process. When only an RM is added, candidate quality may be insufficient for reliable ranking, occasionally causing slight degradation. For example, 0.1\% Pass@1(MCTS) drop in the Qwen3-8B \& AReaL-8B Base configuration and 0.3\% decrease in the trained Qwen3-14B \& AReaL-14B (Heter-MARS$^2$) setting(Table \ref{tab:Tplusresult}). In contrast, once refinement-enhanced feedback and depth guidance are incorporated, these regressions disappear and are replaced by consistent gains across model scales. Collectively, these findings demonstrate that refinement and depth guidance are essential for multi-agent tree search: they correct shallow-search bias, strengthen deep exploration, and enable the RM to function reliably, thereby substantially stabilizing and improving TTS performance.

%% file: sections/5_casestudy.tex
\section{Diversity Analysis of MARS$^2$}\label{case_study}

% \begin{tcolorbox}[takeawaysbox]
% (1) Low-cost AR to SDAR adaptation works for any modern AR base, delivering \textbf{on-par performance with broad applicability.}

% (2) Larger SDAR models tolerate bigger blocks and looser decoding thresholds, enabling \textbf{higher parallel efficiency without sacrificing performance.}

% (3) In SDAR, larger models and higher confidence not only improve quality but also drive faster decoding, \textbf{making accuracy the engine of efficiency.}

% \end{tcolorbox}
%% 咱们的report主要claim的结论是，多体训练能提升单体的上限
%% 分析不能局限在TTS上，TTS只是很小的一个补充
%% 多体训练，比单体训练，的优势
%% 也需要对比一下grpo
% 梳理需要放的实验结果
% best of N
% Sequential Refinement
% abmcts 1,2,3 model
% abmcts_error_bias 1,2,3 model
% 训练后的单体
% 第1节介绍使用的多样性指标
% 第2节介绍推理中的各项指标结果
% 第3节介绍训练后的各项指标结果
\subsection{Diversity Metrics} 
%为了全面评估我们方法生成的代码的算法和认知多样性，我们设计了一个多级、互补的评估框架，该框架涵盖了代码层面的语义分布，算法差异以及推理过程中的认知策略多样性。我们使用pass@K来衡量方法的探索正确解的能力。Average Embedding Clusters (AEC)衡量代码之间的语义多样性。Distinct Algorithms at K (DA@K), Effective number of Algorithms (EA), and Normalized Area Under the Diversity Curve (NAUDC)\cite{lee-etal-2025-diversely}直接量化算法本身的多样性；探索性地，我们引入G-Vendi指标通过梯度空间分布捕捉模型推理路径（CoT）和最终答案中的认知策略差异。
% 重要地，所有的多样性指标（包括AEC,DAK,EA,NAUADC,G-Vendi）均在功能正确的解决方案中评估，确保测量的多样性反映了有意义的算法差异，而不是表面的变化。
To comprehensively evaluate the algorithmic and cognitive diversity of code generated by our method, we design a multi-level, complementary evaluation framework that encompasses semantic distribution at the code level, algorithmic differences, and cognitive strategy diversity in reasoning processes. We employ pass@K to measure the method's capability in exploring correct solutions. Average Embedding Clusters (AEC) quantifies semantic diversity among code samples. Distinct Algorithms at K (DA@K), Effective number of Algorithms (EA) and Normalized Area Under the Diversity Curve (NAUDC) \cite{lee-etal-2025-diversely} directly quantify the diversity of algorithms themselves Exploratively, we introduce the G-Vendi metric\cite{jung2025prismatic} to capture cognitive strategy differences in the model's reasoning paths (CoT) and final answers through gradient space distribution. Importantly, all diversity metrics—including AEC, DA@K, EA, NAUDC, and G-Vendi—are evaluated solely on functionally correct solutions, ensuring that the measured diversity reflects meaningful algorithmic differences rather than superficial variations.

% 我们使用pass@K这是在代码生成评估中广泛使用的指标。形式上，pass@K计算对于给定问题，在K个样本中，至少有一个通过所有测试样例的概率。虽然pass@K主要反映了模型生成正确解的能力，但是它也反映了模型生成代码的多样性：较高的pass@K表明对解空间的更广泛的探索。
\paragraph{pass@K.} We adopt pass@K, a widely used metric in code generation evaluation. Formally, pass@K computes the probability that among K sampled completions for a given problem, at least one passes all test cases. While pass@K primarily reflects a model’s ability to generate correct solutions, it also indirectly captures the diversity of generated code: a higher pass@K suggests broader exploration of the solution space.

\paragraph{Average Embedding Clusters.} 
% 基于embedding的方法很早就被提出，我们使用先进的code embedding 模型jina-code-embeddings-1.5b 来生成代码嵌入并通过CBSCAN聚类对每个问题的正确答案进行聚类，我们以平均聚类个数来衡量算法的语义多样性。AEC通过公式计算，其中Q代表问题集，A_q代表问题q对应的正确答案集。更高的 AEC 表明模型产生的功能正确解在语义空间中分布更广泛，因此表现出更高的语义多样性。
Embedding-based methods have been widely used for code diversity assessment. For each problem, we employ the advanced code embedding model to generate semantic vectors for the correct answers produced by the model. DBSCAN clustering\cite{ester1996density} is then applied to group all correct answers for that problem. We use the average number of clusters to reflect the diversity of the model's generated answers. The average numbers of clusters (AEC) is then computed as:
\begin{equation}
    \text{AEC} = \frac{1}{|Q|} \sum_{q \in Q} C(A_q)
\end{equation}
where $Q$ denotes the set of evaluated problems, and $C(A_q)$ is the numbers of clusters obtained from the correct solution set $A_q$ from problem $q$. A higher AEC suggests that the model produces functionally correct solutions that are more widely dispersed in the semantic space, hence exhibiting higher semantic diversity.
% 遵循\cite{}中的设置，我们使用DA@K，EA以及NAUADC来多角度衡量模型生成代码多样性的能力。我们使用Qwen2.5-7B-Instruct来判断两份代码是否是算法一致的代码从而实现对正确答案集的聚类。我们计算以下指标DA@K,DA@K衡量从解集中抽取K个解能够平均覆盖多少不同算法的答案。EA，衡量解决方案中不同实现方法的分布模式。NAUADC是DA@K随着K增加的积分，衡量不同解集规模下不同算法的平均数量。
\paragraph{DA@K,EA,NAUADC.} 
Following the setup in \cite{lee-etal-2025-diversely}, we adopt DA@K, EA, and NAUADC to comprehensively assess the algorithmic diversity of code generated by models from multiple perspectives. We use a LLM to determine whether two code solutions implement the same algorithm, facilitating the clustering of correct answer sets. We calculate the following metrics: 

DA@K measures the average number of distinct algorithms covered when sampling $K$ solutions from the solution set. Let $M$ denote the number of algorithmic clusters and $s_m$ the size of the $m$-th cluster in a solution set of total size $N$.It is calculated as
\begin{equation}
DA@K = \sum_{m=1}^{M} \left( 1 - \frac{\binom{N - s_m}{K}}{\binom{N}{K}} \right)
\end{equation}

EA characterizes the distribution of solution strategies across clusters. Let $p_m = \frac{s_m}{N}$ be the empirical probability of generating a solution from cluster $m$. It is defined as
\begin{equation}
EA = \exp\left(- \sum_{m=1}^{M} p_m \ln p_m \right)
\end{equation}

NAUADC integrates DA@K over varying sampling budgets to assess the average diversity across solution set sizes. Denoting the maximum evaluation budget as $K_{max}$, we compute
\begin{equation}
    \text{NAUADC} = \frac{1}{K_{\max} - 1} \sum_{k=1}^{K_{\max}} \text{DA@}k
\end{equation}

% 之前的指标主要关注代码级别的多样性。在这项工作中，我们使用 G-Vendi 指标来评估思维链（CoT）推理过程和生成代码的多样性。G-Vendi 通过检查样本在小代理模型的损失梯度空间中的分布情况来捕捉认知策略层面的多样性。follow \cite{} 我们使用轻量级代理模型计算每个样本的损失梯度，对梯度进行归一化，并应用随机投影进行降维。然后，我们从投影后的梯度构建协方差矩阵，并计算其特征值分布的香农熵。G-Vendi 指标定义为该熵的指数：
\paragraph{G-Vendi. }

Previous metrics primarily focus on code-level diversity. We adopt the G-Vendi\cite{jung2025prismatic} metric to assess diversity across both the reasoning process and the generated code. G-Vendi captures diversity at the level of cognitive strategies by examining how samples are distributed in the loss-gradient space of a small proxy model. We compute the loss gradient of each sample using a lightweight proxy model, where $x$ denotes the code problem and $y$ denotes the combined chain-of-thought (CoT) reasoning and final answer:
\begin{equation}
    g_\theta(x, y) = -\nabla \log P(y \mid x; \theta)
\end{equation}

We then normalize the gradients and apply a random projection for dimensionality reduction. We then construct a covariance matrix from the projected gradients and compute the Shannon entropy of its eigenvalue distribution. Specifically, let $\{\lambda_i^K\}$ denote the normalized eigenvalues of the covariance matrix $K$, which form a probability distribution over gradient variation modes. The G-Vendi score is defined as the exponential of this entropy:
\begin{equation}
    \text{G-Vendi}(D) = \exp\left( -\sum_i \lambda_i^K \log \lambda_i^K \right)
\end{equation}

\paragraph{Experimental Setup.} 

% 我们在 LiveCodeBench v6 数据集上评估了基线方法与所提方法的代码生成多样性。我们保留所有待比较方法均能生成至少一个功能正确解的任务。以消除方法间功能正确性差异对多样性评估的干扰，确保所观测的多样性差异纯粹反映方法本身的特性。我们共筛选出95个任务作为评估的子集。
We evaluate the code generation diversity of baseline methods and our proposed approach on the LiveCodeBench v6 dataset. We retain only those tasks for which all methods being compared can generate at least one functionally correct solution, in order to eliminate the confounding effects of differences in functional correctness among methods on the diversity assessment and ensure that the observed diversity differences purely reflect the inherent characteristics of the methods themselves. This filtering results in a total of 95 tasks, which constitute our evaluation subset.

% 实验设置：我们使用jina-code-embeddings-1.5b作为先进的代码嵌入模型，使用Qwen2.5-7B-Instruct(温度设置为0)来判断两个代码是否属于同一算法来计算DA@K,EA等指标，对于DA@K我们选取K=10来反映模型生成的代码多样性，对于NAUADC，我们设置K_max=100以覆盖所有的正确答案。对于G-Vedi指标，我们采用Qwen2.5-0.5B-Instruct来计算梯度损失。
We employ \textbf{jina-code-embeddings-1.5b}\cite{kryvosheieva2025efficient} as the advanced code embedding model. \textbf{Qwen2.5-7B-Instruct} (with temperature set to 0) is used to determine whether two code solutions implement the same algorithm, enabling the clustering required for calculating metrics such as $DA@K$ and $EA$. For $DA@K$, we select $K=20$ to reflect the diversity of code generated by the models. For NAUADC, we set $K_{max}=60$ to cover the range of all correct solutions generated per problem. For G-Vendi, we use \texttt{Qwen2.5-0.5B-Instruct} to compute the loss gradient.

\subsection{Diversity Evaluation}
\begin{tcolorbox}[takeawaysbox]
(1) \textbf{Heterogeneous multi-agent collaboration substantially enhances performance and solution-space exploration capability.} Heter-MARS$^2$ achieves the highest pass@K scores across both 8B (75.4\%) and 14B (81.7\%) model scales, significantly outperforming baseline methods. This demonstrates its superior ability to expand the exploration of high-quality regions within the underlying solution space.

(2) \textbf{System-level diversity consistently increases with both the number of agents and architectural heterogeneity.} Both scaling the number of models and adopting heterogeneous architectures lead to consistent improvements in algorithmic diversity and distributional uniformity. This indicates that the multi-agent paradigm not only broadens the policy space but also mitigates over-reliance on any single solution trajectory.

(3) \textbf{Multi-agent collaboration effectively expands the cognitive strategy space.} The G-Vendi score exhibits a systematic positive correlation with the number of collaborating agents. This confirms that the multi-agent framework fosters greater diversity not only at the code level but also in high-level cognitive reasoning strategies.
\end{tcolorbox}

%% 确定结论分析的角度
%% 1. 多模型的引入可以增加代码多样性和策略多样性（为什么G-Vendi指标上三模型在中等和困难任务上反而降了，这里的分析是因为abmcts算法倾向于生成浅的树导致模型之间的协作不够充分）
%% 2. 并行采样的EA指标最高，这反映了并行采样生成多样性的代码最为均匀，多模型的引入也提高了算法生成的均匀性
%% 3. 训练可以增加代码多样性，已有的训练结果表明经过训练之后，代码多样性的指标都提高了
%% 4. 多模型的RMATS-T+方法能够提高7B模型在复杂任务上的多样性表现，同时提高策略多样性。对于14B模型的改进则不明显
%% 5. 简单任务上，模型体现出思维链多样但算法趋同的趋势，在AEC,DAK,EA,NAUADC等直接衡量代码的指标上，简单任务<中等任务<复杂任务，而在G-Vendi同时衡量思维链和代码的指标上，这一趋势出现了翻转，进一步的实验表明，这是由于简单任务上，模型表现出了更丰富的思维链，但最终的算法上并不多样。
%我们计算了基线方法和我们提出方法的多样性指标，详细的在表。我们的表格揭示了几点核心结论：

We computed diversity metrics for the code generated by the baseline methods and our proposed approach. The results, summarized in Table~\ref{tab:diversity_results}, reveal several key insights: 

% MA-ReMASS达到了最高的pass@K. 在8B模型下，MA-ReMASS取得了最高的pass@K(75.4)，在14B模型下，MA-ReMASS(2-model)同样取得了最高的pass@K(81.7),显著超过了基线方法，这证明了我们提出的方法可以有效扩展模型探索潜在解空间的能力，一定程度上反映了我们的方法可以提高系统生成的多样性。
\paragraph{Heter-MARS$^2$ achieves the highest pass@K scores within both the 8B and 14B model families.}Specifically, in the 8B model family, Heter-MARS$^2$ attains a pass@K of 75.4\%, and in the 14B model family, Heter-MARS$^2$ (2-model) achieves a pass@K of 81.7\%—significantly outperforming the baseline methods. These results demonstrate that our proposed approach effectively enhances the model’s capacity to explore the underlying solution space, which, to some extent, reflects an improvement in the diversity of the generated code.

\paragraph{Algorithmic diversity and distributional uniformity improve as the number of models increases.} This trend is confirmed in Table \ref{tab:diversity_results} vanilla TTS result (14B), where NAUADC rises from 1.659 to 1.962, and is further enhanced by our approach: in the 8B model family, multi-agent collaboration (MA-2model) increases DA@K from 1.554 (SA-1model) to 1.677 and raises NAUADC to 1.750. The concurrent improvement in the EA metric—e.g., reaching 1.544 for 14B Heter-MARS$^2$—demonstrates that the multi-agent mechanism not only expands the policy space but also effectively promotes more uniform algorithmic distributions, thereby mitigating over-reliance on a single solution path.

\begin{table}[H]
\centering
\caption{Performance and diversity comparison of different methods on 8B and 14B models.}
\label{tab:diversity_results}
\small
\begin{tabular}{l c c c c c c}
\toprule
Method & pass@K & AEC & DA@K & EA & NAUADC & G-Vendi \\
\midrule
\multicolumn{7}{c}{\textbf{8B Model}} \\
\midrule
\multicolumn{7}{l}{\textbf{Vanilla-TTS}} \\
Q-8B      & 68.6 & 1.053 & 1.631 & 1.332 & 1.670 & 6.688 \\
A-8B     & 66.9 & 1.052 & 1.641 & 1.348 & 1.682 & 6.946 \\
Q-8B \& A-8B    & 72.0 & 1.295 & 1.634 & 1.321 & 1.686 & 7.146 \\
% Q-8B \& A-8B \& D-8B  & 67.4 & 1.684 & 1.802 & 1.419 & 1.871 & 7.816 \\
\midrule
\multicolumn{7}{l}{\textbf{Vanilla-GRPO}} \\
Q-8B      & 73.7 & 1.274 & {1.733} & 1.375 & 1.787 & 6.690 \\
A-8B      & 69.1 & 1.116 & 1.625 & 1.350 & 1.673 & 6.799 \\
Q-8B \& A-8B    & 72.0 & 1.190 & 1.645 & 1.333 & 1.707 & 7.063 \\
\midrule
\multicolumn{7}{l}{\textbf{Homo-MARS$^2$}} \\
Q-8B       & 71.4 & {1.463} & 1.554 & 1.271 & 1.618 & 5.962 \\
A-8B      & 69.7 & 1.052 & 1.608 & 1.307 & 1.658 & 6.564 \\
Q-8B \& A-8B    & 72.6 & 1.315 & 1.664 & 1.341 & 1.716 & 6.403 \\
\midrule
\multicolumn{7}{l}{\textbf{Heter-MARS$^2$}} \\
Q-8B \& A-8B    & \textbf{75.4} & {1.431} & {1.677} & {1.348} & \textbf{1.750} & {6.901} \\
\midrule
% \multicolumn{7}{l}{\textbf{MARS$^2$-T+}} \\
% Qwen3-8B       & 73.1 & 1.14 & 1.571 & 1.314 & 1.612  & 6.203 \\
% DA-8B$_{\text{Heter-MARS-T+}}$  & 72.0 & 1.337 & 1.680 & 1.357 & 1.724 & 7.445 \\
% \midrule
\multicolumn{7}{c}{\textbf{14B Model}} \\
\midrule
\multicolumn{7}{l}{\textbf{Vanilla-TTS}} \\
Q-14B      & 77.7 & 1.032 & 1.517 & 1.261 & 1.575 & 7.666 \\
A-14B      & 75.4 & 1.000 & 1.594 & 1.320 & 1.659 & 6.908 \\
D-14B     & 68.0 & 1.042 & 1.712 & 1.406 & 1.759 & 7.392 \\
Q-14B \& A-14B   & 79.4 & 1.095 & 1.698 & 1.375 & 1.766 & 7.668 \\
Q-14B \& A-14B \& D-14B  & 79.4 & 1.231 & 1.882 & 1.478 & 1.962 & 8.808 \\
\midrule
\multicolumn{7}{l}{\textbf{Vanilla-GRPO}} \\
Q-14B      & 77.1 & 1.095 & 1.703     & 1.369     & 1.755    & 6.344 \\
A-14B      & 76.0 & 1.063 & 1.705     & 1.405     & 1.753    & 5.970 \\
D-14B      & 73.7 & 1.200 & 1.788 & 1.441 & 1.840 & 7.897 \\
Q-14B \& A-14B   & 78.9  & 1.179 & 1.836     & 1.448     & 1.939    & 6.534 \\
Q-14B \& A-14B \& D-14B  & 79.4   & 1.147 & 1.852    & 1.464    & 1.919     & 8.032 \\
\midrule
\multicolumn{7}{l}{\textbf{Homo-MARS$^2$}} \\
Q-14B      & 78.9 & 1.073 & 1.595 & 1.300 & 1.644 & 7.324 \\
A-14B      & 81.1 & 1.084 & 1.685 & 1.358 & 1.757 & 6.335 \\
D-14B      & 70.9 & 1.168 & 1.692 & 1.373 & 1.739 & 8.335 \\
Q-14B \& A-14B   & 79.4 & 1.042 & 1.687 & 1.367 & 1.749 & 7.250 \\
Q-14B \& A-14B \& D-14B  & 78.9 & 1.147 & 1.837 & 1.445 & 1.889 & 8.750 \\
\midrule
\multicolumn{7}{l}{\textbf{Heter-MARS$^2$}} \\
Q-14B \& A-14B   & \textbf{81.7} & 1.042 & 1.712 & 1.369 & 1.793 & 6.980 \\
Q-14B \& A-14B \& D-14B  & 80.0 & 1.116 & \textbf{1.941}     & \textbf{1.544}     & \textbf{2.033}     & 8.719 \\
% \midrule
% \multicolumn{7}{l}{\textbf{MARS$^2$-T+}} \\
% Qwen-14B      & 74.9 & 1.074 & 1.546 & 1.305 & 1.586 & 6.337 \\
% Q-14B \& A-14B$_{\text{Heter-MARS-T+}}$    & 80.6 & 1.032 & 1.772 & 1.431 & 1.844 & 7.195 \\
% Q-14B \& A-14B \& D-14B$_{\text{Heter-MARS-T+}}$  & 80.6 & 1.063 & 1.870 & 1.482 & 1.954 & 8.002 \\
\bottomrule
\end{tabular}
\end{table}

\begin{figure}[t]
    \centering
    \label{fig:diversity_metric}

    % 第一行
    \begin{subfigure}[b]{0.32\textwidth}
        \centering
        \includegraphics[width=\linewidth]{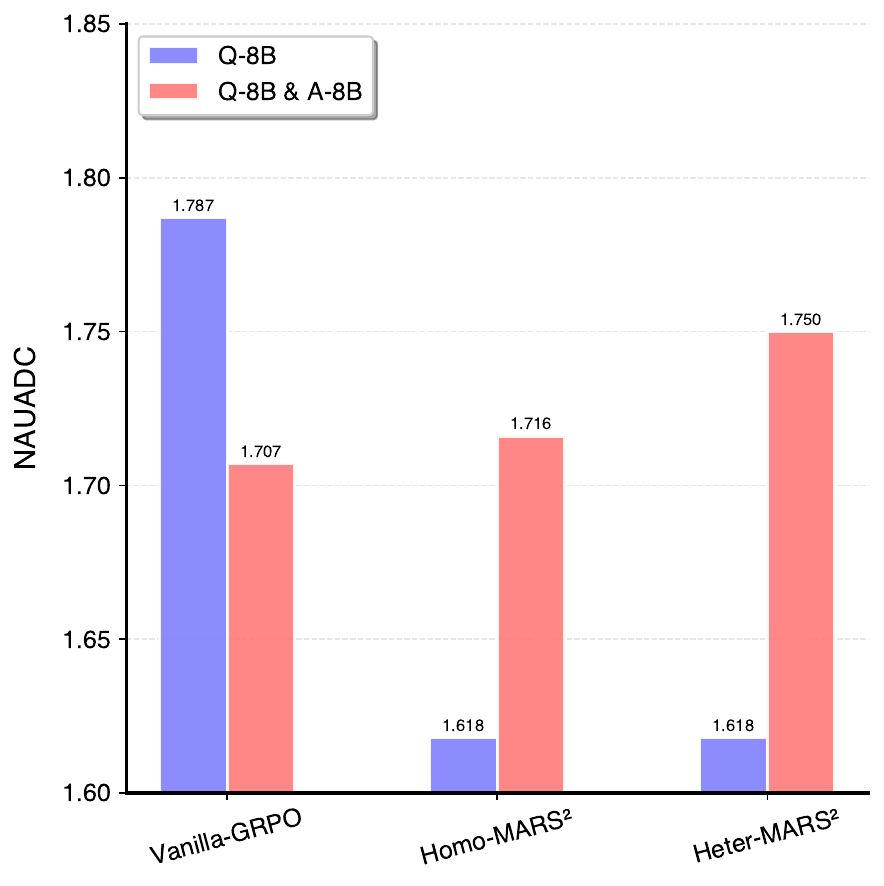}
        \subcaption{Comparison of NAUADC Diversity Trends with Increasing Model Counts on the 8B Model Series}
        \label{fig:nauadc_8B}
    \end{subfigure}
    \hfill
    \begin{subfigure}[b]{0.32\textwidth}
        \centering
        \includegraphics[width=\linewidth]{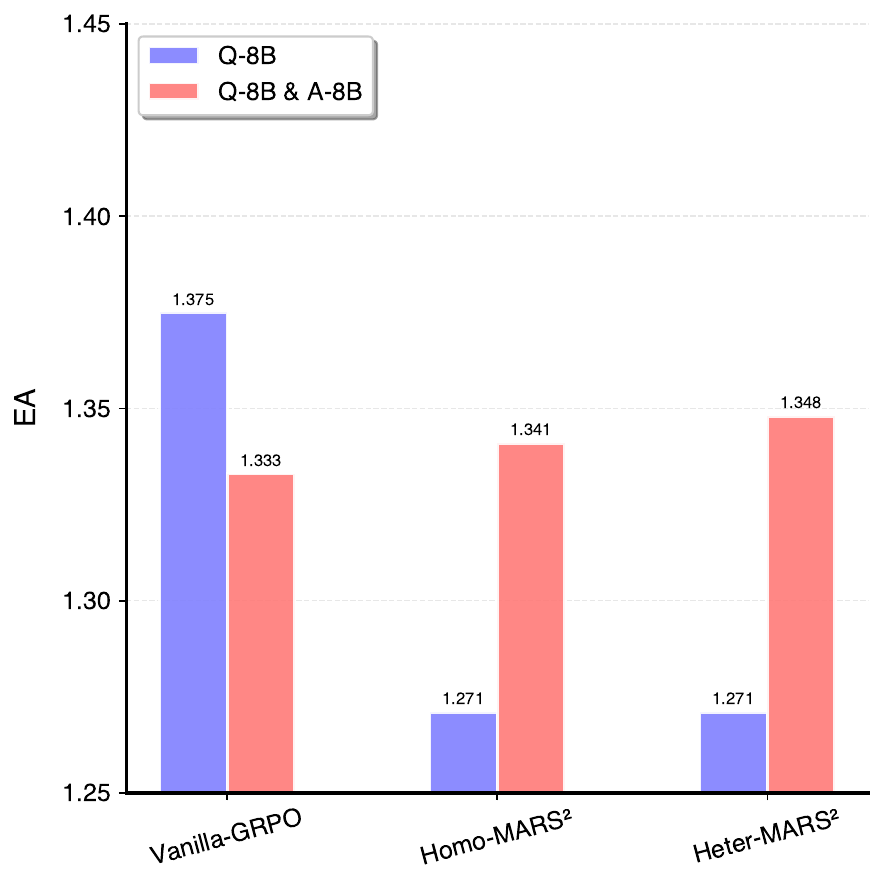}
        \subcaption{Comparison of EA with Increasing Model Counts on the 8B Model Series \\ \quad}
        \label{fig:ea_8B}
    \end{subfigure}
    \hfill
    \begin{subfigure}[b]{0.32\textwidth}
        \centering
        \includegraphics[width=\linewidth]{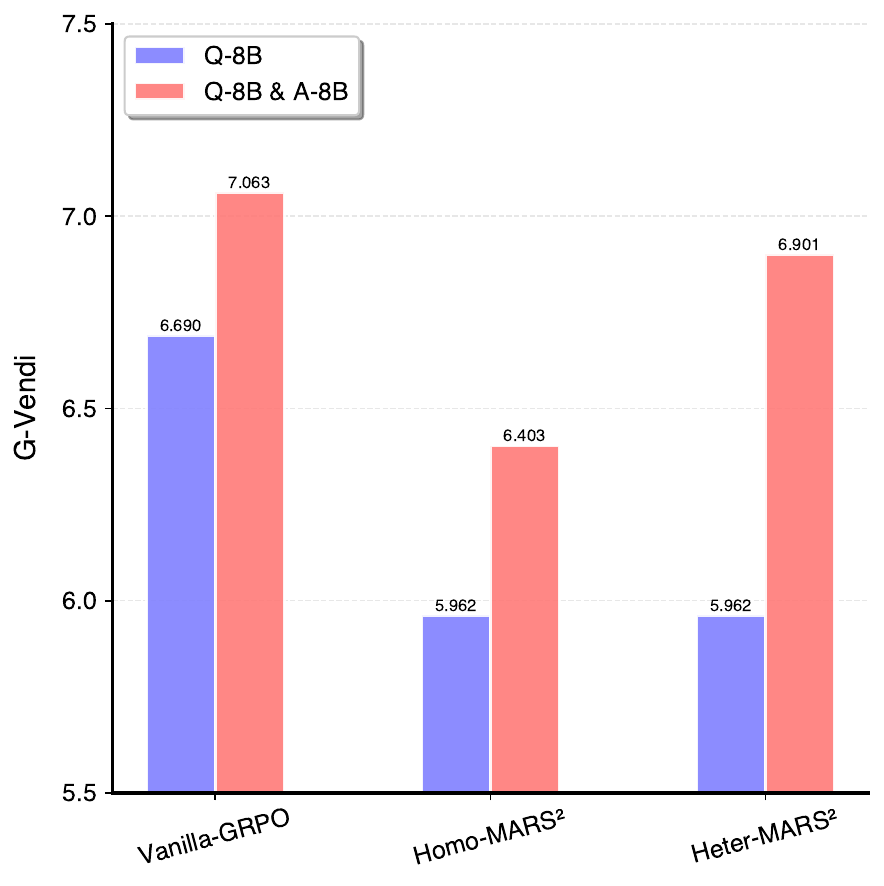}
        \subcaption{Comparison of G-Vendi Metrics with Increasing Model Counts on the 8B Model Series}
        \label{fig:vendi_8B}
    \end{subfigure}

    \vspace{1em} % 行间距

    % 第二行
    \begin{subfigure}[b]{0.32\textwidth}
        \centering
        \includegraphics[width=\linewidth]{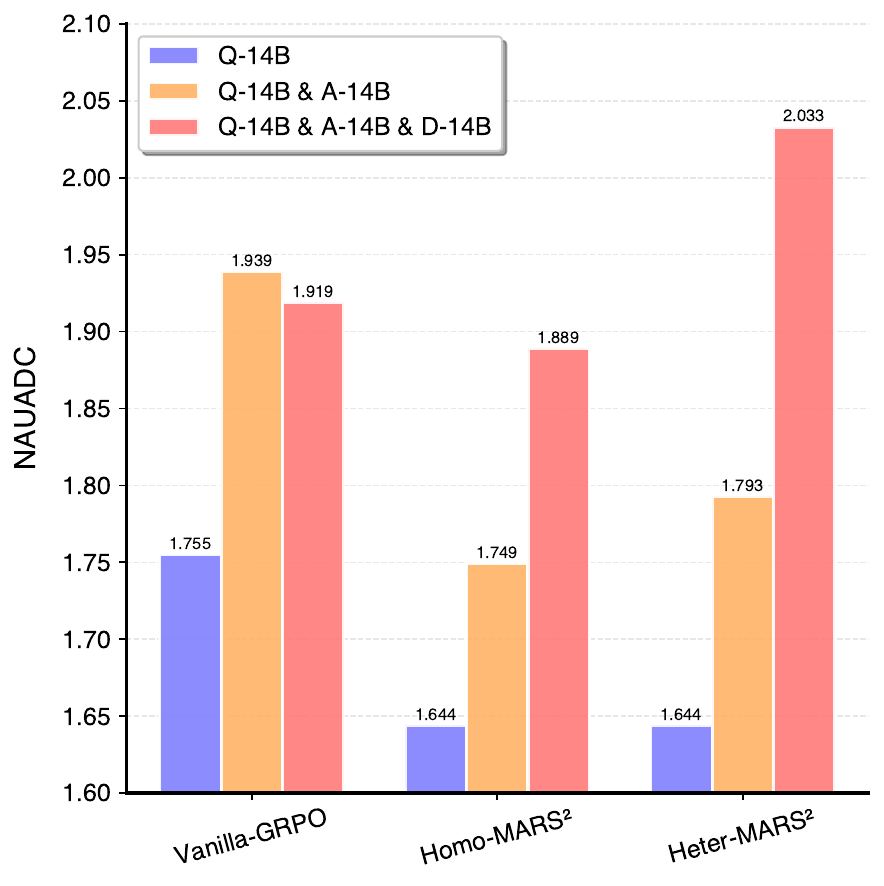}
        \subcaption{Comparison of NAUADC Diversity Trends with Increasing Model Counts on the 14B Model Series}
        \label{fig:nauadc_14B}
    \end{subfigure}
    \hfill
    \begin{subfigure}[b]{0.32\textwidth}
        \centering
        \includegraphics[width=\linewidth]{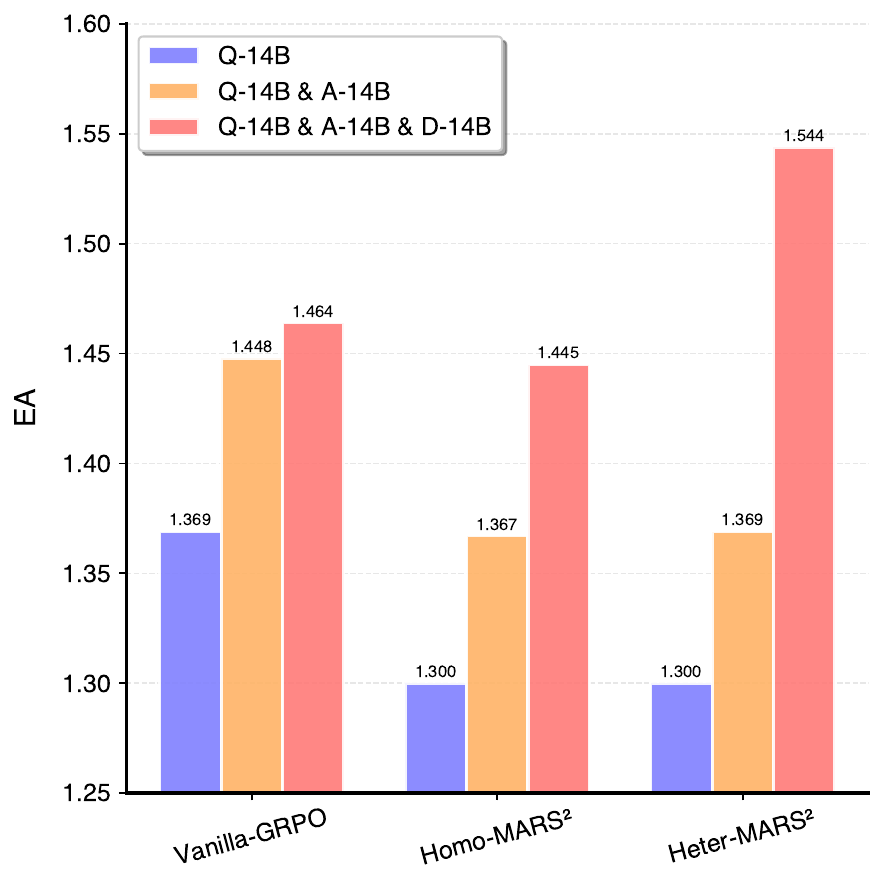}
        \subcaption{Comparison of EA with Increasing Model Counts on the 14B Model Series \\ \quad}
        \label{fig:ea_14B}
    \end{subfigure}
    \hfill
    \begin{subfigure}[b]{0.32\textwidth}
        \centering
        \includegraphics[width=\linewidth]{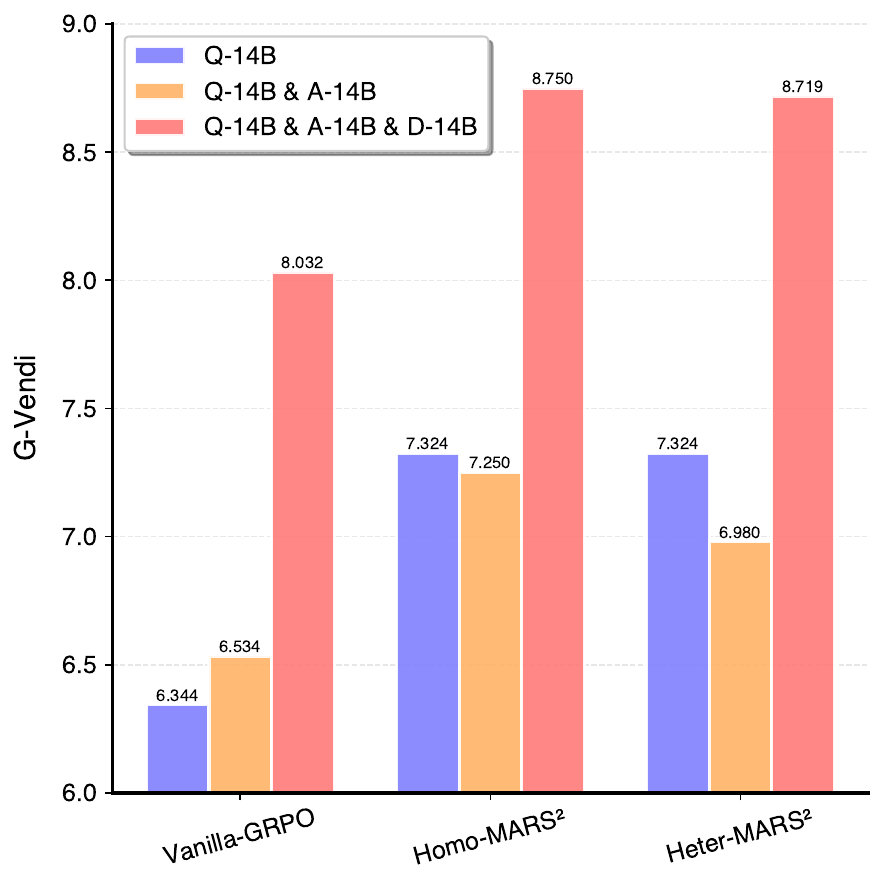}
        \subcaption{Comparison of G-Vendi Metrics with Increasing Model Counts on the 14B Model Series}
        \label{fig:vendi_14B}
    \end{subfigure}
    \caption{Analysis of Diversity metrics Trends with Increasing Model Counts across 8B and 14B Model Series.}
\end{figure}

%算法多样性与分布均匀性随模型数量增加提升。这一趋势在表\ref{tab:diversity_results} 中得到验证where AB-MCTS (14B) NAUADC 由 1.659 升至 1.962，并被我们的方法进一步强化：在 8B 模型上，多体协作（MA-2model）相比单体（SA-1model）将 DA@K 从 1.554 提升至 1.677，NAUADC 提升至 1.750。伴随 EA 指标的同步增长（如 14B MA-ReMASS 达 1.544），这证明多体机制在扩展策略空间的同时，有效促进了算法分布的均匀化，缓解了模型对单一解题路径的依赖。
%\multicolumn{7}{l}{\textbf{Vanilla-T+}} \\
% Q-8B      & 71.4 & 1.074 & 1.669 & 1.352 & 1.711 & 6.858 \\
% Q-8B+A-8B    & 71.4 & 1.095 & 1.666 & 1.343 & 1.708 & 7.362 \\
% Q-8B+A-8B+DS-8B  & 72.0 & 1.474 & 1.862 & 1.408 & 1.949 & 8.361 \\
% \multicolumn{7}{l}{\textbf{Vanilla-T+}} \\
% Qwen3-14B      & 78.3 & 1.031 & 1.615 & 1.334 & 1.660 & 7.651 \\
% Q-14B+A-14B    & 78.9 & 1.084 & 1.649 & 1.334 & 1.709 & 7.803 \\
% Q-14B+A-14B+DC-14B  & 80.0 & 1.179 & 1.869 & 1.501 & 1.946 & 9.168 \\
% MA-ReMASS 比 SA-ReMASS 更进一步提升了算法多样性。在 8B 系列模型中，MA-ReMASS 在同等模型数量下全面超越 SA-ReMASS，其中 DA@K 达到 1.677，NAUADC 达到 1.750（相比 SA 分别提升 +0.013 和 +0.034）。这一趋势在 14B 系列模型中得到了复现与增强（NAUADC 进一步由 1.749 提升至 1.793）。这说明 MA-ReMASS 的联合训练机制能够比单独训练进一步提高算法多样性，产生更多的可行解。
\paragraph{Heter-MARS$^2$ demonstrates superior capability in enhancing algorithmic diversity compared to Homo-MARS$^2$.} At the 8B model scale, Heter-MARS$^2$ consistently outperforms Homo-MARS$^2$ under equivalent model-count configurations, achieving a DA@K of 1.677 and an NAUADC of 1.750—improvements of 0.013 and 0.034, respectively, over the SA baseline. This advantage is further validated and amplified at the 14B scale, where NAUADC increases significantly from 1.749 (SA) to 1.793 (Heter-MARS$^2$). These results indicate that the joint training mechanism employed by Heter-MARS$^2$ effectively overcomes the limitations of independent training, enabling the models to explore a broader algorithmic space and thereby substantially expanding the solution space.

%多智能体协作扩展了认知策略空间。与代码级指标（DA@K、NAUADC）中观察到的趋势一致，G-Vendi 分数与模型数量呈正相关关系。在8B模型系列，这一趋势十分明显，在 14B 模型系列中，从单模型基线过渡到 3 模型协作，所有方法均获得显著提升。值得注意的是，我们的 ReMASS-T+（3model）实现了最高的 G-Vendi 分数 9.168，显著优于 Vanilla-GRPO 基线（8.032）。这说明TTS增强可以扩展认知策略空间。我们对TTS增强方法进行了额外的多样性分析，见附录ref{sec:appendix_diversity_marts}
\paragraph{Multi-agent collaboration expands the cognitive strategy space.} Consistent with the trends observed in code-level diversity metrics (DA@K and NAUADC), the G-Vendi score exhibits a positive correlation with the number of models. This trend is particularly pronounced in the 8B model family, and in the 14B model family, all methods show significant gains when transitioning from a single-model baseline to three-model collaboration. We provide additional diversity analyses of the TTS-enhanced method in Appendix~\ref{sec:appendix_diversity_marts}.

%Notably, our MARS$^2$-T+ (3-model) achieves the highest G-Vendi score of 9.168, substantially outperforming the Vanilla-GRPO baseline (8.032). This result indicates that our TTS augmentation effectively broadens the cognitive strategy space.We provide additional diversity analyses of the TTS-enhanced method in Appendix~\ref{sec:appendix_diversity_marts}.

%\subsection{embedding}
% 单模型/多模型输出50个答案，用coder模型算embedding，余弦相似度聚类or降维可视化，多模型分布更广+多体ab-mcts（系统的能力/个体能力）

%\subsection{Entropy}\label{part:2_1}
% 梯度空间上的分布熵，0.5B模型对一个答案算梯度——降维成1024维向量——多个答案聚合成矩阵，求矩阵的协方差（反映梯度方向的相关性） niups2025

%\subsection{Diversity}\label{eval_setup}
% 大模型评测解法的相似度——聚类——EA指标衡量解法的多样性（第m个簇占总正确解法的数量的比例——算熵）  反映模型不偏向于其中一种解法    emnlp2025

%\subsection{code2text}
% code转换成伪代码，衡量文本的相似度（衡量推理步骤的相似性）

%% file: sections/2_preliminary.tex
% [Unifying Autoregressive and Diffusion-Based Sequence Generation](https://arxiv.org/pdf/2504.06416)
% [Block Diffusion: Interpolating Between Autoregressive and Diffusion Language Models](https://arxiv.org/abs/2503.09573)
% [SSD-LM: Semi-autoregressive Simplex-based Diffusion Language Model for Text Generation and Modular Control](https://arxiv.org/abs/2210.17432)

\section{Preliminary}
\label{sec:preliminary}

\subsection{GRPO}
\label{ssec:grpo}
% GRPO（Group Relative Policy Optimization）是由 DeepSeekMath 提出的一种强化学习算法，摆脱了 PPO（Proximal Policy Optimization）对值模型的依赖。其核心思想是引入组内相对奖励计算优势，通过将同一提示下多个采样输出的平均奖励作为基线来规范化奖励，从而显著提升了策略更新的稳定性和效率。
% GRPO (Group Relative Policy Optimization) is a reinforcement learning algorithm proposed by DeepSeekMath\cite{shao2024deepseekmath}, which eliminates the dependence on critic model found in PPO \cite{schulman2017proximal}. Its core idea is to introduce relative rewards within group to calculate advantages. By using the average reward of multiple sampling outputs under the same prompt as the baseline to normalize the reward, a critic-free training method has been achieved, effectively reducing the training cost.

GRPO (Group Relative Policy Optimization), a reinforcement learning algorithm introduced by DeepSeekMath\cite{shao2024deepseekmath}, eliminates the need for the critic model used in PPO \cite{schulman2017proximal}. Its core idea is to leverage relative rewards within a group of outputs to estimate advantages. Specifically, it uses the average reward of multiple outputs sampled from the same prompt as a baseline for reward normalization. This achieves a critic-free training paradigm, which effectively reduces the computational cost of training.

% 给定一个prompt p, 从old策略模型$\pi_{\theta_{old}}$中采样一组输出${o_1,o_2,...,o_G}$，其中G代表组的大小。每一个输出$o_i$通过奖励模型或环境反馈等方式得到一个标量奖励$r_i$。为了计算优势，通过应用z分数对群体的奖励进行归一化，将优势函数设置为归一化后的奖励。$\bar{r}=\frac{r_i-mean(r_1,r_2,...,r_G)}{std(r_1,r_2,...,r_G)+\epsilon}$

% 其中，$\epsilon$是一个非常小的常数值以保持数值稳定。优势函数被设置为归一化后的奖励，即$\hat{A_i}=\bar{r_i}$
Given a prompt $p$, a set of outputs \{${o_1,o_2,\dots,o_G}$\} is sampled from the old policy model $\pi_{\theta_{old}}$, where $G$ represents the group size. Each output $o_i$ receives a scalar reward $r_i$ through a reward model or environment feedback. To compute the advantage, the rewards for the group are normalized by applying a z-score, with the advantage of all tokens set to the normalized reward:
\begin{equation}
\label{grpo-norm}
\hat{A}_{i,t}=\frac{r_i-\text{mean}(r_1,r_2,\dots,r_G)}{\text{std}(r_1,r_2,\dots,r_G)}
\end{equation}

% 策略通过最大化下述目标被更新，来提高生成相对积极优势的响应
The policy $\pi_{\theta}$ is updated by maximizing the following objective to increase the probability of generating responses with positive relative advantages:

\begin{equation}
\label{grpo-update}
\resizebox{0.9\linewidth}{!}{$
\begin{aligned}
\mathcal{J}_{GRPO}(\theta)&=\mathbb{E}\big[q\sim P(Q), \{o_i\}_{i=1}^G \sim \pi_{\theta_{old}}(O|q)\big] 
\\  &\frac{1}{G}\sum_{i=1}^{G}\frac{1}{|o_i|} \sum_{t=1}^{|o_i|} \bigg\{\text{min}\Big[\frac{\pi_{\theta}(o_{i,t}|q,o_{i,<t})}{\pi_{\theta_{old}}(o_{i,t}|q,o_{i,<t})} \hat{A}_{i,t}, \text{clip} \Big(\frac{\pi_{\theta}(o_{i,t}|q,o_{i,<t})}{\pi_{\theta_{old}}(o_{i,t}|q,o_{i,<t})}, 1-\epsilon, 1+\epsilon \Big) \hat{A}_{i,t} \Big] 
- \beta\mathbb{D}_{KL}(\pi_{\theta}||\pi_{ref}) \bigg\} 
\end{aligned}
$}
\end{equation}

% 其中 $\mathbb{D}_{KL}(\pi_{\theta}||\pi_{ref})$ 衡量当前策略与参考策略之间的差异，而 $\beta$ 控制这种正则化的强度以实现稳定的策略更新。

where $\mathbb{D}_{KL}(\pi_{\theta}||\pi_{ref})$ measures the divergence between the current and reference policies, and $\beta$ controls this regularization strength for stable policy updates.

\subsection{AB-MCTS}
\label{ssec:ab-mcts}

% 介绍 AB-MCTS 的作用、目标和核心优势（深度/广度自适应）
AB-MCTS (Adaptive Branching Monte Carlo Tree Search) is an inference-time computation framework designed to scale LLM reasoning\cite{inoue2025wider}. It acts as a unified strategy, combining the parallel output diversity of repeated sampling with the multi-turn solution refinement of sequential methods. This approach implements an adaptive MCTS that achieves dynamic control over search depth and width.

% 引入多智能体集合，并说明采用的 Multiple GEN Nodes 策略（Algorithm II）
The framework is designed to utilize a set of $m$ heterogeneous or Homogeneous agents (LLMs), denoted as $\{\mathcal{A}_j\}_{j=1}^m$. To effectively capture the local performance context of each agent, we adopt the \textit{Multiple GEN Nodes} strategy, in which every search node conceptually maintains a distinct GEN node for each agent $\mathcal{A}_j$, allowing independent tracking of their exploration potential.
% 介绍核心决策的元动作：GEN 和 CONT
The action space for each agent $\mathcal{A}_j$ is simplified into two distinct meta-actions through node aggregation:
\begin{itemize}
    \item \textbf{GEN (Go Wider):} Represents exploration by generating a new candidate answer, expanding the search width.
    \item \textbf{CON (Go Deeper):} Represents exploitation by refining an existing solution derived from a child node.
\end{itemize}

% 描述两步分层选择流程 (Hierarchical Selection)
The overall selection process operates hierarchically:
\begin{itemize}
    \item \textbf{Model Selection:} The primary decision of which agent to utilize is formulated as a Multi-Armed Bandit (MAB) problem. Thompson Sampling is applied to the aggregated performance statistics of all agents to select the agent $\mathcal{A}_{j^*}$ with the highest current potential for the expansion step.
    \item \textbf{Node Selection:} Once the agent $\mathcal{A}_{j^*}$ is determined, a secondary decision is made within its specific sub-tree. This choice between exploration (GEN) and exploitation (CON) is modeled as a binary bandit problem. Thompson Sampling is used again to choose the specific best action associated with the selected agent $\mathcal{A}_{j^*}$.
\end{itemize}

% 插入决策机制（Beta 分布）的数学细节
For normalized scores $r \in [0, 1]$, the posterior distribution of the expected score for each action $a \in \{\text{GEN}, \text{CON}\}$ is modeled as a Beta distribution $Beta(\alpha_a, \beta_a)$. The parameters are updated using the history of observed scores $\mathcal{D}_a = \{r_1, \dots, r_K\}$ associated with that action:
\begin{equation}
    \alpha_a = \tilde{\alpha} + \sum_{k=1}^{K} r_k, \quad \beta_a = \tilde{\beta} + \sum_{k=1}^{K} (1 - r_k)
\end{equation}
where $\tilde{\alpha}$ and $\tilde{\beta}$ are prior hyperparameters.

% 总结：说明在 m=1 时的退化情况
This heterogeneous multi-agent framework naturally accommodates the homogeneous multi-role scenario: when the system contains only one agent ($m=1$) with different roles, the Model Selection step automatically selects that agent, and the search relies solely on the Node Selection step to adaptively balance exploration (GEN) and exploitation (CON).

%% file: sections/7_related_work.tex
\section{Related Works}
\label{sec:related_work}
%Our work, which introduces a novel Multi-LRM Scaling paradigm to systematically unify test-time scaling and reinforcement learning through collaborative multi-agent interactions, is positioned at the intersection of two rapidly advancing research areas: LLM-based multi-agent test-time scaling and multi-agent reinforcement learning.

\subsection{Reinforcement Learning}
Recent progress in reinforcement learning has efficiently strengthened reasoning and alignment capabilities of LLM by facilitating policy optimization\cite{zhang2025survey,liu2025reinforcement,zhang2025landscape}. 
\subsubsection{Reinforcement Learning for LLM}
Several representative methods, such as PPO\cite{schulman2017proximal} and VC-PPO\cite{yuan2025s}, rely on explicit value estimation to provide stable training dynamics. While value-free methods including GRPO\cite{shao2024deepseekmath}, Dr.GRPO\cite{liu2025understanding}, DAPO\cite{yu2025dapo} and GSPO\cite{zheng2025group} reduce dependence on critic models and yield more scalable optimization. Preference optimization-based methods, such as DPO\cite{rafailov2023direct},  Focused-DPO\cite{zhang2025focused} and Selective DPO\cite{gao2025principled}, further simplify the learning pipeline by aligning models directly with preference data, bypassing reward modeling altogether. These methods collectively provide diverse optimization pathways for LLM training, advancing model performance.

\subsubsection{Multi-Agent Reinforcement Learning}

Whereas supervised fine-tuning excel at enhancing individual LLMs but cannot effectively capture emergent multi-agent dynamics (e.g., long-horizon coordination, role specialization, and credit assignment), reinforcement learning has been increasingly employed to directly optimize collaborative policies in LLM-based multi-agent systems, targeting sustained improvements in collective performance~\cite{sun2024llm,li2024survey}. A common paradigm uses external validator scores as reward signals to jointly fine-tune multiple agents, as demonstrated by MAPoRL~\cite{park2025maporl}. To better handle agent heterogeneity and training instability, subsequent works incorporate group-aware optimization and stability mechanisms. JoyAgents-R1~\cite{han2025joyagents} combines GRPO with marginal-benefit selection and adaptive memory to enable stable co-evolution of diverse agents. MAGRPO~\cite{liu2025llm} further introduces a centralized group advantage function that explicitly optimizes joint policies for cooperative tasks. For large-scale systems with highly specialized agents, MarsRL~\cite{Marsrl2025} proposes agentic pipeline parallelism paired with agent-specific rewards, achieving significant reasoning gains in open-source multi-agent settings. Together, these methods mark a clear evolution from basic shared-reward fine-tuning to group-level and scalable policy optimization.

\subsection{Test-Time Scaling}

Test-Time Scaling (TTS) has emerged as a compelling, training-agnostic methodology for augmenting the performance of LLMs by judiciously increasing the computational budget during the inference phase\cite{xia2025generative,zhang2025and}. In this section, test-time scaling is categorized based on the paradigms of single-agent and multi-agent systems. %While TTS can be broadly applied, its methods differ depending on whether a single agent or multiple agents are involved.

\subsubsection{Single-Agent}  

Parallel Scaling is a common test-time scaling method\cite{zhang2025and}, in which a single LLM-based agent generates multiple candidate responses to the same query, followed by an aggregation mechanism that selects or refines the final output. One of the most widely studied methods is Best-of-N\cite{snell2025scaling} strategy, where $N$ independent samples are generated, and the one with the highest likelihood or score is selected. Another common aggregation method is Majority Voting\cite{brown2024large,wang2022self,nguyen2024consistent}, which selects the most frequently generated answer across multiple reasoning paths. Another prominent method is Sequential Scaling\cite{zhang2025and}, which explicitly directs the LLM-based agent to perform iterative computations based on intermediate steps. Chain-of-Thought (CoT)\cite{wei2022chain,li2025system,chen2025towards,ji2025test} prompts the model to generate detailed, step-by-step reasoning trajectory. Building upon this, researchers have proposed hybrid scaling methods\cite{xia2025generative,zhang2025and} combining the complementary benefits of parallel and sequential scaling, such as Tree-of-Thoughts (ToT)\cite{yao2023tree}, Graph-of-Thoughts (GoT)\cite{besta2024graph,yao2024got}, and Monte Carlo Tree Search (MCTS)-based reasoning methods\cite{xie2024monte,zhou2023language}. They significantly enhance the LLM's reasoning capabilities by exploring a wider search space.

\subsubsection{Multi-Agent}

%Extensive research has explored leveraging multi-agent collaboration to expand the space of potential solutions, thereby mitigating the capability limitations and decision-making biases of LLM-based single-agent\cite{wan2025fano}. 

%While single-agent TTS focuses on exploring multiple reasoning paths within a single model, multi-agent TTS leverages collaboration among agents to overcome inherent limitations and biases of single agent. 

Single-agent test-time scaling focuses on exploring diverse reasoning paths within a single model to expand the search space.  In contrast, multi-agent test-time scaling leverages structured collaboration among agents, integrating their complementary advantages to effectively overcome the inherent limitations and intrinsic biases of single-agent systems\cite{wan2025fano}.
A prominent avenue within multi-agent test-time scaling focuses on consensus. The seminal Multi-Agent Debate framework \cite{liang2024encouraging,chan2023chateval,chern2024combating} establishes a debate-centric structure to promote divergent ideation, wherein multiple agents engage in a reciprocal “tit-for-tat” dialogue under the supervision of an adjudicator. 

%Multi-Agent Verification\cite{lifshitz2025multi} further scales computational resources by deploying multiple verifiers to enhance reliability and robustness of the final output. 

A second critical method involves structured search. \cite{inoue2025wider}introduces Adaptive Branching Monte Carlo Tree Search (AB-MCTS), which leverages multi-agent collaboration for test-time scaling by unifying distributed exploration of solution spaces and collective feedback-driven exploitation. 
%In addition, \cite{song2025ctts} proposes Collective Test-Time Scaling(CTTS)  that integrates agent collaboration search and mixture of reward models to enhance LLM inference. 
Structured planning is another focus. Frameworks such as Chain-of-Agents\cite{li2025chain} and PlanGEN\cite{parmar2025plangen} concentrate on the generation of intricate planning and reasoning trajectories via multi-agent collaboration. The integration of external tools has also been a focus, exemplified by TUMIX\cite{chen2025tumix} , which employs a heterogeneous mixture of tool-use strategies. This paradigm has been successfully applied to specialized domains, including METAL\cite{li2025metal} for chart generation and Multi$^2$\cite{cao2025multi2} for multi-document processing. 

%% file: sections/8_part4_or_appendix.tex
\section{Training Details}
\subsection{Training Infrastructure}\label{training_details}
\begin{table}[h]
    \caption{Training configurations and hyperparameters for Heter-MARS$^2$.}
    \label{tab:training-config}
    \centering
\small
\begin{tabularx}{\textwidth}{lX}
\hline
\textbf{Category} & \textbf{Parameter} \\
\hline

\multicolumn{2}{l}{\textbf{Model \& Workflow Setup}} \\
\hline
MCTS Nodes & 16 \\
Max Prompt Length & 4096 \\
Max Generation Length & 32768 \\
Eval Generation Length & 32768 \\
Max Sequence Length & 40000 \\
Overlong Buffer Length & 2048 \\

\hline
\multicolumn{2}{l}{\textbf{Cluster Configuration}} \\
\hline
Reference Model Nodes & 1 node × 8 GPUs ( per model ) \\
Actor Nodes & 1 node × 8 GPUs ( per model ) \\
VLLM Engines & 8 ( per model ) \\
Tensor Parallel Size & 1 \\
VLLM GPU Memory Utilization & 0.85 \\

\hline
\multicolumn{2}{l}{\textbf{Training Hyperparameters}} \\
\hline
Training Batch Size & 256 \\
Micro Train Batch Size & 1 \\
Rollout Batch Size & 512 \\
Micro Rollout Batch Size & 1 \\
Number of Samples per Prompt & 1 \\
Learning Rate (Actor) & 1e-6 \\
Learning Rate (Critic) & 9e-6 \\
Initial KL Coefficient & 1e-3 \\
Gamma & 1.0 \\
KL Estimator & k3 \\
Zero Stage & 3 \\
Precision & bfloat16 \\

\hline
\multicolumn{2}{l}{\textbf{RL / PPO Settings}} \\
\hline
Use KL Loss & Yes \\
Normalize Reward & Yes \\
Use Packing Samples & Yes \\
Dynamic Filtering Reward Range & [0, 1] \\
Dynamic Filtering for Agents & Enabled \\
IS Correction (vLLM) & Enabled \\
IS Truncated Threshold & 2 \\

\hline
\multicolumn{2}{l}{\textbf{Inference Settings}} \\
\hline
Temperature & 1.0 \\
Top-p & 1.0 \\
Eval Temperature & 1.0 \\
Eval Samples per Prompt & 1 \\

\hline
\end{tabularx}
\end{table}

% \subsection{Training Algorithm}\label{MAMCTS-GRPO_algo}

% \begin{algorithm}
% \caption{\textbf{MARS$^2$} Training Procedure}
% \begin{algorithmic}[1]
% \Require Training prompt set $\mathcal{D}$; agent set $\{A_1,\ldots,A_K\}$; buffer thresholds $\{B_1,\ldots,B_K\}$
% \Ensure Updated agent parameters $\{\theta_1,\ldots,\theta_K\}$
% \State \textbf{Initialization}
% \State Initialize agent policies $\{\pi_{\theta_k}\}_{k=1}^{K}$
% \State Initialize buffers $\{\mathcal{B}_k\}_{k=1}^{K}$
% \For{each training iteration}
%     \State Sample a batch of prompts $q \sim \mathcal{D}$
%     \For{each prompt $q$}
%         \State Perform rollout to obtain trajectory tree $\mathcal{T}$
%         \State Compute rewards and advantages for all nodes in $\mathcal{T}$
%     \EndFor
%     \State Build optimization group from Monte Carlo tree
%     \For{each agent indexed by $k$}
%         \State Append data for agent $A_k$ to buffer $\mathcal{B}_k$
%     \EndFor
%     \For{each agent indexed by $k$}
%         \If{$|\mathcal{B}_k| \ge B_k$}
%             \State Compute \textbf{MARS$^2$} loss $\mathcal{L}_k$

%             \State Update parameters: $\theta_k \leftarrow \theta_k - \eta \nabla_{\theta_k}\mathcal{L}_k$
%             \State Clear buffer $\mathcal{B}_k$
%         \EndIf
%     \EndFor
% \EndFor
% \end{algorithmic}
% \end{algorithm}

\section{Detailed Experiments of MARS$^2$}
\subsection{Heter-MARS$^2$ and Homo-MARS$^2$}

\begin{table}[H]
    \caption{Comparison of MARS$^2$-T performance before and after training. For brevity, \textbf{DA-8B} denotes the combination of Qwen3-8B and AReaL-boba-2-8B, \textbf{DA-14B} denotes the combination of Qwen3-14B and AReaL-boba-2-14B, and \textbf{TA-14B} represents the three-model ensemble of Qwen3-14B, AReaL-boba-2-14B, and DeepCoder-14B-Preview. }
     % \vspace{\baselineskip}
    \label{rl-tts}
    \centering
    \small
    \setlength{\tabcolsep}{5.5pt}
    \begin{tabular}{cccccc}
    \toprule%第一道横线
    \textbf{single/multi agent} & \textbf{Pass@1}   & \multicolumn{3}{c}{\textbf{Pass@1(MCTS)}} & \textbf{Pass@N} \\
    \hline
     &  & budget=16 & budget=37 & budget=60 & budget=60 \\
    \midrule%第三道横线 
    Qwen-8B (w/o training) & 50.3 & 56.0 & 56.6 & 54.3 & 68.6\\
    Qwen-8B (w/ training) & 55.4 & 56.6 & 56.6 & 60.6 & 71.4\\
    AReaL-8B (w/o training) & 51.0 & 54.9 & 54.9 & 56.0 & 66.9\\
    AReaL-8B (w/ training) & 55.4 & 57.7 & 59.4 & 58.3 & 69.7\\
    
    \bottomrule
    DA-8B (w/o training) & 50.3/51.4 & 56.0 & 55.4 & 57.2 & 69.7\\
    DA-8B (w/ training) & 58.3/54.9 & 64.0 & 61.7 & 61.7 & 75.4\\
    
    \bottomrule
    Qwen-14B (w/o training) & 56.0 & 61.1 & 61.7 & 63.4 & 75.4\\
    Qwen-14B (w/ training) & 61.1 & 64.0 & 66.3 & 65.1 & 78.9\\
    AReaL-14B (w/o training) & 58.4 & 61.1 & 65.7 & 64.6 & 74.3\\
    AReaL-14B (w/ training) & 62.3 & 66.3 & 72.6 & 68.0 & 81.1\\
    DeepCoder-14B (w/o training) & 50.8 & 56.0 & 57.1 & 57.1 & 67.4\\
    DeepCoder-14B (w/ training) & 54.3 & 58.3 & 56.6 & 60.6 & 70.9\\

    \bottomrule
    DA-14B (w/o training) & 56.0/58.4 & 61.7 & 60.0 & 62.9 & 74.9\\
    DA-14B (w/ training) & 61.7/64.6 & 67.7 & 68.0 & 68.9 & 79.4\\
    TA-14B (w/o training) & 56.0/58.4/50.8 & 61.1 & 62.3 & 61.1 & 76.6\\
    TA-14B (w/ training) & 56.6/64/52 & 68.0 & 65.7 & 66.9 & 80.0\\

    \bottomrule    \end{tabular}
    \par\scriptsize\vspace{0.2em}
    %\par\tiny\vspace{0.1em}
\end{table}

\subsection{Test-time Search Efficiency}
% 首先我们对训练前的模型进行了详细的TTS
We conduct a comprehensive evaluation of MARS$^2$-T+ to validate its search efficiency and scalability before training.
\begin{itemize}
    \item \textbf{System Performance.} We evaluate MARS$^2$-T+ algorithm by comparing its performance with LLM-based single-agent’s initial performance as well as with standard Best-of-N algorithm. Empirical results in Table \ref{test-time scaling} indicate that MARS$^2$-T+ algorithm achieves consistently superior performance. These results collectively demonstrate the efficiency of the MARS$^2$-T+ algorithm.
    
    \item \textbf{Scaling Trend (Compute \& Agents).} We identify scaling trends across two dimensions. Regarding generation budgets, performance improves steadily as more budget is allocated, systematically capitalizing on additional computational resources until stabilizing at a budget of XX. Regarding agent scaling, multi-agent MCTS consistently outperforms the strongest single-agent setup. The performance continues to improve as the number of agents increases, beyond which the system saturates. This highlights the effectiveness of multi-agent collaboration in enhancing exploration efficiency.
    
    \item \textbf{Ablation Studies.} We validate the impact of reward model and error message. The integration of reward model for node selection consistently outperforms the baseline strategy that selects nodes in reverse generation order, proving its effectiveness in prioritizing high-quality reasoning trajectories. Furthermore, augmenting MARS$^2$-T+ with error message feedback yields superior results compared to the version without feedback, indicating that leveraging error signals can steer search and enhance overall performance.
\end{itemize}
% ABMCTS的限制

\begin{table}[H]
    \caption{\textbf{Performance of single-agent and multi-agent test-time scaling.} Q-8B denotes Qwen3-8B; A-8B denotes AReaL-boba-2-8B; DS-Q-8B denotes DeepSeek-R1-0528-Qwen3-8B; DS-L-8B denotes DeepSeek-R1-Distill-Llama-8B; Q-14B denotes Qwen3-14B; A-14B denotes AReaL-boba-2-14B; DC-14B denotes Deepcoder-14B-preview ; DS-Q-14B denotes DeepSeek-R1-Distill-Qwen-14B.}
    \label{test-time scaling}
    \centering
    \small
    \setlength{\tabcolsep}{2pt}
    \begin{tabular}{ccccccc}
    \toprule
    \textbf{\makecell{single/multi \\agent}} &
    \textbf{\makecell{initial\\performance}} &
    \textbf{\makecell{Best of N\\(Pass@1)}} &
    \multicolumn{3}{c}{\textbf{\makecell{System Performance\\(Pass@1)}}} &
    \textbf{\makecell{System Performance\\(Pass@K)}} \\
    \cline{4-6}
     & & N=150 & budget=60 & budget=100 & budget=150 & budget=150 \\
    \midrule
    Q-8B  & 0.503 & 0.554 & 0.543 & 0.576 & 0.570 & 0.703\\
    A-8B  & 0.510 & 0.526 & 0.560 & 0.556 & 0.538 & 0.675\\
    \midrule
    DA-8B & - & - & 0.572 & 0.543 & 0.560 & 0.714 \\
    \midrule
    Q-14B & 0.560 & 0.600 & 0.634 & 0.636 & 0.630 & 0.766\\
    A-14B & 0.584 & 0.594 & 0.646 & 0.652 & 0.674 & 0.760\\
    DC-14B & 0.508 & 0.514 & 0.571 & 0.539 & 0.536 & 0.737\\
    DS-Q-14B & 0.530 & 0.566 & 0.571 & 0.571 & 0.571 & 0.680\\
    \midrule
    DA-14B & - & - & 0.629 & 0.623 & 0.646 & 0.760\\
    TA-14B & - & - & 0.611 & 0.608 & 0.615 & 0.771\\
    QA-14B & - & - & 0.623 & 0.657 & 0.628 & 0.789\\
    \bottomrule
    \end{tabular}
    \par\scriptsize\vspace{0.2em}
\end{table}
\subsection{The Limitations of AB-MCTS}
\label{sec:appendix_mamcts}
% 我们观察到ABMCTS在解决复杂代码问题时仍然存在一些限制，这些限制在面向多智能体时显得尤为重要

% AB-MCTS 将失败的节点仅视为提供二值反馈（例如“通过/未通过”），这限制了模型从错误中学习的能力。因此，它未能利用更丰富的诊断信号——例如输入输出不匹配或运行时错误日志——来引导更高效的迭代式优化过程。
\paragraph{Insufficient error utilization.}
AB-MCTS treats failed nodes as providing only binary feedback (e.g., pass/fail), which limits the model’s ability to learn from mistakes. Consequently, it fails to exploit richer diagnostic signals—such as input–output mismatches or runtime error traces—that could otherwise guide more effective iterative refinement. We provide the original refinement prompt template used in AB-MCTS to illustrate this limitation. The template for our enhanced error-feedback mechanism is presented in Appendix~\ref{sec:error_feedback}.

% 我们给出原有方法的refine的prompt模板以说明这种情况，我们在附录中提供了错误信息集成的模板以解决这些限制见附录
\paragraph{Limited depth for refinement. }
% 我们统计了ABMCTS-A在livecodebench-v6测试集上的节点分布，我们选用总的175个任务，每个任务扩展100个节点。As detailed in Tables，我们发现ABMCTS倾向于生成宽但是更浅的树，95%的节点分布在前三层，这一倾向在多智能体场景上更加明显。

% 模型    第一层     第二层     第三层     第四层     大于等于第五层
% Qwen3-8B  46.514 36.377 13.234 3.291 0.511
% Areal-8B  41.143 42.297 13.543 2.503 0.515
% DeepSeek-R1-0528-Qwen3-8B 54.194 32.149 11.417 2.023 0.217
% Qwen3-8B\&Areal-8B 45.326 40.234 12.560 1.686 0.191
% Qwen3-8B\&Areal-8B\&DeepSeek-R1-0528-Qwen3-8B 66.589 28.743 4.314 0.326 0.029
The algorithm tends to generate wide but shallow trees, which severely impedes the deep iterative refinement required to solve complex code generation problems. We evaluate AB-MCTS and its multi-agent variants on the \texttt{LiveCodeBench-v6} benchmark, using 175 tasks with 100 node expansions per task. As detailed in Table~\ref{tab:depth_distribution}, AB-MCTS exhibits a strong bias toward shallow exploration: over 95\% of nodes are concentrated within the first three levels. This tendency becomes even more pronounced in multi-agent settings, indicating that the algorithm does not fully exploit the collaborative advantages necessary for solving complex tasks.

\begin{table}[h]
\centering
\small
\caption{Node depth distribution (\%) on \texttt{LiveCodeBench-v6} with 175 tasks (100 node expansions per task). Most nodes lie within the first three levels, indicating shallow exploration behavior.
\textbf{Q}, \textbf{A}, and \textbf{D} denote \textit{Qwen3-8B}, \textit{Areal-8B}, and \textit{DeepSeek-R1-0528-Qwen3-8B}, respectively. 
\textbf{Q+A} and \textbf{Q+A+D} represent 2-agent and 3-agent configurations.}
\label{tab:depth_distribution}
\begin{tabular}{lccccc}
\toprule
\textbf{Model} & \textbf{1st layer} & \textbf{2nd layer} & \textbf{3rd layer} & \textbf{4th layer} & \textbf{$\ge$5th layer} \\
\midrule
Qwen3-8B & 46.51 & 36.38 & 13.23 & 3.29 & 0.51 \\
Areal-8B & 41.14 & 42.30 & 13.54 & 2.50 & 0.52 \\
DeepSeek-R1-0528-Qwen3-8B & 54.19 & 32.15 & 11.42 & 2.02 & 0.22 \\
2 model(Q+A) & 45.33 & 40.23 & 12.56 & 1.69 & 0.19 \\
3 model(Q+A+D) & 66.59 & 28.74 & 4.31 & 0.33 & 0.03 \\
\bottomrule
\end{tabular}
\end{table}

%为了避免任务难度对于节点分布的影响，我们选取了难度为困难的样本作为一个子任务集，统计模型在该任务子集上的节点分布，detailed in Tables 我们发现尽管相比于简单任务，算法展示出了向深扩展的倾向，但是绝大多数的节点仍然分布在前三层，多智能体仍然显示出拓展更浅的倾向性。
% 模型    第一层     第二层     第三层     第四层     大于等于第五层
% Qwen3-8B  35.425, 38.8375, 19.025, 5.6625, 1.05
% Areal-8B  31.9625, 47.05, 16.575, 3.5, 0.9125
% DeepSeek-R1-0528-Qwen3-8B 38.825, 40.6125, 17.0, 3.1625, 0.4
% Qwen3-8B\&Areal-8B 35.5125, 43.55, 17.9, 2.7375, 0.3
% Qwen3-8B\&Areal-8B\&DeepSeek-R1-0528-Qwen3-8B 62.525, 31.875, 5.15, 0.4, 0.05

To further control for task difficulty, we analyze a subset containing only \textit{hard-level} samples. The depth distribution (Table~\ref{tab:depth_difficult}) reveals a similar pattern: despite a slight shift toward deeper exploration, most nodes still cluster in the first three layers, especially in multi-agent variants.

\begin{table}[h]
\centering
\caption{Node depth distribution (\%) on the hard subset of \texttt{LiveCodeBench-v6}. 
The shallow-search tendency remains evident in multi-agent variants.}
\label{tab:depth_difficult}
\small
\begin{tabular}{lccccc}
\toprule
\textbf{Model} & \textbf{1st layer} & \textbf{2nd layer} & \textbf{3rd layer} & \textbf{4th layer} & \textbf{$\ge$5th layer} \\
\midrule
Qwen3-8B & 35.43 & 38.84 & 19.03 & 5.66 & 1.05 \\
Areal-8B & 31.96 & 47.05 & 16.58 & 3.50 & 0.91 \\
DeepSeek-R1-0528-Qwen3-8B & 38.83 & 40.61 & 17.00 & 3.16 & 0.40 \\
2 model(Q+A) & 35.51 & 43.55 & 17.90 & 2.74 & 0.30 \\
3 model(Q+A+D) & 62.53 & 31.88 & 5.15 & 0.40 & 0.05 \\
\bottomrule
\end{tabular}
\end{table}

% 这表示现有的算法并没有充分利用到多智能体之间的协同优势——限制了解决复杂任务所需的协同反复迭代
This shallow search behavior suggests that AB-MCTS has not fully exploited the potential of multi-agent collaboration, thereby limiting its ability to achieve deeper iterative refinement on challenging code generation tasks.

\subsection{Prompt of MARS$^2$-T+}
\begin{tcolorbox}[title=Vanilla-TTS Prompt]
\label{box:ab-mcts-prompt}
You are an expert Python programmer. You will be given a question (problem specification) and will generate a correct Python program that matches the specification and passes all tests. Here's your last qwestion and your answer: 

Question:\{\}

Your previous code:\{\}

Result: Wrong

Summary

Your solution is correct for 0 problems among 1, please re-implement your code. In addition to passing the test cases, try to ensure that your code can handle various input cases as much as possible, including edge cases and exceptions.

Answer: (use the provided format with backticks)
\end{tcolorbox}

\section{Reward Model Details}
\label{sec:appendix_rm}
% 1. 数据构造
\subsection{Training Data Construction}
\label{sec:appendix_rm_data}
\paragraph{Data Source and Generation.}
We utilize the \texttt{DeepCoder} dataset as the problem pool. For each problem $x$, we employ a simple parallel sampling method. Specifically, we use each policy model $\mathcal{A}_j$ from our pool (e.g., \texttt{Qwen3-8B}, \texttt{areal-boba-2-8B}, etc.) to independently generate $N=8$ candidate solutions $y$ for the same prompt. 
% 中文注释：
% 我们利用 DeepCoder 数据集作为问题池。
% 对于每个问题 x，我们采用一个简单的并行采样方法。具体来说，我们使用我们模型池中的*每个*策略模型 A_j (例如 Qwen3-8B, areal-boba-2-8B 等)为同一个 prompt 独立生成 N=8 个候选解法 y。

The collected nodes from all models are then aggregated. 
% 中文注释：
% 从所有模型收集到的节点随后被聚合。

\paragraph{Ground-Truth Labeling.}
Each generated solution $y$ is executed against the problem's private test cases to obtain the ground-truth environment reward $r \in \{0, 1\}$, where $r=1$ signifies passing all test cases and $r=0$ otherwise.

\paragraph{Data Formatting.}
From this labeled corpus, we derive two distinct training formats, which correspond to two complementary training objectives:

% 错误信息集成prompt
\begin{tcolorbox}[title=MARS$^2$-T+Prompt]
\label{sec:error_feedback}
You are an expert Python programmer. You will be given a question (problem specification) and will generate a correct Python program that matches the specification and passes all tests. Here's your last question and your answer: 

Question:{}

Format: Read the inputs from stdin solve the problem and write the answer to stdout (do not directly test on the sample inputs). Enclose your code within delimiters as follows. Ensure that when the python program runs, it reads the inputs, runs the algorithm and writes output to STDOUT.

```python

YOUR CODE HERE

```

Your previous code:{}

Next, please re-implement your code. In addition to passing the test cases, try to ensure that your code can handle various input cases as much as possible, including edge cases and exceptions.

TASK: Deep Code Repair and Generalization

Here is the test report from the last execution:

Execution Summary:
- Passed {} out of {} test cases ({} pass rate).

Detailed Failure Analysis:

--- Failure on Test Case 1 ---

Reason: Wrong Answer

Input:{}

Your code's output: {}

Expected output: {}

WARNING: Do not write special-case code (e.g., using an `if` statement for a specific input) just to fix the failures listed above. This "patching" approach will fail on unseen test cases.

Your Goal: Treat the failed cases as clues to find and fix the fundamental algorithmic or logical flaw in the code.

Please follow this thinking framework to provide a generalized solution:

1.  Pattern Analysis:

Examine all failed test cases. Are they edge cases (e.g., empty lists, zero, max values)? Data type issues (e.g., integer vs. float)? Or inputs with a specific pattern?

Compare successful vs. failed cases and ask: "Under what conditions does my algorithm break?"

2.  Root Cause Deduction:

Based on the pattern, pinpoint the specific defect in the code. Is it a wrong loop condition? Improper variable initialization? A missing base case in recursion? Or a flaw in the overall algorithm logic?

3.  General Fix Strategy:

Formulate a plan that corrects the root cause. This fix must naturally pass all known failed cases and generalize to similar, unseen ones.

4.  Provide Final Code:

Based on your general fix strategy, provide the complete and immediately runnable corrected code.

Do not add any explanations, comments, or dialogue before or after the code.

Answer: (use the provided format with backticks)

\end{tcolorbox}

% 每个生成的解法 y 通过私有测试用例执行以获得真值奖励 r 属于 {0, 1}。
% 从这个已标注的语料库中，我们派生出两种不同针对不同训练目标的的训练格式：
\begin{itemize}[leftmargin=*]
    \item \textbf{Pointwise Data:} Each tuple is formatted as $(x, y, r)$. This data is used to train the RM to predict the absolute correctness (i.e., pass probability) of a solution.
    % 逐点数据：每个元组格式化为 (x, y, r)。该数据用于训练RM预测解法的绝对正确性（即通过概率）。
    
    \item \textbf{Pairwise Data:} We generate preference pairs $(x, y_w, y_l)$, where $y_w$ ("winner") is a preferred solution over $y_l$ ("loser"). These pairs are primarily constructed by sampling solutions with different ground-truth rewards (e.g., $r(y_w)=1$ and $r(y_l)=0$) for the same problem $x$. This format is crucial for training the model on relative ranking.
    % 偏好对：我们生成偏好对 (x, y_w, y_l)，其中 y_w ("赢家") 是比 y_l ("输家") 更优的解法。
    % 这些对主要是通过为同一问题 x 采样具有不同真值奖励（例如 r(y_w)=1 和 r(y_l)=0）的解法来构建的。
    % 这种格式对于训练模型的相对排序能力至关重要。
\end{itemize}

\subsection{Training Objectives}
\label{sec:appendix_rm_loss}
We initialize our RM $R_\theta$ from the \texttt{Skywork-Reward-V2-Qwen3-8B} checkpoint. The model takes $(x, y)$ as input and outputs a scalar score $s = R_\theta(x, y)$. We train this model using two complementary objectives:
% 我们从 Skywork-Reward-V2-Qwen3-8B 检查点初始化 RM。 模型输入 (x, y) 并输出标量分数 s。我们使用两种互补的目标训练模型：
\paragraph{Mean Squared Error (MSE) Loss.}
As defined in the main paper (Equation \ref{eq:mse_loss}), this objective trains the RM as a fine-grained predictor of pass probability. We apply a sigmoid function $\sigma(\cdot)$ to the output score and optimize against the binary ground-truth label $r$:
%正如主论文中定义的（公式 \ref{eq:mse_loss}），此目标将RM训练为通过率的细粒度预测器。 我们对输出分数应用 sigmoid 函数，并使用MSE损失 相对真值标签 r 进行优化：
\begin{equation}
\label{eq:appendix_mse_loss}
L_{MSE} = (\sigma(R_\theta(x, y)) - r)^2
\end{equation}

\paragraph{Bradley-Terry (BT) Pairwise Loss.}
To explicitly train the RM on relative preferences, we utilize the Bradley-Terry model. For a given preference pair $(x, y_w, y_l)$, the loss function encourages the score of the winning solution $y_w$ to be higher than that of the losing solution $y_l$:
% 中文注释：
% 为了在相对偏好上明确训练RM，我们利用了 Bradley-Terry 模型。对于给定的偏好对 (x, y_w, y_l)，损失函数鼓励赢家解法 y_w 的分数高于输家解法 y_l：
\begin{equation}
\label{eq:appendix_bt_loss}
L_{BT} = -\log(\sigma(R_\theta(x, y_w) - R_\theta(x, y_l)))
\end{equation}
In our experiments, we compare the performance of RMs trained with $L_{MSE}$ versus those trained with $L_{BT}$ on our benchmark.
% 在我们的实验中，我们比较了使用 L_MSE 训练的 RM 与使用 L_BT 训练的 RM 的性能。

\subsection{Benchmark Construction for RM Evaluation}
\paragraph{Benchmark Source.} The benchmarks are derived from tasks in \texttt{livecodebench\_v6} (175 tasks), which are entirely separate from our \texttt{DeepCoder}-based training data.

\paragraph{Generation Process.} To generate the dataset, we employed two distinct groups of reasoning models, categorized by parameter size. The first group consists of 14B-parameter models: \texttt{Qwen3-14B}, \texttt{areal-boba-2-14B}, and \texttt{deepcoder-14B-preview}. The second group comprises 8B-parameter models: \texttt{Qwen3-8B} and \texttt{areal-boba-2-8B}. We utilized these models to run AB-MCTS on the held-out tasks, collecting a diverse set of solutions and their corresponding ground-truth outcomes.

\paragraph{Benchmark Instances.}
\label{sec:appendix_rm_bench}
We constructed two distinct sets of benchmarks to evaluate performance across different training stages. Both sets encompass solutions generated by the 8B and 14B model groups. For these extracted benchmarks, we specifically ensured a balanced distribution between data with a score of 1 and a score of 0.
% 我们根据生成模型的训练阶段构建了两组不同的基准。两组都包含由 8B 和 14B 模型组生成的解。此外，对于这些提取的基准，我们特别确保了分数为 1 和分数为 0 的数据之间的平衡。

\begin{itemize}[leftmargin=*]
    \item \textbf{Pre-train Benchmark (150 nodes):} Constructed using solutions generated by the models prior to training. This dataset collects the top 150 nodes per task.
    % 训练前基准：使用模型在训练前生成的解构建。该数据集收集每个任务的前 150 个节点。
    
    \item \textbf{Post-train Benchmark (60 nodes):} Created using solutions generated by the models after training. This dataset collects the top 60 nodes per task.
    % 训练后基准：使用模型在训练后生成的解构建。该数据集收集每个任务的前 60 个节点。
\end{itemize}

\paragraph{Evaluation Metrics Definition.}
To comprehensively evaluate the Reward Model's ranking capability and robustness against score distribution shifts, we employ three key metrics on the pointwise benchmark data:
% 为了全面评估奖励模型的排序能力以及对分数分布偏移的鲁棒性，我们在逐点基准数据上采用了三个关键指标：

\begin{itemize}[leftmargin=*]
    \item \textbf{AUC-ROC:} The Area Under the Receiver Operating Characteristic curve. This metric evaluates the model's ability to rank positive samples higher than negative ones across all possible decision thresholds, serving as a calibration-independent measure of discriminative power.
    % AUC-ROC：ROC 曲线下的面积。该指标评估模型在所有可能的决策阈值下将正样本排在负样本之前的能力，是一种独立于校准之外的判别能力衡量标准。

    \item \textbf{Spearman Correlation:} The rank-order correlation coefficient between the predicted reward scores and the ground truth labels. It assesses the monotonicity of the predictions, ensuring that better responses consistently receive higher scores regardless of the absolute value range.
    % 斯皮尔曼相关系数：预测奖励分数与真实标签之间的秩相关系数。它评估预测的单调性，确保更好的回复无论绝对数值范围如何，都能获得更高分数。

    \item \textbf{Adaptive Accuracy:} The binary classification accuracy calculated using the median of the predicted logits as the dynamic decision threshold (instead of a fixed threshold like 0 or 0.5). This metric mitigates the impact of uncalibrated score distributions (e.g., negative bias) and focuses on the model's intrinsic ability to separate positive and negative samples.
    % 自适应准确率：使用预测 Logit 的中位数作为动态决策阈值（而不是固定的 0 或 0.5）计算的二分类准确率。该指标减轻了未校准分数分布（例如负向偏差）的影响，专注于模型区分正负样本的内在能力。
\end{itemize}

\section{Diversity Analysis of Our TTS-Enhanced Method}
\label{sec:appendix_diversity_marts}

% 我们在5.2中主要分析了ReMASS训练方法的多样性指标，在本section中，我们将分析我们提出的tts增强方法的多样性表现。
In Section 5.2, we primarily analyzed the diversity metrics of the MARS$^2$ training method. In this section, we evaluate the diversity performance of our proposed TTS-enhanced method.

\begin{figure}[H]
    \centering
    % 第一行
    \begin{subfigure}[b]{0.32\textwidth}
        \centering
        \includegraphics[width=\linewidth]{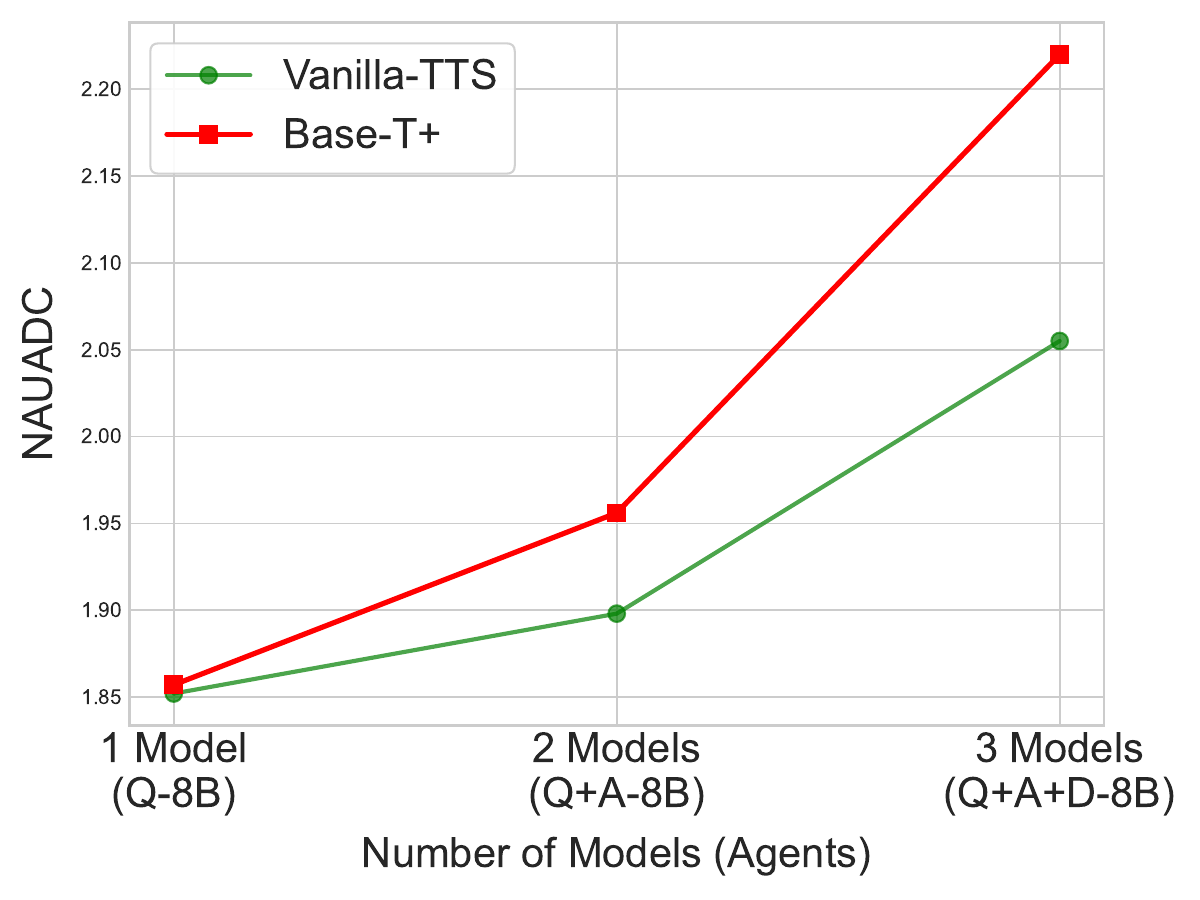}
        \subcaption{Comparison of NAUADC Diversity Trends with Increasing Model Counts on the 8B Model Series}
        \label{fig:nauadc_marts8B}
    \end{subfigure}
    \hfill
    \begin{subfigure}[b]{0.32\textwidth}
        \centering
        \includegraphics[width=\linewidth]{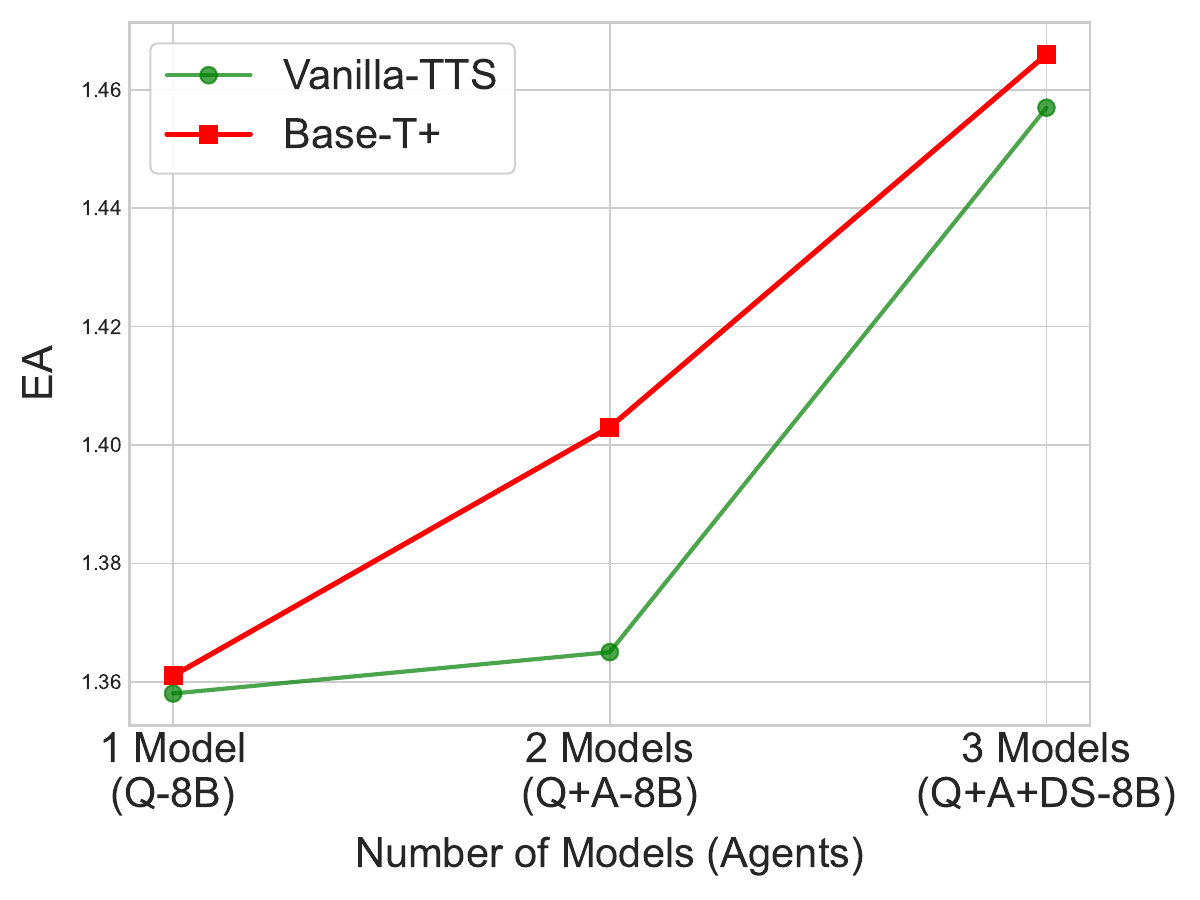}
        \subcaption{Comparison of EA with Increasing Model Counts on the 8B Model Series}
        \label{fig:ea_marts8B}
    \end{subfigure}
    \hfill
    \begin{subfigure}[b]{0.32\textwidth}
        \centering
        \includegraphics[width=\linewidth]{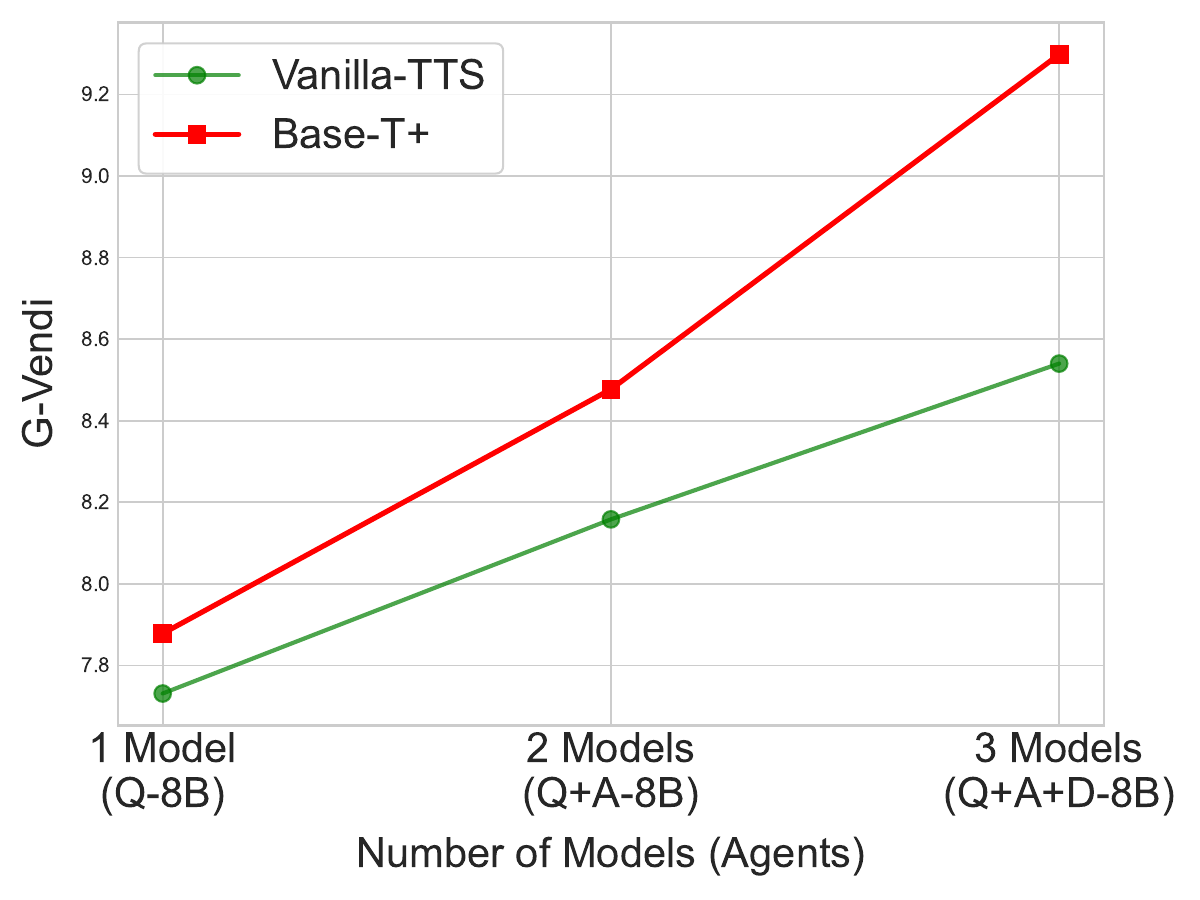}
        \subcaption{Comparison of G-Vendi Metrics with Increasing Model Counts on the 8B Model Series}
        \label{fig:vendi_marts8B}
    \end{subfigure}
    \caption{Analysis of Diversity metrics Trends with Increasing Model Counts across 8B and 14B Model Series.}
    \label{fig:diversity_metric_marts}
\end{figure}

% 我们将节点budgets从60扩展到100个节点以呈现方法的潜力，我们筛选了AB-MCTS和MARTS Method(ref 3.3.1)能够共同完成的104个任务，计算其多样性指标，我们得到了如下的图像。\ref{fig:diversity_metric_marts}
We extend the node budget from 60 to 100 to better demonstrate the potential of the approach. We select 104 tasks that can be successfully completed by both AB-MCTS and the MARS$^2$ method (see Section~3.3.1) and compute their diversity metrics. The results are visualized in Figure~\ref{fig:diversity_metric_marts}.

% 随着模型数量的增加，NAUADC,EA,G-Vendi指标均增加，这再次验证了我们的主要结论：Algorithmic diversity and distributional uniformity improve as the number of models increases. 此外，我们提出的TTS-增强方法 MARTS在所有的多样性指标上都高于AB-MCTS（特别是在多模型设置上），这说明MARTS能够同时提升算法多样性，算法分布均匀，以及策略多样性。证明了我们方法的有效。
As the number of models increases, the NAUADC, EA, and G-Vendi metrics all improve, further corroborating our main conclusion: \textbf{algorithmic diversity and distributional uniformity increase with the number of models}. Moreover, our proposed TTS-enhanced method, MARS$^2$, consistently outperforms AB-MCTS across all diversity metrics, particularly in multi-model settings. This demonstrates that MARS$^2$ effectively enhances algorithmic diversity, distributional uniformity, and strategic diversity.

%为了进一步对我们方法的多样性增加机理，我们将任务根据难度进行区分，将其分成43个简单任务，35个中等任务，以及26个困难任务。我们分别在这些任务上计算8B系列模型的多样性指标，结果如表xx所示。\ref{tab:diversity_results}
To further investigate the mechanism by which our method enhances diversity, we categorize tasks by difficulty into 43 easy, 35 medium, and 26 hard tasks. We compute diversity metrics for the 8B-class models on each category separately, with results shown in Table~\ref{tab:diversity_marts}.

\begin{table}[H]
\centering
\small
\caption{Detailed diversity metric statistics for 8B series models.}
\label{tab:diversity_marts}
\setlength{\tabcolsep}{1.5pt}
\renewcommand{\arraystretch}{1.00}
\begin{tabular}{l c ccc ccc ccc ccc ccc}
\toprule
\multirow{2}{*}{Method} & \multirow{2}{*}{Pass@K}
& \multicolumn{3}{c}{AEC}
& \multicolumn{3}{c}{DA@K}
& \multicolumn{3}{c}{EA}
& \multicolumn{3}{c}{NAUADC}
& \multicolumn{3}{c}{G-Vendi} \\
\cmidrule(lr){3-5} \cmidrule(lr){6-8} \cmidrule(lr){9-11} \cmidrule(lr){12-14} \cmidrule(lr){15-17}
& & E. & M. & H.
  & E. & M. & H.
  & E. & M. & H.
  & E. & M. & H.
  & E. & M. & H. \\
\midrule
\multicolumn{17}{l}{\textbf{AB-MCTS}} \\
Q-8B            & 0.725 & 1.09 & 1.09 & 1.00 & 1.30 & 1.50 & 1.71 & 1.23 & 1.39 & 1.53 & 1.50 & 1.95 & 2.31 & 8.67 & 7.60 & 6.34 \\
A-8B            & 0.691 & 1.09 & 1.06 & 1.00 & 1.19 & 1.60 & 1.72 & 1.14 & 1.44 & 1.54 & 1.33 & 2.07 & 2.29 & 9.39 & 7.50 & 5.96 \\
Q-8B+A-8B       & 0.737 & 1.23 & 1.40 & 1.50 & 1.32 & 1.55 & 1.76 & 1.24 & 1.38 & 1.55 & 1.58 & 2.07 & 2.19 & 9.43 & 8.07 & 6.17 \\
Q-8B+A-8B+DS-8B & 0.697 & 1.47 & 1.94 & 1.96 & 1.44 & 1.58 & 1.91 & 1.33 & 1.45 & 1.68 & 1.74 & 2.12 & 2.48 & 12.20 & 6.93 & 4.66 \\
\midrule
\multicolumn{17}{l}{\textbf{MARS$^2$}} \\
Q-8B                 & 0.742 & 1.14 & 1.09 & 1.19 & 1.33 & 1.51 & 1.73 & 1.24 & 1.36 & 1.57 & 1.51 & 1.96 & 2.29 & 8.78 & 7.66 & 6.67 \\
Q-8B+A-8B            & 0.754 & 1.07 & 1.09 & 1.23 & 1.30 & 1.61 & 1.83 & 1.22 & 1.45 & 1.64 & 1.52 & 2.09 & 2.50 & 10.05 & 7.99 & 6.54 \\
Q-8B+A-8B+DS-8B  & 0.737 & 1.21 & 1.77 & 1.92 & 1.37 & 1.67 & 2.05 & 1.27 & 1.48 & 1.77 & 1.72 & 2.37 & 2.84 & 12.79 & 7.89 & 5.42 \\

\bottomrule
\end{tabular}
\end{table}
Based on this table, we obtain several intriguing findings.

% MARTS 在高难度任务上显著增强了算法多样性： 与 AB-MCTS 相比，MARTS 方法在多样性指标（如 DA@K, EA, NAUADC）上表现更佳，这种优势在 Medium（中等） 和 Hard（困难） 任务中尤为明显。例如，在三模型集成（Q+A+DS）设置下，MARTS 在 Hard 难度的 NAUADC 指标上达到了 2.84，显著高于 AB-MCTS 的 2.48。这表明 MARTS 的多智能体协作机制能有效激发模型在处理复杂问题时的发散性思维，避免陷入单一的错误路径。
\paragraph{MARS$^2$ significantly enhances algorithmic diversity on high-difficulty tasks.} Compared to AB-MCTS, the MARS$^2$ method achieves consistently higher scores on diversity metrics such as DA@K, EA, and NAUADC, with the improvement most pronounced on medium and hard tasks. For instance, under the three-model ensemble setting (Q+A+DS), MARS$^2$ attains a NAUADC score of 2.84 on hard tasks, substantially outperforming AB-MCTS's score of 2.48. This indicates that MARS$^2$’s multi-agent collaboration mechanism effectively promotes divergent reasoning when tackling complex problems, helping to avoid convergence to a single erroneous trajectory.

%策略多样性（G-Vendi）呈现“Easy > Medium > Hard”的递减趋势： 无论是在单模型还是多模型设置中，G-Vendi 指标均表现出随着任务难度增加而显著下降的趋势。这可能是因为简单任务（Easy）允许模型生成更为丰富且发散的思维链（Chain-of-Thought），即存在多种不同的推理路径都能通向正确答案；而在面对困难任务时，有效的推理路径往往受到严格的逻辑约束，导致解空间收缩，思维路径的异质性随之降低。
\paragraph{Strategic diversity (G-Vendi) exhibits a clear decreasing trend of ``Easy > Medium > Hard''.}Across both single-model and multi-model settings, the G-Vendi score consistently declines as task difficulty increases. This pattern likely arises because easy tasks permit models to generate richer and more divergent chains of thought—multiple distinct reasoning pathways can lead to the correct answer. In contrast, hard tasks impose stringent logical constraints, narrowing the solution space and thereby reducing the heterogeneity of viable reasoning trajectories.

\paragraph{Cumulative diversity gains from heterogeneous model integration. }Comparing the results of single-model (Q-8B), two-model (Q+A), and three-model (Q+A+DS) setups reveals a steady improvement in all diversity metrics (particularly AEC and NAUADC) as more heterogeneous models are integrated. notably, while the three-model ensemble under MARS$^2$ achieves the highest diversity scores (e.g., a G-Vendi of 12.79 on Easy tasks), its Pass@K (0.737) is slightly lower than that of the two-model ensemble (0.754). This indicates that extreme diversity does not always translate linearly into accuracy improvements; in some cases, introducing highly divergent models may introduce noise, highlighting the need to balance diversity and precision.

%% file: paper.bbl
\begin{thebibliography}{76}
\providecommand{\natexlab}[1]{#1}
\providecommand{\url}[1]{\texttt{#1}}
\expandafter\ifx\csname urlstyle\endcsname\relax
  \providecommand{\doi}[1]{doi: #1}\else
  \providecommand{\doi}{doi: \begingroup \urlstyle{rm}\Url}\fi

\bibitem[Ahmad et~al.(2025)Ahmad, Narenthiran, Majumdar, Ficek, Jain, Huang, Noroozi, and Ginsburg]{ahmad2025opencodereasoning}
Wasi~Uddin Ahmad, Sean Narenthiran, Somshubra Majumdar, Aleksander Ficek, Siddhartha Jain, Jocelyn Huang, Vahid Noroozi, and Boris Ginsburg.
\newblock Opencodereasoning: Advancing data distillation for competitive coding.
\newblock \emph{arXiv preprint arXiv:2504.01943}, 2025.

\bibitem[Besta et~al.(2024)Besta, Blach, Kubicek, Gerstenberger, Podstawski, Gianinazzi, Gajda, Lehmann, Niewiadomski, Nyczyk, et~al.]{besta2024graph}
Maciej Besta, Nils Blach, Ales Kubicek, Robert Gerstenberger, Michal Podstawski, Lukas Gianinazzi, Joanna Gajda, Tomasz Lehmann, Hubert Niewiadomski, Piotr Nyczyk, et~al.
\newblock Graph of thoughts: Solving elaborate problems with large language models.
\newblock In \emph{Proceedings of the AAAI conference on artificial intelligence}, volume~38, pages 17682--17690, 2024.

\bibitem[Brown et~al.(2024)Brown, Juravsky, Ehrlich, Clark, Le, R{\'e}, and Mirhoseini]{brown2024large}
Bradley Brown, Jordan Juravsky, Ryan Ehrlich, Ronald Clark, Quoc~V Le, Christopher R{\'e}, and Azalia Mirhoseini.
\newblock Large language monkeys: Scaling inference compute with repeated sampling.
\newblock \emph{arXiv preprint arXiv:2407.21787}, 2024.

\bibitem[Cao et~al.(2025)Cao, Zhang, Li, Wei, Li, Joty, and Carenini]{cao2025multi2}
Juntai Cao, Xiang Zhang, Raymond Li, Jiaqi Wei, Chuyuan Li, Shafiq Joty, and Giuseppe Carenini.
\newblock Multi2: Multi-agent test-time scalable framework for multi-document processing.
\newblock In \emph{Proceedings of The 5th New Frontiers in Summarization Workshop}, pages 135--156, 2025.

\bibitem[Chan et~al.(2023)Chan, Chen, Su, Yu, Xue, Zhang, Fu, and Liu]{chan2023chateval}
Chi-Min Chan, Weize Chen, Yusheng Su, Jianxuan Yu, Wei Xue, Shanghang Zhang, Jie Fu, and Zhiyuan Liu.
\newblock Chateval: Towards better llm-based evaluators through multi-agent debate.
\newblock \emph{arXiv preprint arXiv:2308.07201}, 2023.

\bibitem[Chen et~al.(2025{\natexlab{a}})Chen, Qin, Liu, Peng, Guan, Wang, Hu, Zhou, Gao, and Che]{chen2025towards}
Qiguang Chen, Libo Qin, Jinhao Liu, Dengyun Peng, Jiannan Guan, Peng Wang, Mengkang Hu, Yuhang Zhou, Te~Gao, and Wanxiang Che.
\newblock Towards reasoning era: A survey of long chain-of-thought for reasoning large language models.
\newblock \emph{arXiv preprint arXiv:2503.09567}, 2025{\natexlab{a}}.

\bibitem[Chen et~al.(2025{\natexlab{b}})Chen, Yan, Sun, Ma, Zhang, Wang, Yin, Yang, and Mao]{chen2025improving}
Yiqun Chen, Lingyong Yan, Weiwei Sun, Xinyu Ma, Yi~Zhang, Shuaiqiang Wang, Dawei Yin, Yiming Yang, and Jiaxin Mao.
\newblock Improving retrieval-augmented generation through multi-agent reinforcement learning.
\newblock \emph{arXiv preprint arXiv:2501.15228}, 2025{\natexlab{b}}.

\bibitem[Chen et~al.(2025{\natexlab{c}})Chen, Zhang, Yan, Wang, Huang, Yin, and Mao]{chen2025mao}
Yiqun Chen, Erhan Zhang, Lingyong Yan, Shuaiqiang Wang, Jizhou Huang, Dawei Yin, and Jiaxin Mao.
\newblock Mao-arag: Multi-agent orchestration for adaptive retrieval-augmented generation.
\newblock \emph{arXiv preprint arXiv:2508.01005}, 2025{\natexlab{c}}.

\bibitem[Chen et~al.(2026)Chen, Yan, Yang, Zhang, Zhao, Wang, Yin, and Mao]{chen2026beyond}
Yiqun Chen, Lingyong Yan, Zixuan Yang, Erhan Zhang, Jiashu Zhao, Shuaiqiang Wang, Dawei Yin, and Jiaxin Mao.
\newblock Beyond monolithic architectures: A multi-agent search and knowledge optimization framework for agentic search.
\newblock \emph{arXiv preprint arXiv:2601.04703}, 2026.

\bibitem[Chen et~al.(2025{\natexlab{d}})Chen, Chen, Meng, Yin, Li, Fan, Wang, Pfister, and Yoon]{chen2025tumix}
Yongchao Chen, Jiefeng Chen, Rui Meng, Ji~Yin, Na~Li, Chuchu Fan, Chi Wang, Tomas Pfister, and Jinsung Yoon.
\newblock Tumix: Multi-agent test-time scaling with tool-use mixture.
\newblock \emph{arXiv preprint arXiv:2510.01279}, 2025{\natexlab{d}}.

\bibitem[Chern et~al.(2024)Chern, Fan, and Liu]{chern2024combating}
Steffi Chern, Zhen Fan, and Andy Liu.
\newblock Combating adversarial attacks with multi-agent debate.
\newblock \emph{arXiv preprint arXiv:2401.05998}, 2024.

\bibitem[El-Kishky et~al.(2025)El-Kishky, Wei, Saraiva, Minaiev, Selsam, Dohan, Song, Lightman, Clavera, Pachocki, et~al.]{el2025competitive}
Ahmed El-Kishky, Alexander Wei, Andre Saraiva, Borys Minaiev, Daniel Selsam, David Dohan, Francis Song, Hunter Lightman, Ignasi Clavera, Jakub Pachocki, et~al.
\newblock Competitive programming with large reasoning models.
\newblock \emph{arXiv preprint arXiv:2502.06807}, 2025.

\bibitem[Ester et~al.(1996)Ester, Kriegel, Sander, Xu, et~al.]{ester1996density}
Martin Ester, Hans-Peter Kriegel, J{\"o}rg Sander, Xiaowei Xu, et~al.
\newblock A density-based algorithm for discovering clusters in large spatial databases with noise.
\newblock In \emph{kdd}, volume~96, pages 226--231, 1996.

\bibitem[Fu et~al.(2025)Fu, Gao, Shen, Zhu, Mei, He, Xu, Wei, Mei, Wang, et~al.]{areal}
Wei Fu, Jiaxuan Gao, Xujie Shen, Chen Zhu, Zhiyu Mei, Chuyi He, Shusheng Xu, Guo Wei, Jun Mei, Jiashu Wang, et~al.
\newblock Areal: A large-scale asynchronous reinforcement learning system for language reasoning, 2025.

\bibitem[Gao et~al.(2025)Gao, Li, Liu, Xie, Zhao, and Xu]{gao2025principled}
Chengqian Gao, Haonan Li, Liu Liu, Zeke Xie, Peilin Zhao, and Zhiqiang Xu.
\newblock Principled data selection for alignment: The hidden risks of difficult examples.
\newblock \emph{arXiv preprint arXiv:2502.09650}, 2025.

\bibitem[Guo et~al.(2025)Guo, Yang, Zhang, Song, Zhang, Xu, Zhu, Ma, Wang, Bi, et~al.]{guo2025deepseek}
Daya Guo, Dejian Yang, Haowei Zhang, Junxiao Song, Ruoyu Zhang, Runxin Xu, Qihao Zhu, Shirong Ma, Peiyi Wang, Xiao Bi, et~al.
\newblock Deepseek-r1: Incentivizing reasoning capability in llms via reinforcement learning.
\newblock \emph{arXiv preprint arXiv:2501.12948}, 2025.

\bibitem[Han et~al.(2025)Han, Hu, Wei, Zhang, Guo, Lu, and Zhang]{han2025joyagents}
Ai~Han, Junxing Hu, Pu~Wei, Zhiqian Zhang, Yuhang Guo, Jiawei Lu, and Zicheng Zhang.
\newblock Joyagents-r1: Joint evolution dynamics for versatile multi-llm agents with reinforcement learning.
\newblock \emph{arXiv preprint arXiv:2506.19846}, 2025.

\bibitem[Hong et~al.(2024)Hong, Zhuge, Chen, Zheng, Cheng, Zhang, Wang, Wang, Yau, Lin, Zhou, Ran, Xiao, Wu, and Schmidhuber]{hong2024metagptmetaprogrammingmultiagent}
Sirui Hong, Mingchen Zhuge, Jiaqi Chen, Xiawu Zheng, Yuheng Cheng, Ceyao Zhang, Jinlin Wang, Zili Wang, Steven Ka~Shing Yau, Zijuan Lin, Liyang Zhou, Chenyu Ran, Lingfeng Xiao, Chenglin Wu, and Jürgen Schmidhuber.
\newblock Metagpt: Meta programming for a multi-agent collaborative framework, 2024.
\newblock URL \url{https://arxiv.org/abs/2308.00352}.

\bibitem[Inoue et~al.(2025)Inoue, Misaki, Imajuku, Kuroki, Nakamura, and Akiba]{inoue2025wider}
Yuichi Inoue, Kou Misaki, Yuki Imajuku, So~Kuroki, Taishi Nakamura, and Takuya Akiba.
\newblock Wider or deeper? scaling llm inference-time compute with adaptive branching tree search.
\newblock \emph{arXiv preprint arXiv:2503.04412}, 2025.

\bibitem[Jain et~al.(2024)Jain, Han, Gu, Li, Yan, Zhang, Wang, Solar-Lezama, Sen, and Stoica]{livecodebench}
Naman Jain, King Han, Alex Gu, Wen-Ding Li, Fanjia Yan, Tianjun Zhang, Sida Wang, Armando Solar-Lezama, Koushik Sen, and Ion Stoica.
\newblock Livecodebench: Holistic and contamination free evaluation of large language models for code.
\newblock \emph{arXiv preprint arXiv:2403.07974}, 2024.

\bibitem[Ji et~al.(2025)Ji, Li, Ye, Wu, Xu, Mo, and Zhang]{ji2025test}
Yixin Ji, Juntao Li, Hai Ye, Kaixin Wu, Jia Xu, Linjian Mo, and Min Zhang.
\newblock Test-time computing: from system-1 thinking to system-2 thinking.
\newblock \emph{arXiv e-prints}, pages arXiv--2501, 2025.

\bibitem[Jin et~al.(2025{\natexlab{a}})Jin, Peng, Zhang, Tang, Metaxas, and Che]{jin2025two}
Can Jin, Hongwu Peng, Qixin Zhang, Yujin Tang, Dimitris~N Metaxas, and Tong Che.
\newblock Two heads are better than one: Test-time scaling of multi-agent collaborative reasoning.
\newblock \emph{arXiv preprint arXiv:2504.09772}, 2025{\natexlab{a}}.

\bibitem[Jin et~al.(2025{\natexlab{b}})Jin, Zhou, Zhang, Peng, Zhang, Pavone, Han, Hong, Che, and Metaxas]{jin2025your}
Can Jin, Yang Zhou, Qixin Zhang, Hongwu Peng, Di~Zhang, Marco Pavone, Ligong Han, Zhang-Wei Hong, Tong Che, and Dimitris~N Metaxas.
\newblock Your reward function for rl is your best prm for search: Unifying rl and search-based tts.
\newblock \emph{arXiv preprint arXiv:2508.14313}, 2025{\natexlab{b}}.

\bibitem[Jung et~al.(2025)Jung, Han, Lu, Hallinan, Acuna, Prabhumoye, Patwary, Shoeybi, Catanzaro, and Choi]{jung2025prismatic}
Jaehun Jung, Seungju Han, Ximing Lu, Skyler Hallinan, David Acuna, Shrimai Prabhumoye, Mostafa Patwary, Mohammad Shoeybi, Bryan Catanzaro, and Yejin Choi.
\newblock Prismatic synthesis: Gradient-based data diversification boosts generalization in llm reasoning.
\newblock \emph{arXiv preprint arXiv:2505.20161}, 2025.

\bibitem[Kryvosheieva et~al.(2025)Kryvosheieva, Sturua, G{\"u}nther, Martens, and Xiao]{kryvosheieva2025efficient}
Daria Kryvosheieva, Saba Sturua, Michael G{\"u}nther, Scott Martens, and Han Xiao.
\newblock Efficient code embeddings from code generation models.
\newblock \emph{arXiv preprint arXiv:2508.21290}, 2025.

\bibitem[Lee et~al.(2025)Lee, Chon, Jang, Lee, and Yu]{lee-etal-2025-diversely}
Seonghyeon Lee, HeeJae Chon, Joonwon Jang, Dongha Lee, and Hwanjo Yu.
\newblock How diversely can language models solve problems? exploring the algorithmic diversity of model-generated code.
\newblock In Christos Christodoulopoulos, Tanmoy Chakraborty, Carolyn Rose, and Violet Peng, editors, \emph{Findings of the Association for Computational Linguistics: EMNLP 2025}, Suzhou, China, November 2025. Association for Computational Linguistics.
\newblock ISBN 979-8-89176-335-7.

\bibitem[Li et~al.(2025{\natexlab{a}})Li, Wang, Gu, Chang, and Peng]{li2025metal}
Bingxuan Li, Yiwei Wang, Jiuxiang Gu, Kai-Wei Chang, and Nanyun Peng.
\newblock Metal: A multi-agent framework for chart generation with test-time scaling.
\newblock \emph{arXiv preprint arXiv:2502.17651}, 2025{\natexlab{a}}.

\bibitem[Li et~al.(2025{\natexlab{b}})Li, Lin, Jiang, Cao, Liu, Zhang, Huang, Chen, Sun, Wang, et~al.]{li2025chain}
Weizhen Li, Jianbo Lin, Zhuosong Jiang, Jingyi Cao, Xinpeng Liu, Jiayu Zhang, Zhenqiang Huang, Qianben Chen, Weichen Sun, Qiexiang Wang, et~al.
\newblock Chain-of-agents: End-to-end agent foundation models via multi-agent distillation and agentic rl.
\newblock \emph{arXiv preprint arXiv:2508.13167}, 2025{\natexlab{b}}.

\bibitem[Li et~al.(2024)Li, Wang, Zeng, Wu, and Yang]{li2024survey}
Xinyi Li, Sai Wang, Siqi Zeng, Yu~Wu, and Yi~Yang.
\newblock A survey on llm-based multi-agent systems: workflow, infrastructure, and challenges.
\newblock \emph{Vicinagearth}, 1\penalty0 (1):\penalty0 9, 2024.

\bibitem[Li et~al.(2025{\natexlab{c}})Li, Zhang, Zhang, Zhang, Liu, Yao, Xu, Zheng, Wang, Chen, et~al.]{li2025system}
Zhong-Zhi Li, Duzhen Zhang, Ming-Liang Zhang, Jiaxin Zhang, Zengyan Liu, Yuxuan Yao, Haotian Xu, Junhao Zheng, Pei-Jie Wang, Xiuyi Chen, et~al.
\newblock From system 1 to system 2: A survey of reasoning large language models.
\newblock \emph{arXiv preprint arXiv:2502.17419}, 2025{\natexlab{c}}.

\bibitem[Liang et~al.(2024)Liang, He, Jiao, Wang, Wang, Wang, Yang, Shi, and Tu]{liang2024encouraging}
Tian Liang, Zhiwei He, Wenxiang Jiao, Xing Wang, Yan Wang, Rui Wang, Yujiu Yang, Shuming Shi, and Zhaopeng Tu.
\newblock Encouraging divergent thinking in large language models through multi-agent debate.
\newblock In \emph{Proceedings of the 2024 conference on empirical methods in natural language processing}, pages 17889--17904, 2024.

\bibitem[Liao et~al.(2025)Liao, Wen, Wang, and Zhang]{liao2025marft}
Junwei Liao, Muning Wen, Jun Wang, and Weinan Zhang.
\newblock Marft: Multi-agent reinforcement fine-tuning.
\newblock \emph{arXiv preprint arXiv:2504.16129}, 2025.

\bibitem[Liu et~al.(2025{\natexlab{a}})Liu, Zeng, Xiao, He, Liu, Wang, Yan, Shen, Zhang, Xu, et~al.]{liu2025skywork}
Chris~Yuhao Liu, Liang Zeng, Yuzhen Xiao, Jujie He, Jiacai Liu, Chaojie Wang, Rui Yan, Wei Shen, Fuxiang Zhang, Jiacheng Xu, et~al.
\newblock Skywork-reward-v2: Scaling preference data curation via human-ai synergy.
\newblock \emph{arXiv preprint arXiv:2507.01352}, 2025{\natexlab{a}}.

\bibitem[Liu et~al.(2025{\natexlab{b}})Liu, Yang, Qian, Yin, Wang, Li, Liu, Zhai, Liu, and Zhang]{liu2025reinforcement}
Keliang Liu, Dingkang Yang, Ziyun Qian, Weijie Yin, Yuchi Wang, Hongsheng Li, Jun Liu, Peng Zhai, Yang Liu, and Lihua Zhang.
\newblock Reinforcement learning meets large language models: A survey of advancements and applications across the llm lifecycle.
\newblock \emph{arXiv preprint arXiv:2509.16679}, 2025{\natexlab{b}}.

\bibitem[Liu et~al.(2025{\natexlab{c}})Liu, Gao, Zhao, Zhang, Li, Qi, Ouyang, and Zhou]{liu2025can}
Runze Liu, Junqi Gao, Jian Zhao, Kaiyan Zhang, Xiu Li, Biqing Qi, Wanli Ouyang, and Bowen Zhou.
\newblock Can 1b llm surpass 405b llm? rethinking compute-optimal test-time scaling.
\newblock \emph{arXiv preprint arXiv:2502.06703}, 2025{\natexlab{c}}.

\bibitem[Liu et~al.(2025{\natexlab{d}})Liu, Du, Yang, Li, and Qiu]{Marsrl2025}
Shulin Liu, Dong Du, Tao Yang, Yang Li, and Boyu Qiu.
\newblock Marsrl: Advancing multi-agent reasoning system via reinforcement learning with agentic pipeline parallelism.
\newblock 2025{\natexlab{d}}.

\bibitem[Liu et~al.(2025{\natexlab{e}})Liu, Chen, Liang, Lyu, and Amato]{liu2025llm}
Shuo Liu, Tianle Chen, Zeyu Liang, Xueguang Lyu, and Christopher Amato.
\newblock Llm collaboration with multi-agent reinforcement learning.
\newblock \emph{arXiv preprint arXiv:2508.04652}, 2025{\natexlab{e}}.

\bibitem[Liu et~al.(2025{\natexlab{f}})Liu, Chen, Li, Qi, Pang, Du, Lee, and Lin]{liu2025understanding}
Zichen Liu, Changyu Chen, Wenjun Li, Penghui Qi, Tianyu Pang, Chao Du, Wee~Sun Lee, and Min Lin.
\newblock Understanding r1-zero-like training: A critical perspective.
\newblock \emph{arXiv preprint arXiv:2503.20783}, 2025{\natexlab{f}}.

\bibitem[Luo et~al.(2023)Luo, Sun, Xu, Zhao, Lou, Tao, Geng, Lin, Chen, and Zhang]{luo2023wizardmath}
Haipeng Luo, Qingfeng Sun, Can Xu, Pu~Zhao, Jianguang Lou, Chongyang Tao, Xiubo Geng, Qingwei Lin, Shifeng Chen, and Dongmei Zhang.
\newblock Wizardmath: Empowering mathematical reasoning for large language models via reinforced evol-instruct.
\newblock \emph{arXiv preprint arXiv:2308.09583}, 2023.

\bibitem[Luo et~al.(2025)Luo, Tan, Huang, Patel, Ariyak, Wu, Shi, Xin, Cai, Weber, et~al.]{deepcoder2025}
Michael Luo, Sijun Tan, Roy Huang, Ameen Patel, Alpay Ariyak, Qingyang Wu, Xiaoxiang Shi, Rachel Xin, Colin Cai, Maurice Weber, et~al.
\newblock Deepcoder: A fully open-source 14b coder at o3-mini level.
\newblock \emph{Notion Blog}, 2025.

\bibitem[{Moonshot AI}(2026)]{moonshotai_kimi_k2.5}
{Moonshot AI}.
\newblock Kimi k2.5: Visual agentic intelligence.
\newblock \url{https://www.kimi.com/blog/kimi-k2-5.html}, 2026.

\bibitem[Muennighoff et~al.(2025)Muennighoff, Yang, Shi, Li, Fei-Fei, Hajishirzi, Zettlemoyer, Liang, Cand{\`e}s, and Hashimoto]{muennighoff2025s1}
Niklas Muennighoff, Zitong Yang, Weijia Shi, Xiang~Lisa Li, Li~Fei-Fei, Hannaneh Hajishirzi, Luke Zettlemoyer, Percy Liang, Emmanuel Cand{\`e}s, and Tatsunori~B Hashimoto.
\newblock s1: Simple test-time scaling.
\newblock In \emph{Proceedings of the 2025 Conference on Empirical Methods in Natural Language Processing}, pages 20286--20332, 2025.

\bibitem[Nguyen et~al.(2024)Nguyen, Mekala, Dong, and Shang]{nguyen2024consistent}
Alex Nguyen, Dheeraj Mekala, Chengyu Dong, and Jingbo Shang.
\newblock When is the consistent prediction likely to be a correct prediction?
\newblock \emph{arXiv preprint arXiv:2407.05778}, 2024.

\bibitem[Pan et~al.(2025)Pan, Cemri, Agrawal, Yang, Chopra, Tiwari, Keutzer, Parameswaran, Ramchandran, Klein, et~al.]{pan2025multiagent}
Melissa~Z Pan, Mert Cemri, Lakshya~A Agrawal, Shuyi Yang, Bhavya Chopra, Rishabh Tiwari, Kurt Keutzer, Aditya Parameswaran, Kannan Ramchandran, Dan Klein, et~al.
\newblock Why do multiagent systems fail?
\newblock In \emph{ICLR 2025 Workshop on Building Trust in Language Models and Applications}, 2025.

\bibitem[Park et~al.(2025)Park, Han, Guo, Ozdaglar, Zhang, and Kim]{park2025maporl}
Chanwoo Park, Seungju Han, Xingzhi Guo, Asuman~E Ozdaglar, Kaiqing Zhang, and Joo-Kyung Kim.
\newblock Maporl: Multi-agent post-co-training for collaborative large language models with reinforcement learning.
\newblock In \emph{Proceedings of the 63rd Annual Meeting of the Association for Computational Linguistics (Volume 1: Long Papers)}, pages 30215--30248, 2025.

\bibitem[Parmar et~al.(2025)Parmar, Liu, Goyal, Chen, Le, Mishra, Mobahi, Gu, Wang, Nakhost, et~al.]{parmar2025plangen}
Mihir Parmar, Xin Liu, Palash Goyal, Yanfei Chen, Long Le, Swaroop Mishra, Hossein Mobahi, Jindong Gu, Zifeng Wang, Hootan Nakhost, et~al.
\newblock Plangen: A multi-agent framework for generating planning and reasoning trajectories for complex problem solving.
\newblock \emph{arXiv preprint arXiv:2502.16111}, 2025.

\bibitem[Rafailov et~al.(2023)Rafailov, Sharma, Mitchell, Manning, Ermon, and Finn]{rafailov2023direct}
Rafael Rafailov, Archit Sharma, Eric Mitchell, Christopher~D Manning, Stefano Ermon, and Chelsea Finn.
\newblock Direct preference optimization: Your language model is secretly a reward model.
\newblock \emph{Advances in neural information processing systems}, 36:\penalty0 53728--53741, 2023.

\bibitem[Schulman et~al.(2017)Schulman, Wolski, Dhariwal, Radford, and Klimov]{schulman2017proximal}
John Schulman, Filip Wolski, Prafulla Dhariwal, Alec Radford, and Oleg Klimov.
\newblock Proximal policy optimization algorithms.
\newblock \emph{arXiv preprint arXiv:1707.06347}, 2017.

\bibitem[Shao et~al.(2024)Shao, Wang, Zhu, Xu, Song, Bi, Zhang, Zhang, Li, Wu, et~al.]{shao2024deepseekmath}
Zhihong Shao, Peiyi Wang, Qihao Zhu, Runxin Xu, Junxiao Song, Xiao Bi, Haowei Zhang, Mingchuan Zhang, YK~Li, Yang Wu, et~al.
\newblock Deepseekmath: Pushing the limits of mathematical reasoning in open language models.
\newblock \emph{arXiv preprint arXiv:2402.03300}, 2024.

\bibitem[Snell et~al.(2025)Snell, Lee, Xu, and Kumar]{snell2025scaling}
Charlie~Victor Snell, Jaehoon Lee, Kelvin Xu, and Aviral Kumar.
\newblock Scaling llm test-time compute optimally can be more effective than scaling parameters for reasoning.
\newblock In \emph{The Thirteenth International Conference on Learning Representations}, 2025.

\bibitem[Sun et~al.(2024)Sun, Huang, and Pompili]{sun2024llm}
Chuanneng Sun, Songjun Huang, and Dario Pompili.
\newblock Llm-based multi-agent reinforcement learning: Current and future directions.
\newblock \emph{arXiv preprint arXiv:2405.11106}, 2024.

\bibitem[Wan et~al.(2025)Wan, Gao, Mu, Nakov, Wang, and Chen]{wan2025fano}
Kaiyang Wan, Lang Gao, Honglin Mu, Preslav Nakov, Yuxia Wang, and Xiuying Chen.
\newblock A fano-style accuracy upper bound for llm single-pass reasoning in multi-hop qa.
\newblock \emph{arXiv preprint arXiv:2509.21199}, 2025.

\bibitem[Wang et~al.(2024)Wang, Li, Shao, Xu, Dai, Li, Chen, Wu, and Sui]{wang2024math}
Peiyi Wang, Lei Li, Zhihong Shao, Runxin Xu, Damai Dai, Yifei Li, Deli Chen, Yu~Wu, and Zhifang Sui.
\newblock Math-shepherd: Verify and reinforce llms step-by-step without human annotations.
\newblock In \emph{Proceedings of the 62nd Annual Meeting of the Association for Computational Linguistics (Volume 1: Long Papers)}, pages 9426--9439, 2024.

\bibitem[Wang et~al.(2022)Wang, Wei, Schuurmans, Le, Chi, Narang, Chowdhery, and Zhou]{wang2022self}
Xuezhi Wang, Jason Wei, Dale Schuurmans, Quoc Le, Ed~Chi, Sharan Narang, Aakanksha Chowdhery, and Denny Zhou.
\newblock Self-consistency improves chain of thought reasoning in language models.
\newblock \emph{arXiv preprint arXiv:2203.11171}, 2022.

\bibitem[Wei et~al.(2022)Wei, Wang, Schuurmans, Bosma, Xia, Chi, Le, Zhou, et~al.]{wei2022chain}
Jason Wei, Xuezhi Wang, Dale Schuurmans, Maarten Bosma, Fei Xia, Ed~Chi, Quoc~V Le, Denny Zhou, et~al.
\newblock Chain-of-thought prompting elicits reasoning in large language models.
\newblock \emph{Advances in neural information processing systems}, 35:\penalty0 24824--24837, 2022.

\bibitem[Wu et~al.(2023)Wu, Bansal, Zhang, Wu, Li, Zhu, Jiang, Zhang, Zhang, Liu, Awadallah, White, Burger, and Wang]{wu2023autogenenablingnextgenllm}
Qingyun Wu, Gagan Bansal, Jieyu Zhang, Yiran Wu, Beibin Li, Erkang Zhu, Li~Jiang, Xiaoyun Zhang, Shaokun Zhang, Jiale Liu, Ahmed~Hassan Awadallah, Ryen~W White, Doug Burger, and Chi Wang.
\newblock Autogen: Enabling next-gen llm applications via multi-agent conversation, 2023.
\newblock URL \url{https://arxiv.org/abs/2308.08155}.

\bibitem[Xia et~al.(2025)Xia, Qin, Li, Ma, Fan, Chern, Zou, Zhou, Hu, Jin, et~al.]{xia2025generative}
Shijie Xia, Yiwei Qin, Xuefeng Li, Yan Ma, Run-Ze Fan, Steffi Chern, Haoyang Zou, Fan Zhou, Xiangkun Hu, Jiahe Jin, et~al.
\newblock Generative ai act ii: Test time scaling drives cognition engineering.
\newblock \emph{arXiv preprint arXiv:2504.13828}, 2025.

\bibitem[Xie et~al.(2024)Xie, Goyal, Zheng, Kan, Lillicrap, Kawaguchi, and Shieh]{xie2024monte}
Yuxi Xie, Anirudh Goyal, Wenyue Zheng, Min-Yen Kan, Timothy~P Lillicrap, Kenji Kawaguchi, and Michael Shieh.
\newblock Monte carlo tree search boosts reasoning via iterative preference learning.
\newblock \emph{arXiv preprint arXiv:2405.00451}, 2024.

\bibitem[Xue et~al.(2025)Xue, Zhou, Zhang, Zhang, Li, Zhang, Yin, Torr, Ouyang, and Bai]{xue2025comas}
Xiangyuan Xue, Yifan Zhou, Guibin Zhang, Zaibin Zhang, Yijiang Li, Chen Zhang, Zhenfei Yin, Philip Torr, Wanli Ouyang, and Lei Bai.
\newblock Comas: Co-evolving multi-agent systems via interaction rewards.
\newblock \emph{arXiv preprint arXiv:2510.08529}, 2025.

\bibitem[Yang et~al.(2025)Yang, Li, Yang, Zhang, Hui, Zheng, Yu, Gao, Huang, Lv, et~al.]{yang2025qwen3}
An~Yang, Anfeng Li, Baosong Yang, Beichen Zhang, Binyuan Hui, Bo~Zheng, Bowen Yu, Chang Gao, Chengen Huang, Chenxu Lv, et~al.
\newblock Qwen3 technical report.
\newblock \emph{arXiv preprint arXiv:2505.09388}, 2025.

\bibitem[Yao et~al.(2025)Yao, Liu, Zhang, Dong, Shang, and Gao]{yaoyour}
Feng Yao, Liyuan Liu, Dinghuai Zhang, Chengyu Dong, Jingbo Shang, and Jianfeng Gao.
\newblock Your efficient rl framework secretly brings you off-policy rl training, august 2025.
\newblock \emph{URL https://fengyao. notion. site/off-policy-rl}, 2025.

\bibitem[Yao et~al.(2023)Yao, Yu, Zhao, Shafran, Griffiths, Cao, and Narasimhan]{yao2023tree}
Shunyu Yao, Dian Yu, Jeffrey Zhao, Izhak Shafran, Tom Griffiths, Yuan Cao, and Karthik Narasimhan.
\newblock Tree of thoughts: Deliberate problem solving with large language models.
\newblock \emph{Advances in neural information processing systems}, 36:\penalty0 11809--11822, 2023.

\bibitem[Yao et~al.(2024)Yao, Li, and Zhao]{yao2024got}
Yao Yao, Zuchao Li, and Hai Zhao.
\newblock Got: Effective graph-of-thought reasoning in language models.
\newblock In \emph{Findings of the Association for Computational Linguistics: NAACL 2024}, pages 2901--2921, 2024.

\bibitem[Yu et~al.(2025)Yu, Zhang, Zhu, Yuan, Zuo, Yue, Dai, Fan, Liu, Liu, et~al.]{yu2025dapo}
Qiying Yu, Zheng Zhang, Ruofei Zhu, Yufeng Yuan, Xiaochen Zuo, Yu~Yue, Weinan Dai, Tiantian Fan, Gaohong Liu, Lingjun Liu, et~al.
\newblock Dapo: An open-source llm reinforcement learning system at scale.
\newblock \emph{arXiv preprint arXiv:2503.14476}, 2025.

\bibitem[Yuan et~al.(2025)Yuan, Yue, Zhu, Fan, and Yan]{yuan2025s}
Yufeng Yuan, Yu~Yue, Ruofei Zhu, Tiantian Fan, and Lin Yan.
\newblock What's behind ppo's collapse in long-cot? value optimization holds the secret.
\newblock \emph{arXiv preprint arXiv:2503.01491}, 2025.

\bibitem[Zhang et~al.(2024)Zhang, Huang, Zhou, Li, and Ouyang]{zhang2024accessing}
Di~Zhang, Xiaoshui Huang, Dongzhan Zhou, Yuqiang Li, and Wanli Ouyang.
\newblock Accessing gpt-4 level mathematical olympiad solutions via monte carlo tree self-refine with llama-3 8b.
\newblock \emph{arXiv preprint arXiv:2406.07394}, 2024.

\bibitem[Zhang et~al.(2025{\natexlab{a}})Zhang, Geng, Yu, Yin, Zhang, Tan, Zhou, Li, Xue, Li, et~al.]{zhang2025landscape}
Guibin Zhang, Hejia Geng, Xiaohang Yu, Zhenfei Yin, Zaibin Zhang, Zelin Tan, Heng Zhou, Zhongzhi Li, Xiangyuan Xue, Yijiang Li, et~al.
\newblock The landscape of agentic reinforcement learning for llms: A survey.
\newblock \emph{arXiv preprint arXiv:2509.02547}, 2025{\natexlab{a}}.

\bibitem[Zhang et~al.(2025{\natexlab{b}})Zhang, Liu, Zhu, Tian, Zeng, Jia, Fan, Lv, Zuo, Jiang, Liu, Wang, Wang, Zhao, Hua, Wang, Wang, Gao, Long, Sun, Ma, Cui, Bai, Ding, Qi, and Zhou]{marti2025}
Kaiyan Zhang, Runze Liu, Xuekai Zhu, Kai Tian, Sihang Zeng, Guoli Jia, Yuchen Fan, Xingtai Lv, Yuxin Zuo, Che Jiang, Ziyang Liu, Jianyu Wang, Yuru Wang, Ruotong Zhao, Ermo Hua, Yibo Wang, Shijie Wang, Junqi Gao, Xinwei Long, Youbang Sun, Zhiyuan Ma, Ganqu Cui, Lei Bai, Ning Ding, Biqing Qi, and Bowen Zhou.
\newblock Marti: A framework for multi-agent llm systems reinforced training and inference, 2025{\natexlab{b}}.
\newblock URL \url{https://github.com/TsinghuaC3I/MARTI}.

\bibitem[Zhang et~al.(2025{\natexlab{c}})Zhang, Li, Li, Dong, and Jin]{zhang2025focused}
Kechi Zhang, Ge~Li, Jia Li, Yihong Dong, and Zhi Jin.
\newblock Focused-dpo: Enhancing code generation through focused preference optimization on error-prone points.
\newblock \emph{arXiv preprint arXiv:2502.11475}, 2025{\natexlab{c}}.

\bibitem[Zhang et~al.(2025{\natexlab{d}})Zhang, Lyu, Sun, Wang, Zhang, Guo, Wang, King, Liu, and Ma]{zhang2025and}
Qiyuan Zhang, Fuyuan Lyu, Zexu Sun, Lei Wang, Weixu Zhang, Zhihan Guo, Yufei Wang, Irwin King, Xue Liu, and Chen Ma.
\newblock What, how, where, and how well? a survey on test-time scaling in large language models.
\newblock \emph{CoRR}, 2025{\natexlab{d}}.

\bibitem[Zhang et~al.(2025{\natexlab{e}})Zhang, Lyu, Sun, Wang, Zhang, Hua, Wu, Guo, Wang, Muennighoff, et~al.]{zhang2025survey}
Qiyuan Zhang, Fuyuan Lyu, Zexu Sun, Lei Wang, Weixu Zhang, Wenyue Hua, Haolun Wu, Zhihan Guo, Yufei Wang, Niklas Muennighoff, et~al.
\newblock A survey on test-time scaling in large language models: What, how, where, and how well?
\newblock \emph{arXiv preprint arXiv:2503.24235}, 2025{\natexlab{e}}.

\bibitem[Zhao et~al.(2025)Zhao, Hu, Wang, Hou, Zhang, Ding, and Zhao]{zhao2025stronger}
Yujie Zhao, Lanxiang Hu, Yang Wang, Minmin Hou, Hao Zhang, Ke~Ding, and Jishen Zhao.
\newblock Stronger-mas: Multi-agent reinforcement learning for collaborative llms.
\newblock \emph{arXiv preprint arXiv:2510.11062}, 2025.

\bibitem[Zheng et~al.(2025)Zheng, Liu, Li, Chen, Yu, Gao, Dang, Liu, Men, Yang, et~al.]{zheng2025group}
Chujie Zheng, Shixuan Liu, Mingze Li, Xiong-Hui Chen, Bowen Yu, Chang Gao, Kai Dang, Yuqiong Liu, Rui Men, An~Yang, et~al.
\newblock Group sequence policy optimization.
\newblock \emph{arXiv preprint arXiv:2507.18071}, 2025.

\bibitem[Zhou et~al.(2023)Zhou, Yan, Shlapentokh-Rothman, Wang, and Wang]{zhou2023language}
Andy Zhou, Kai Yan, Michal Shlapentokh-Rothman, Haohan Wang, and Yu-Xiong Wang.
\newblock Language agent tree search unifies reasoning acting and planning in language models.
\newblock \emph{arXiv preprint arXiv:2310.04406}, 2023.

\bibitem[Zhoubian et~al.(2025)Zhoubian, Zhang, and Tang]{zhoubian2025rest}
Sining Zhoubian, Dan Zhang, and Jie Tang.
\newblock Rest-rl: Achieving accurate code reasoning of llms with optimized self-training and decoding.
\newblock \emph{arXiv preprint arXiv:2508.19576}, 2025.

\bibitem[Zuo et~al.(2025)Zuo, Zhang, Sheng, Qu, Cui, Zhu, Li, Zhang, Long, Hua, et~al.]{zuo2025ttrl}
Yuxin Zuo, Kaiyan Zhang, Li~Sheng, Shang Qu, Ganqu Cui, Xuekai Zhu, Haozhan Li, Yuchen Zhang, Xinwei Long, Ermo Hua, et~al.
\newblock Ttrl: Test-time reinforcement learning.
\newblock \emph{arXiv preprint arXiv:2504.16084}, 2025.

\end{thebibliography}
